\documentclass[preprint,5p,times,twocolumn]{elsarticle}

\usepackage{amsmath}
\usepackage{amsthm}
\usepackage{subcaption}
\usepackage{microtype}
\usepackage{booktabs}
\usepackage{tikz}
\usetikzlibrary{calc, arrows}
\usepackage[colorlinks]{hyperref}

\newcommand{\Real}{\mathbb{R}}
\newcommand{\Natural}{\mathbb{N}}

\newcommand{\sign}{\mathrm{sign}}

\newcommand{\plane}{\pi}
\newcommand{\sfunc}{\phi}
\newcommand{\planenormal}{\nu}
\newcommand{\proj}{\Pi}
\newcommand{\sproj}{\Gamma}
\newcommand{\distribution}{\mathcal{D}}
\newcommand{\accuracy}{\operatorname{acc}}
\newcommand{\susceptibility}{\operatorname{sus}}
\newcommand{\undetectability}{\operatorname{rand}}
\newcommand{\notword}{\operatorname{not}}

\newcommand{\suchthat}{\,:\,}
\newcommand{\given}{\,\,|\,\,}

\newtheorem{theorem}{Theorem}
\newtheorem{lemma}[theorem]{Lemma}

\newtheorem{definition}[theorem]{Definition}
\newtheorem{corollary}[theorem]{Corollary}

\begin{document}

\begin{frontmatter}

\title{How Adversarial Attacks Can Disrupt Seemingly Stable Accurate Classifiers}

\author[kings]{Oliver J. Sutton}
\author[kings]{Qinghua Zhou}
\author[kings]{Ivan Y. Tyukin}
\author[leicester]{Alexander N. Gorban}
\author[kings]{Alexander Bastounis}
\author[edinburgh]{Desmond J. Higham}

\affiliation[kings]{
    organization={Department of Mathematics, King's College London},
    city={London},
    country={UK}
}

\affiliation[leicester]{
    organization={School of Computing and Mathematical Sciences, University of Leicester},
    city={Leicester},
    country={UK}
}

\affiliation[edinburgh]{
    organization={School of Mathematics, University of Edinburgh},
    city={Edinburgh},
    country={UK}
}

\begin{abstract}
    Adversarial attacks dramatically change the output of an otherwise accurate learning system using a seemingly inconsequential modification to a piece of input data.
    Paradoxically, empirical evidence indicates that even systems which are robust to large random perturbations of the input data remain susceptible to small, easily constructed, adversarial perturbations of their inputs.
    Here, we show that this may be seen as a fundamental feature of classifiers working with high dimensional input data.
    We introduce a simple generic and generalisable framework for which key behaviours observed in practical systems arise with high 
    probability---notably the simultaneous susceptibility of the (otherwise accurate) model to easily constructed adversarial attacks, and robustness to random perturbations of the input data.
    We confirm that the same phenomena are directly observed in practical neural networks trained on standard image classification problems, where even large additive random noise fails to trigger the adversarial instability of the network.
    A surprising takeaway is that even small margins separating a classifier's decision surface from training and testing data can hide adversarial susceptibility from being detected using randomly sampled perturbations.
    Counter-intuitively, using additive noise during training or testing is therefore inefficient for eradicating or detecting adversarial examples, and more demanding adversarial training is required.
    \end{abstract}








\end{frontmatter}

\section{Introduction}

Adversarial attacks aim to slightly modify a piece of input data in such a way as to significantly change the output of a model. 
The sensitivity of neural networks to small perturbations like these has been widely studied since they were first reported in deep networks in~\cite{szegedy2013intriguing}.
Simple algorithms exist which enable a malicious attacker to produce adversarial perturbations quite easily in many cases
\cite{chaubey2020universal}.
Recent works~\cite{bastounis2021mathematics, Bastounis2023530} 
have shown that such instabilities are somewhat inevitable, even in relatively small networks consisting of just two layers where the number of neurons is linear in the input data dimension.
It is remarkable, therefore, that these same instabilities are rarely triggered by random perturbations to the input data -- even when these random perturbations may be much larger than destabilising adversarial perturbations.

\begin{figure*}
\begin{subfigure}{0.49\linewidth}
    \centering
    \includegraphics[width=0.9\linewidth]{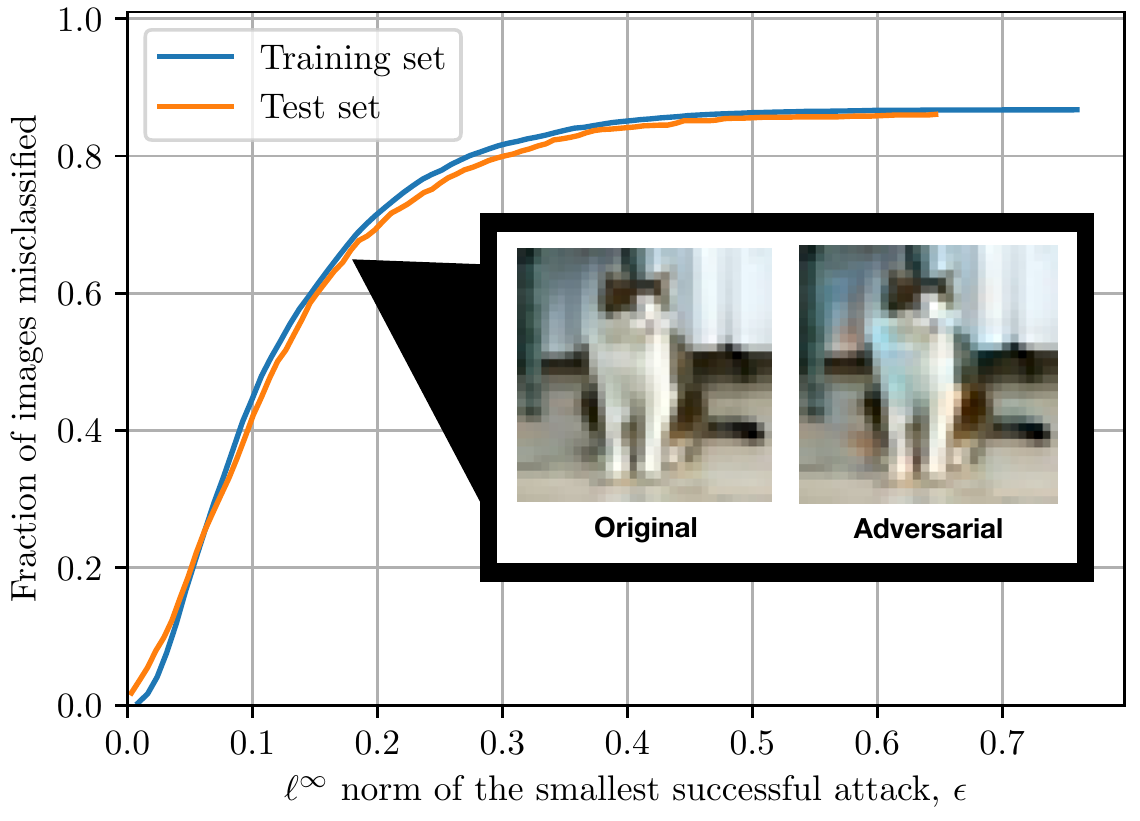}
    \caption{Cumulative histogram of sizes of successful adversarial attacks.}
    \label{fig:adversarial}
\end{subfigure}
\begin{subfigure}{0.49\linewidth}
    \centering
    \includegraphics[width=0.9\linewidth]{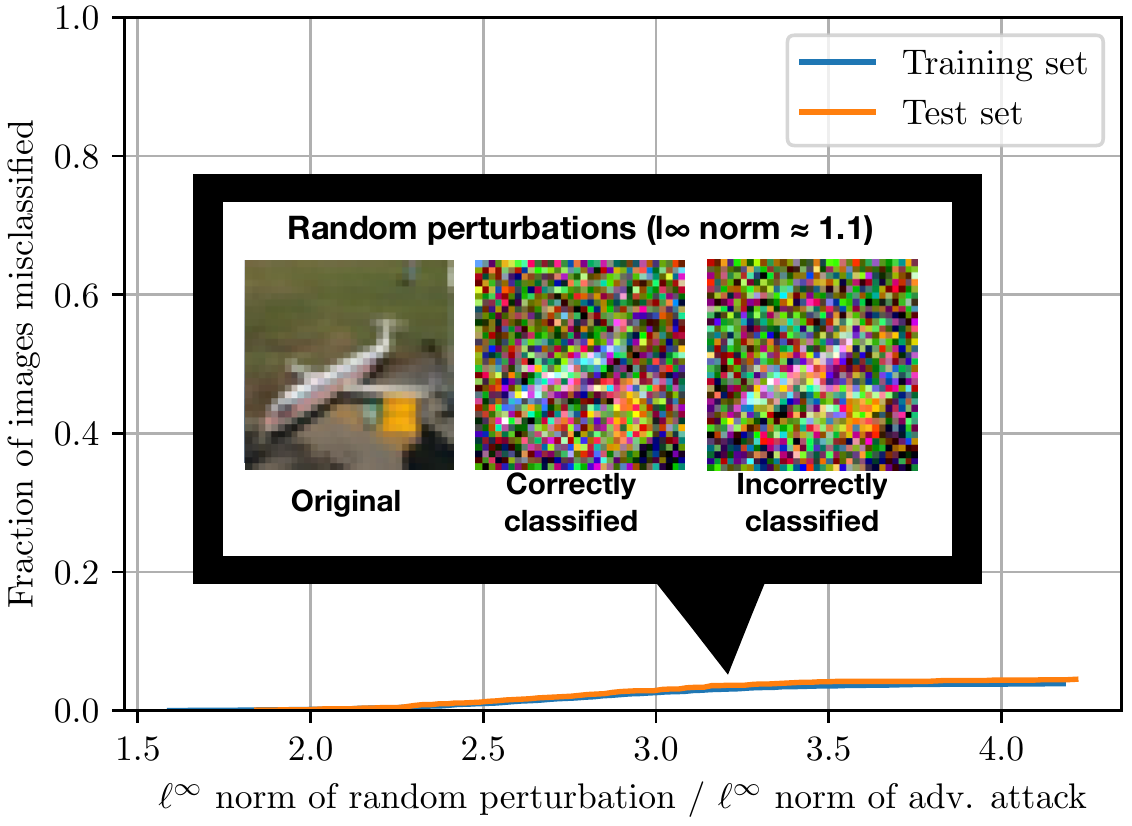}
    \caption{Cumulative histogram of sizes of successful random perturbations.}
    \label{fig:randomPerturbation}
\end{subfigure}
\caption{Histograms showing the fraction of images from the `aeroplane-vs-cat' binary classification problem (from the CIFAR-10 dataset) which were misclassified after either (a) an adversarial attack (as the fraction of ordinarily correctly classified images) or (b) a random perturbation of different sizes (as the fraction of images which were susceptible to adversarial attacks), measured as the maximum absolute change to an individual pixel channel (the $\ell^{\infty}$ norm). For adversarial attacks, this represents the smallest misclassifying attack in the adversarial direction. For the random perturbations, we record the smallest $\ell^{\infty}$ norm among 2000 misclassifying perturbations sampled from the Euclidean ball with radius $5\epsilon$, where $\epsilon$ is the Euclidean norm of the smallest successful adversarial attack found for each image. Examples are shown at the size of their respective perturbation norms. Full details of the experimental results are given in Section~\ref{sec:full-experiments}.}
\label{fig:histograms}
\end{figure*}

This \emph{paradox of apparent stability} is demonstrated in Figure~\ref{fig:histograms} for a standard convolutional neural network trained on CIFAR-10 images~\cite{cifar10}.
Although the majority of images in both the training and test data sets are susceptible to small adversarial attacks (panel (\emph{a})), random perturbations even an order of magnitude larger mostly fail to cause the images to be misclassified (panel (\emph{b})).
Further experimental results on other image classification datasets, including using pre-trained foundation models, are summarised in Table~\ref{tbl:benchmark-summary} and discussed further in Section~\ref{sec:experiments}.

Several explanations for the causes of adversarial examples have been proposed in the literature.
An early work on the subject~\cite{goodfellow2014explaining} suggested that adversarial images simply live in regions of the data space to which the data distribution assigns low probability.
A variant of this idea, discussed in~\cite{khoury2018geometry}, suggests that adversarial attacks perturb inputs in a way that moves them in an orthogonal direction to the local data manifold.
This results in adversarial images which exist in a region of data space where no training data could have been sampled, and the decision surfaces of the network are therefore relatively pathological.
Other suggested mechanisms include the dimpled manifold hypothesis \cite{shamir2021dimpled}, boundary tilting \cite{tanay2016boundary}, and the existence of uncountably large families of special distributions for which instabilities are expected \cite{bastounis2021mathematics}. 
However, none of these frameworks rigorously account for and explain the paradoxical simultaneous robustness of these classifiers to random perturbations whose size could be several times larger than that of the adversarial ones.

Here, we suggest a resolution to the paradox rooted in ideas from the
theory of concentration of measure, and related properties
of high dimensional probability distributions.
The simple, realistic framework we introduce captures the key features of the paradox which are observed in practice (precise definitions of these terms are given in Section~\ref{sec:theoretical-results}):
\begin{description}
\item[Accuracy:] The classifier correctly labels non-perturbed data.
\item[Apparent robustness/stability:] There is a vanishingly small probability that a sampled data point will be misclassified after a large random perturbation is applied to it.
\item[Vulnerability:] Yet, with high probability, any sampled data point is susceptible to a very small adversarial perturbation that changes the predicted class.
\item[Computability:] An optimal destabilizing perturbation can be computed from knowledge of the loss function gradient.
\end{description}

Our theoretical investigation reveals a tension between different notions of what it means for a classifier to be stable, a subtlety which is rarely discussed in practice. 
A problem may be \emph{deterministically unstable} in the sense that for a given data point there exists 
an arbitrarily small destabilising perturbation which may be exploited
by an attacker, while the fact that this instability is extremely unlikely to be triggered by random noise renders the problem \emph{probabilistically stable}.
This is a dangerous situation for a performance-critical classifier: even though the performance appears excellent at test time, adversarial instabilities and the lack of deterministic robustness lurk awaiting an unscrupulous attacker and cannot be efficiently detected at random.
An important feature of our theoretical framework (developed in Section~\ref{sec:theoretical-results}) for understanding the paradox of apparent stability is that it can be studied at various levels of generality.
This enables us to distil the fundamental origins of the paradox without unnecessary technical details, through a hierarchy of models incorporating different levels of complexity.

Our findings are directly supported by extensive experimental results, summarised in Section~\ref{sec:experiments} and discussed in detail in Section~\ref{sec:full-experiments}.
The results demonstrate the paradox of apparent stability, and reveal some of its real-world implications.
We show the effectiveness of simple adversarial attacks on a variety of different standard datasets and models, and contrast this against their robustness to large random perturbations.
These experiments confirm the predictions of our theoretical results: to observe cases in which random perturbations cause labels to swap, many perturbations must be taken, and with a significantly larger amplitude than that of the smallest adversarial perturbation affecting the same image. 
An immediate practical consequence of this is to shed new light on algorithms which aim to ensure or certify adversarial robustness by adding random noise to inputs, such as those discussed in~\cite{cohen2019certified, li2019certified, ye2024unit}.
Our investigation reveals that this is computationally inefficient in high dimension, since it requires an exponentially large number of perturbed samples per data point to expect to observe just one which causes a misclassification of even a highly susceptible input.
We also find that in genuinely high-dimensional settings adding random noise at training time causes a significant degradation to the trained model's accuracy, which appears to outweigh any marginal improvements in adversarial robustness. The relevant spectrum of tasks to which the above applies includes popular image classification problems. This implies that data pre-processing involving an appropriate dimensionality reduction may be needed to bring out the benefits of robustness induced by random data augmentation at training.

To study the paradox of apparent stability, we begin with a simple model which nonetheless exhibits key features of the paradox, and build it up to expose how different phenomena appear and persist as the model becomes more general.
We first consider a single fixed data point (i.e. without any data sampling process) sitting close to a (locally) linear decision surface in Section~\ref{sec:fixedDataPoint}.
We prove that in this situation, the probability of randomly sampling a perturbation which causes the data point to be misclassified is exponentially small in the data dimension.
This already shows that dimensionality is a fundamental component in trying to understand the relationship between random and adversarial perturbations, and therefore in resolving the paradox of apparent stability.
This result also shows that algorithms which aim to detect or defend against adversarial susceptibility using random data perturbations, such as those discussed in~\cite{cohen2019certified, li2019certified, ye2024unit}, may in fact require computational complexities which are exponentially large with respect to the data dimension.

We build on this in Section~\ref{sec:two-balls-model} by considering a binary classification problem with a linear classifier.
Data from both classes are sampled from distributions satisfying a mild non-degeneracy condition, formulated as a simplified version of the Smeared Absolute Continuity (SmAC) condition introduced in~\cite{gorban2018correction} (see Definition~\ref{def:smac}).
Despite its simplicity, we prove that this setting already exhibits the four characteristics above of the paradox of apparent stability.
We then significantly generalise the setup in Section~\ref{sec:general-model} to show that these phenomena persist when data are sampled from two arbitrary distributions and classified using nonlinear decision surfaces.
Once again, this setup reveals the same fundamental characteristics of the paradox of apparent stability.
This setup admits various further generalisations, which are discussed in Section~\ref{sec:commentsOnGeneralisbility}.

As a counterpoint to the findings discussed above, Section~\ref{sec:hemispheres_model} reveals a subtle, yet important, modification which can be made to the setup which causes the discrepancy between adversarial and random perturbations to disappear.
Specifically, we construct a scenario in which the typical distance from a sampled data point to the classifiers (linear) decision boundary approaches zero in high dimensions.
In this case, the probability of a random perturbation to input data causing a misclassification is separated away from zero by a constant for arbitrarily large data dimension, rather than exponentially decreasing as in the previous scenarios.
Having small distances from typical data points to the decision surface is clearly undesirable in any practical application, since it means that the classifier itself is extremely sensitive to small changes in the data.
However, this setup reveals that it is in fact the presence of an appropriate margin between typical data points and the decision surface which manifests itself as the paradox of simultaneous apparent robustness to large random perturbations and vulnerability to small adversarial attacks.

The paper is organised as follows.
In Section~\ref{sec:notation} we introduce mathematical notation that is used throughout the text.
Section \ref{sec:experiments} gives an overview of the paradox of apparent stability as it manifests itself in real image classification problems.
Our main theoretical results are presented in Section~\ref{sec:theoretical-results}, introducing a hierarchy of simple models which expose the fundamental origins of the paradox.
The alternative model analysed in Section~\ref{sec:hemispheres_model} reveals the link between the existence of non-zero classification margins and the ability to determine susceptibility to adversarial examples using random perturbations.
In Section~\ref{sec:discussion} we discuss these analytical and empirical findings and relate them to prior work and knowledge in the area. 
Section~\ref{sec:full-experiments} provides a comprehensive description of our experimental setup and numerical results.
Section~\ref{sec:conclusion} concludes the paper.
Proofs of all statements and auxiliary technical results are provided in the Appendix.

\section{Notation}\label{sec:notation}

We use the following notation throughout:
\begin{itemize}
\item 
$x \cdot y$ denotes the inner product of 
$x,y \in \Real^n$ and 
$\|x\| = \sqrt{x \cdot y}$ denotes the Euclidean ($\ell^2$) norm,
\item the $\ell^{1}$ norm of a vector in $\mathbb{R}^n$ is defined to be the sum of the absolute values of its components,
\item the $\ell^{\infty}$ norm of a vector in $\mathbb{R}^n$ is defined to be the maximum of the absolute values of its components,
\item $\mathbb{B}^n_r(c) = \{x \in \Real^n \suchthat \|x - c\| \leq r\}$ denotes the the Euclidean ball in $\Real^n$ with radius $r > 0$ centered at $c \in \Real^n$, and we use the abbreviation $\mathbb{B}^n = \mathbb{B}^n_1(0)$,
\item $V^n = \frac{\pi^{\frac{n}{2}}}{\Gamma(\frac{n}{2} + 1)}$ denotes the $n$-dimensional volume of $\mathbb{B}^n$ (the unit ball, usually assumed centred at $0$), and $V^n_{\operatorname{cap}}(r, h)$ denotes the volume of the cap with height $h$ of the $n$-dimensional ball of radius $r$, i.e. the volume of the set $\{x \in \mathbb{R}^n \suchthat \|x\| < r \text{ and } x_1 > 1 - h\}$, where $x_1 = x \cdot \mathbf{e}_1$ and $\mathbf{e}_1 = (1, 0, \dots, 0)^{\top} \in \mathbb{R}^n$; if $S\subset\Real^n$ then $V^n(S)$ denotes the $n$-dimensional volume of the set $S$,
\item for a set $S \subset \mathbb{R}^n$, we use $\mathcal{U}(S)$ to denote the uniform distribution on $S$, and $\mathbb{I}_S : S \to \{0, 1\}$ to denote the indicator function of $S$, such that $\mathbb{I}_S(x) = 1$ for $x \in S$ and 0 otherwise,
\item the function $\Phi : \mathbb{R} \to \mathbb{R}$ denotes the standard Gaussian cumulative distribution function 
\[
\Phi(s)=\frac{1}{\sqrt{2\pi}}\int_{-\infty}^{s} e^{-\frac{\xi^2}{2}}d\xi.
\]
\end{itemize}

\section{The paradox of apparent stability demonstrated on standard datasets}\label{sec:experiments}

The phenomenon of simultaneous susceptibility to adversarial attacks and robustness to random noise can be clearly demonstrated using standard image classification benchmark datasets.
Here, we summarise results presented in detail in Section~\ref{sec:full-experiments}, calculated using CIFAR-10~\cite{cifar10}, Fashion MNIST~\cite{fashionMNIST}, the German Traffic Sign Recognition Benchmark (GTSRB)~\cite{GTSRB}, and ImageNet~\cite{imagenet}.
Our experimental setup is described in detail in Section~\ref{sec:experimentalSetup}.

To present the phenomenon in the simplest possible setting, we split {each of CIFAR-10, Fashion-MNIST and GTSRB} into a set of binary classification problems, one for each pair of classes in the dataset.
A separate network (each with the same convolutional structure in the form of a truncated VGG network~\cite{simonyan2015deep}) was trained using Tensorflow~\cite{tensorflow} for each of these problems, and each point in the training and test set was assessed for its susceptibility to adversarial examples using a gradient-based attack algorithm. 
On images which were susceptible to an adversarial attack with Euclidean norm $\epsilon$, we applied 2000 perturbations randomly sampled from the Euclidean ball with radius $\delta\epsilon$ for each value of $\delta$ in the set $\{1, 2, 5, 10\}$. 
This measures the sensitivity of the network to random perturbations around the training and test images.

We complement this investigation of binary classification problems with an analogous analysis of the adversarial and random susceptibility of pre-trained models (with VGG19~\cite{simonyan2015deep} and ResNet50~\cite{he2016deep} architectures, from Tensorflow) using images from the ImageNet validation set.
These models are trained to classify ImageNet images into 1,000 classes, and demonstrate the presence of the paradox of apparent stability in real-world models.

\begin{table*}[!h]
    \centering
    \scalebox{0.85}{
    \begin{tabular}{c|c|c|c|c|c}
         & \textbf{CIFAR-10} & \textbf{Fashion MNIST} & \textbf{GTSRB} & \textbf{ImageNet} (ResNet50) & \textbf{ImageNet} (VGG19) \\
        \midrule
        Accuracy  & 99.70\%, 95.80\% & 99.51\%, 99.4\%  & 98.32\%, 98.51\% & -, 70.8\% & -, 66.52\%  \\
        Adversarial attack susceptibility & 91.88\%, 89.96\% & 53.58\%, 53.01\% & 77.53\%, 77.00\% & -, 94.2\% & -, 97.07\% \\
        Random attack susceptibility ($\delta = 2$) & 0.02\%, 0.17\% & 0.07\%, 0.09\%     & 0.36\%, 0.36\% & - & -  \\
        Random attack susceptibility ($\delta = 5$) & 2.65\%, 2.57\% & 10.71\%, 13.35\%     & 5.76\%, 5.1\% & - & -  \\
        Random attack susceptibility ($\delta = 10$) & 41.19\%, 40.57\% & 56.84\%, 57.43\%     & 39.26\%, 36.07\% & -, 2.5\% & -, 1.4\% \\
        Input dimension & $32\times 32\times 3$ & $28\times28\times1$     & $30\times 30\times 3$ & $224 \times 224 \times 3$ & $224 \times 224 \times 3$ \\
        Number of classes & 10 & 10 & 6 & 1000 & 1000
    \end{tabular}
    }
    \caption{A summary of the performance of networks trained on different standard image classification benchmark datasets. We split each dataset into a set of binary classification problems, and the results in this table are reported in the form `train, test' and as the median over the binary classification problems of the form `class $i$-vs-class $i+1$' within each dataset.
    The symbol `-' denotes values which were not computed.
    Full details of the experimental setup and results are given in Sections~\ref{sec:experimentalSetup} and \ref{sec:experimentalResults}. Adversarial attack susceptibility is measured as the proportion of images in both classes of each problem which were susceptible to an adversarial attack. Random attack susceptibility with fixed $\delta$ is measured as the proportion of adversarially susceptible images which were misclassified after applying any of 2,000 randomly sampled perturbations with Euclidean norm up to $\delta$ times as large as that of the smallest adversarial perturbation identified on that image. The results on GTSRB were computed using a subset of classes in the original dataset, see Section~\ref{sec:experimentResults:gtsrb} for details. Results for ImageNet were evaluated using pre-trained ResNet50 and VGG19 models from Tensorflow on images from the ImageNet validation set, as described in Section~\ref{sec:experimentResults:imagenet}.}
    \label{tbl:benchmark-summary}
\end{table*}

The 
empirical essence of the
phenomenon is illustrated in Figure~\ref{fig:histograms} using the CIFAR-10 dataset: while the networks were easily fooled by relatively small adversarial perturbations which appear to make little perceptual difference to the image, they were remarkably robust to randomly sampled perturbations.
Here we demonstrate this in the broadly representative case of the `aeroplane-vs-cat' binary classification problem.
Comparing the inset examples in Figures~\ref{fig:adversarial} and~\ref{fig:randomPerturbation},  the modification made by the adversarial perturbation 
does not alter the overall perception of the image as that of a `cat'. Moreover, it is difficult to tell by just looking at these images which one of them has been subjected to an adversarial attack.
It is nearly equally difficult to make out the aeroplane in the (correctly classified) randomly perturbed image.
Note that, since the original images have pixel channel values in $[0, 1]$, a perturbation with $\ell^{\infty}$ norm greater than 1 represents a drastic change to the contents of the image, yet one which was rarely able to cause the network to misclassify its input.

A summary of the accuracy and susceptibility of the classifiers is presented in Table~\ref{tbl:benchmark-summary}, although the figures alone make it clear that even when the random perturbations are sampled to be five times as large as the known adversarial perturbation ($\delta = 5$), they still mostly fail to trigger the adversarial susceptibility of the network.
The effects are broadly consistent across all the datasets we examined, with negligible difference between the training and test data.
Further details of the experimental setup and full results for this and the remaining classification problems are explored in Section~\ref{sec:experimentalResults}.

We also provide the results of experiments on CIFAR-10 into incorporating additive random noise to data at training time to assess the impact this may have on adversarial susceptibility (the experimental setup is described in Section~\ref{sec:supplementary:randomTraining} and the results are presented in Section~\ref{sec:experimentResults:cifar10}).
The conclusion we draw from these experiments is that training with even large random perturbations does not significantly decrease the susceptibility to adversarial attacks, and is responsible for a large drop in accuracy.

\section{The essence of the paradox}\label{sec:theoretical-results}

To understand the origins of the paradox of apparent stability, we show how a hierarchy of simple, yet reasonably generic, theoretical models can explain the behaviour observed empirically.
First, in Section~\ref{sec:fixedDataPoint} we show that for a fixed data point close to a model's (locally linear) decision boundary, randomly sampled noise is very unlikely to detect adversarial instabilities in high dimensions.
Since this example does not assume any sampling distribution for the data point, it provides a generic setup for understanding the difference between random and adversarial perturbations.

We generalise this to a second scenario in Section~\ref{sec:two-balls-model}, where data from two classes are sampled from a reasonably general class of distributions, and classified using a linear classifier.
The data distributions are only assumed to satisfy a mild non-degeneracy condition (known as the SmAC condition; Definition~\ref{def:smac}).
Despite its apparent simplicity, this setup already simultaneously presents all of the symptoms of the paradox of apparent stability: with high probability, data points are accurately classified (Theorem~\ref{thm:accuracy}) and susceptible to small adversarial perturbations (Theorem~\ref{thm:susceptibility}), yet with high probability randomly sampled perturbations do not cause data to be misclassified (Theorem~\ref{thm:undetectability}).
Moreover, gradient-based algorithms are efficient for constructing the adversarial attack (Theorem~\ref{thm:gradientBasedAttack}), and successful attacks are even universal in sense that they also cause other data points with the same class to be misclassified with high probability (Theorem~\ref{thm:universality}).

This setup is generalised further in Section~\ref{sec:general-model}, providing versions of the same key results.
No assumptions are placed on the data distributions, and the results require only that the classifier's decision boundary is a Lipschitz warping of a plane in its normal direction.
We also show how the results from Section~\ref{sec:two-balls-model} may be obtained as corollaries of these general results.

Further generalisations of our results are considered in Section~\ref{sec:commentsOnGeneralisbility}.

\begin{figure}
    \begin{center}
        \begin{tikzpicture}[scale=0.75]
            \def\outerRadius{2cm}
            \def\innerRadius{2cm}
            \def\separatorX{0.85}
            \coordinate (y) at (3, 0);
            \coordinate (c) at (0, 0);
            \coordinate (separatorTop) at (0.75, 2.2);
            \coordinate (separatorBottom) at (0.75, -2.2);
            \coordinate (separatorMidpoint) at ($(separatorTop)!0.5!(separatorBottom)$);
            \coordinate (midpoint) at ($(c)!0.5!(y)$);
            \coordinate (smallCircleEdge) at ($(midpoint) - (0,\innerRadius)$);
            \coordinate (bigCircleEdge) at ($(c) + (0,\outerRadius)$);
            \coordinate (smallRadiusMidpoint) at ($(midpoint)!0.5!(smallCircleEdge)$);
            \coordinate (bigRadiusMidpoint) at ($(c)!0.5!(bigCircleEdge)$);
            \def\firstcircle{(0,0) circle (\outerRadius)}
            \def\secondcircle{(midpoint) circle (\innerRadius)}
            \def\firstrectangle{(-\outerRadius,-\outerRadius) rectangle ($(\separatorX*\outerRadius,\outerRadius)$)}
            \def\secondrectangle{($(\separatorX*\outerRadius,-\outerRadius)$) rectangle ($(\outerRadius,\outerRadius)$)}

            \colorlet{circle edge}{black}
            \colorlet{circle area}{yellow!20}
            \colorlet{rectangle edge}{blue!20}
            
            \tikzset{filled/.style={fill=circle area},
            outline/.style={draw=circle edge, thick}}
            \draw[outline, dashed] \secondcircle;
            \fill (midpoint) circle [fill, radius=2pt, anchor=south] node[anchor=south west] {$x$};
            \draw (smallRadiusMidpoint) node[anchor=west] {$\delta$};
            \draw[<->, dotted] ($(midpoint) - (0, 0.1)$) -- (smallCircleEdge);
            \draw[<->, dotted] ($(separatorMidpoint)$) -- ($(midpoint) - (0.05, 0)$);
            \draw (separatorTop) -- (separatorBottom);
            \draw ($(midpoint)!0.5!(separatorMidpoint)$) node[anchor=north] {$\epsilon$};
        \end{tikzpicture}
    \end{center}
    \caption{A data point $x$ and the (locally linear) decision surface of a classifier $f$ (solid line). The point $x$ is susceptible to an adversarial attack of size $\epsilon$, and randomly perturbed using random noise of size $\leq \delta$. These perturbed points are sampled from the within dashed ball.}
    \label{fig:fixedDataPoint}
\end{figure}

\subsection{{Random perturbations are inefficient for detecting adversarial instability}}\label{sec:fixedDataPoint}
We first consider the simple setup illustrated in Figure~\ref{fig:fixedDataPoint}.
Suppose we wish to estimate the susceptibility of a fixed data point $x$ to adversarial attack.
This may be measured as the distance from $x$ to the decision surface of a classifier.
For simplicity, we suppose that this decision surface is locally linear near $x$, and we denote the shortest distance from $x$ to the decision surface by $\epsilon$.
We may therefore say that the point $x$ is susceptible to an adversarial attack of size $\epsilon$, since this is the smallest perturbation which would push $x$ across the decision boundary.
In keeping with the setup of the paradox, we attempt to estimate the (unknown) size of $\epsilon$ by randomly perturbing the data point $x$ using noise of size $\delta$.
We may do this by measuring the proportion of random perturbations which fall on the opposite side of the decision surface.
Unfortunately, as described in Theorem~\ref{thm:fixedDataPoint}, this process is extremely inefficient in high dimensions.

\begin{theorem}[Random perturbations are inefficient for detecting adversarial instability]\label{thm:fixedDataPoint}
    Let $x \in \mathbb{R}^n$ and let $\Pi$ be a planar decision surface passing distance $\epsilon > 0$ from $x$.
    Suppose (without loss of generality since the setup is invariant to rigid translations) that $\Pi$ passes through the origin.
    Suppose that points $z$ are sampled uniformly from a ball of radius $\delta \geq \epsilon$ around $x$.
    Then, the probability of sampling a point $z$ with a different classification from $x$ decreases exponentially with the dimension $n$.
    Specifically, if $\Pi$ has normal vector $\nu$ (with $\|\nu\| = 1$) then
    \begin{equation*}
        P(z \sim \mathcal{U}(\mathbb{B}_{\delta}^{n}(x)) \suchthat \sign(z \cdot \nu) \neq \sign(x \cdot \nu)) \leq \frac{1}{2} \Big(1 - \frac{\epsilon^2}{\delta^2} \Big)^{\frac{n}{2}}.
    \end{equation*}
\end{theorem}

This theorem is proved in Section~\ref{sec:proof:fixedDataPoint}.
The clear implication is that exponentially many perturbed data samples would be required to expect to find any which are misclassified.
This remains true even when the sampled noise is much larger than the size $\epsilon$ of the adversarial attack affecting $x$.
Since this does not depend on any data distribution of $x$, it provides a first hint at the foundations of the paradox.

It is interesting to ask whether this finding is due to the choice of sampling noise uniformly from a ball around $x$.
The answer to this question is `no', due to concentration properties of data distributions in high dimensions.
For example, points sampled from a uniform distribution on the cube $[-1, 1]^d$ and those sampled from a Gaussian distribution (with mean $0$ and unit variance) both concentrate such that $\frac{1}{\sqrt{d}}\|x\|$ is almost constant with high probability in high dimensions~\cite{ledoux2001concentration}.
Data from these distributions therefore behaves very similarly to data sampled uniformly from a ball (albeit a ball with radius growing with $\sqrt{d}$).

\subsection{{A simple theoretical model captures the essence of the paradox}}\label{sec:two-balls-model}

To more completely understand the paradox, in this section we show how it manifests itself in the case where data points sampled from two classes are classified using a linear classifier.
We adopt the simple yet reasonably flexible assumption that each data class is sampled from a distribution supported somewhere within a ball and satisfying a mild non-degeneracy requirement (Definition~\ref{def:smac}).
A significantly 
more general version
of this model is analysed in Section~\ref{sec:general-model}, with fewer constraints on the distributions and a classifier which is permitted to use a more general nonlinear decision surface. 
The results and conclusions remain largely qualitatively similar. In particular, the simultaneous co-existence of high accuracy, the typicality of data susceptible to adversarial attacks, and the rarity of destabilising random perturbations with bounded Euclidean norm all extend to the more general model with nonlinear decision boundary (see Theorems \ref{thm:supplementary:generalisation:accuracy}, \ref{thm:supplementary:generalisation:susceptibility}, \ref{thm:supplementary:generalisation:undetectability} and Corollaries \ref{corr:supplementary:generalisation:accuracy}, \ref{corr:supplementary:generalisation:susceptibility}, \ref{corr:supplementary:generalisation:undetectability}).

Let us now formally define the setup considered in this section. To assess the probabilities and typicality of events we need to define an appropriate class of data distributions. This class should be sufficiently flexible to capture uncertainties and the lack of precise knowledge about data distributions while also tractable enough to enable a mathematical assessment of the setup.
One such class of distributions is those satisfying the Smeared Absolute Continuity property~\cite{gorban2018correction}.
\begin{definition}[Smeared Absolute Continuity (SmAC)~\cite{gorban2018correction}]
    Let $x_1,\dots,x_M\in\mathbb{R}^n$ be random variables. 
    The joint distribution of $x_1,\dots,x_M$ has the SmAC property if there exist constants $\alpha>0$, $\beta\in(0,1)$, and $\gamma>0$ such that for every positive integer $n$, any convex set $S\subset\Real^n$ such that
    \[
    \frac{V^n(S)}{V^n(\mathbb{B}^n)}\leq \alpha^n,
    \]
    any index $i\in\{1,2,\dots,M\}$ and any points $y_1,\dots,y_{i-1}, y_{i+1},\dots,y_M\in\Real^n$, we have
    \[
    P(x_i\in\mathbb{B}^n\setminus S \given x_j=y_j, \ \forall \ j\neq i)\geq 1 - \gamma \beta^n.
    \]
\end{definition}

In this work, however, we do not wish to consider a joint distribution over multiple random variables as our main case focuses on a single point drawn from a distribution.
Furthermore, to make the technical analysis simpler and clearer it is beneficial to assume the existence of a probability density function that is associated with the probability measure.
Therefore, in what follows we adopt the following restricted single-particle version of the SmAC property which, for the sake of brevity, will be referred to as SmAC in the rest of the paper:

\begin{definition}[Single-particle SmAC with bounded density]\label{def:smac}
    A distribution $\mathcal{D}$ on $\mathbb{R}^n$ is said to satisfy the \emph{single-particle smeared absolute continuity condition with bounded density} if it possesses a density $p : \mathbb{R}^n \to \mathbb{R}_{\geq 0}$ and there exists a centre point $c \in \mathbb{R}^n$ and radius $r > 0$ such that $p(x) > 0$ only for points $x$ in the ball $\mathbb{B}^n_{r}(c)$, and there exists a constant growth parameter $A > 0$ such that
    \begin{align*}
        \sup_{x \in \mathbb{B}^n_r(c)} p(x) \leq \frac{A}{V^n r^n}.
    \end{align*}
\end{definition}

We note that if the growth property is satisfied with $A = 1$, then the distribution is simply the uniform distribution on the ball $\mathbb{B}^n_r(c)$.

Suppose that two classes of data are each sampled from data distributions $\mathcal{D}_0$ and $\mathcal{D}_1$ on $\mathbb{R}^n$, each satisfying Definition~\ref{def:smac}.
For simplicity, we suppose that these distributions are each supported in a ball with radius 1, with centres given by $c_0 = -\epsilon \mathbf{e}_1$ for class 0 and $c_1 = \epsilon \mathbf{e}_1$ for class 1.
We further suppose that both distributions satisfy the growth bound with the same parameter $A$.
For brevity, we also define the combined distribution $\mathcal{D}_{\epsilon}$ which samples a point from $\mathcal{D}_0$ with label 0 with probability $\frac{1}{2}$, and samples a point from $\mathcal{D}_1$ with label 1 with probability $\frac{1}{2}$.
The geometry of this setup is illustrated in Figure~\ref{fig:closeSpheres}.

\begin{figure}
    \begin{center}
        \begin{tikzpicture}[scale=0.75]
            \def\outerRadius{2cm}
            \def\innerRadius{2cm}
            \def\separatorX{0.85}
            \coordinate (y) at (3, 0);
            \coordinate (c) at (0, 0);
            \coordinate (separatorTop) at (0.75, 2.2);
            \coordinate (separatorBottom) at (0.75, -2.2);
            \coordinate (separatorMidpoint) at ($(separatorTop)!0.5!(separatorBottom)$);
            \coordinate (midpoint) at ($(c)!0.5!(y)$);
            \coordinate (smallCircleEdge) at ($(midpoint) - (0,\innerRadius)$);
            \coordinate (bigCircleEdge) at ($(c) + (0,\outerRadius)$);
            \coordinate (smallRadiusMidpoint) at ($(midpoint)!0.5!(smallCircleEdge)$);
            \coordinate (bigRadiusMidpoint) at ($(c)!0.5!(bigCircleEdge)$);
            \def\firstcircle{(0,0) circle (\outerRadius)}
            \def\secondcircle{(midpoint) circle (\innerRadius)}
            \def\firstrectangle{(-\outerRadius,-\outerRadius) rectangle ($(\separatorX*\outerRadius,\outerRadius)$)}
            \def\secondrectangle{($(\separatorX*\outerRadius,-\outerRadius)$) rectangle ($(\outerRadius,\outerRadius)$)}

            \colorlet{circle edge}{black}
            \colorlet{circle area}{yellow!20}
            \colorlet{rectangle edge}{blue!20}
            
            \tikzset{filled/.style={fill=circle area},
            outline/.style={draw=circle edge, thick}}
            \draw[outline] \firstcircle;
            \draw[outline] \secondcircle;
            \fill (c) circle [fill, radius=2pt, anchor=west];
            \fill (midpoint) circle [fill, radius=2pt, anchor=south];
            \draw (smallRadiusMidpoint) node[anchor=east] {$1$};
            \draw (bigRadiusMidpoint) node[anchor=north west] {$1$};
            \draw[<->, dotted] ($(midpoint) - (0, 0.1)$) -- (smallCircleEdge);
            \draw[<->, dotted] ($(c) + (0, 0.1)$) -- (bigCircleEdge);
            \draw[<->, dotted] ($(c) + (0.05, 0)$) -- ($(separatorMidpoint)$);
            \draw[<->, dotted] ($(separatorMidpoint)$) -- ($(midpoint) - (0.05, 0)$);
            \draw[dashed] (separatorTop) -- (separatorBottom);
            \draw ($(midpoint)!0.5!(separatorMidpoint)$) node[anchor=north] {$\epsilon$};
            \draw ($(c)!0.5!(separatorMidpoint)$) node[anchor=north] {$\epsilon$};
        \end{tikzpicture}
    \end{center}
    \caption{Two unit balls with centres separated by distance $2\epsilon$, and the decision surface of the classifier $f$ (dashed).}
    \label{fig:closeSpheres}
\end{figure}

The classification function $f : \mathbb{R}^n \to \{0, 1\}$ with the highest accuracy which can be defined for this data model without further knowledge of the distributions is given by the simple linear separator
\begin{align}\label{eq:classifier}
    f(x) = 
    \begin{cases}
        0 &\text{ if } x_1 < 0,
        \\
        1 &\text{ otherwise}.
    \end{cases}
\end{align}
This classifier does not necessarily return the correct label in all cases since, for $\epsilon \in (0, 1)$, the two data classes overlap.
Despite this, misclassified points are rare in the high dimensional setting, even when the two balls from which points are sampled have only a small separation between their centres.
More precisely, the probability that this classifier is correct converges exponentially to 1 as the data dimension grows.
This result is proven in Section~\ref{sec:accuracyProof}.

\begin{theorem}[The classifier is accurate]\label{thm:accuracy}
    For any $\epsilon > 0$, the probability that the classifier applies the correct label to a randomly sampled data point grows exponentially to 1 with dimension $n$, specifically
    \begin{align*}
        P((x, \ell) \sim \mathcal{D}_\epsilon : f(x) = \ell)
        \geq
        1 - \frac{1}{2} A (1 - \epsilon^2)^{\frac{n}{2}}.
    \end{align*}
\end{theorem}

\begin{figure*}
    \centering
    \begin{subfigure}{0.49\linewidth}
        \centering
        \includegraphics[width=0.6\linewidth]{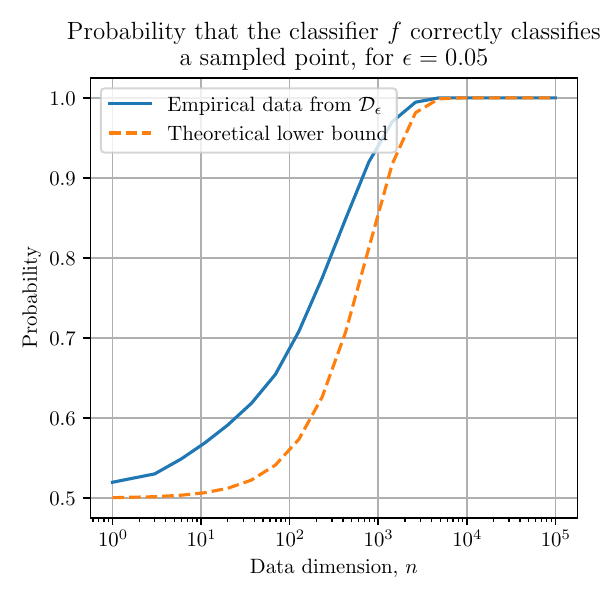}
        \caption{Accuracy of the classifier (Theorem~\ref{thm:accuracy})}
        \label{fig:empiricalAccuracy}
    \end{subfigure}%
    \begin{subfigure}{0.5\linewidth}
        \centering
        \includegraphics[width=\linewidth]{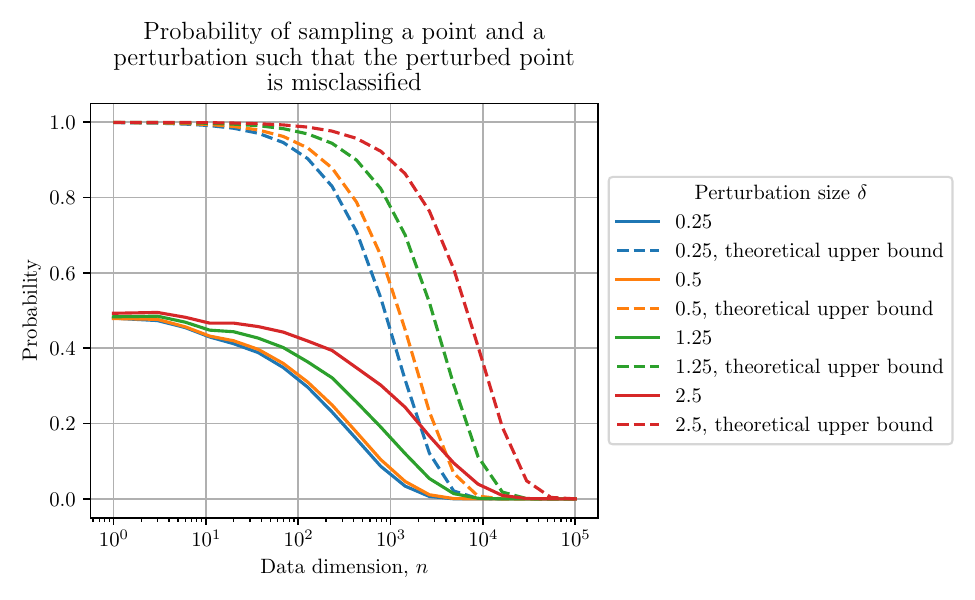}
        \caption{Destabilisation probability (Theorem~\ref{thm:undetectability})}
        \label{fig:perturbationProbability}
    \end{subfigure}
    \caption{Comparison of the theoretical bounds in Theorems~\ref{thm:accuracy} and~\ref{thm:undetectability} against empirical results computed using 10,000 data points sampled from $\mathcal{D}_{\epsilon}$, with $\epsilon = 0.05$, and 10,000 perturbations sampled from $\mathcal{U}(\mathbb{B}^n_{\delta})$ for various values of $\delta$. We see that, even for perturbations 50 times larger than the separation distance between the balls (i.e. $\delta = 2.5$), the probability of randomly sampling a perturbation which changes the classification of a random data point is very small in high dimensions.}
\end{figure*}

The sharpness of this result is verified empirically in Figure~\ref{fig:empiricalAccuracy}, computed for $A = 1$.
We observe that by $n = 10,000$, the probability of sampling a point which will be misclassified is virtually 0.
To put this and the following results into context, the $32\times32\times3$ images used in CIFAR-10 have 3,072 attributes, while the size of $256\times256\times3$ commonly used for the images in ImageNet have 196,608 attributes, placing them firmly within the range of dimensionalities where the effects described here are active.

On the other hand, even accurately classified points in this model are still close to the decision surface since the ball centres are only separated by distance $\epsilon$.
Because of this, for any $\delta > \epsilon$, there are points sampled from each class which are susceptible to an adversarial attack $s \in \mathbb{R}^n$ with $\|s\| \leq \delta$ which causes $f$ to predict the wrong class.
Moreover, in high dimensions, data points sampled from such a distribution concentrate at distance $\epsilon$ from this decision surface, meaning that the probability of sampling a point which is susceptible to an adversarial attack is high.
This may be encapsulated in the following result, which is proven 
in Section~\ref{sec:susecptibilityProof}.

\begin{theorem}[Susceptible data points are typical]\label{thm:susceptibility}
    For any $\epsilon \geq 0$ and $\delta \in [\epsilon, 1 + \epsilon]$,  the probability that a randomly sampled data point is susceptible to an adversarial attack with Euclidean norm $\delta$ grows exponentially to 1 with the dimension $n$, specifically
    \begin{align*}
        P\big((x, \ell) &\sim \mathcal{D}_\epsilon \suchthat \text{there exists } s \in \mathbb{B}^n_{\delta} 
        \text{ such that } f(x + s) \neq \ell \big)
        \\&
        \geq
        1 - \frac{1}{2} A (1 - (\delta - \epsilon)^2)^{\frac{n}{2}}.
    \end{align*}
\end{theorem}

Although this susceptibility may therefore be viewed as typical in high dimensions, however, the probability of detecting it by sampling random perturbations of data points is paradoxically very small, as shown by the following result which 
is proven in Section~\ref{sec:undetectabilityProof}.

\begin{theorem}[Destabilising perturbations are rare]\label{thm:undetectability}
    For any $\delta > \epsilon \geq 0$, the probability that a randomly selected perturbation with Euclidean norm $\delta$ causes a randomly sampled data point to be misclassified is bounded from above as:
    \begin{align*}
        P\big( (x, \ell) \sim \mathcal{D}_\epsilon, s \sim \mathcal{U}(\mathbb{B}^n_\delta) &\suchthat f(x + s) \neq \ell \big)
        \\&
        \leq
        A
        \Big(1 - \Big(\frac{\epsilon}{1 + \delta} \Big)^2 \Big)^{\frac{n}{2}}.
    \end{align*}
In particular, when $\delta$ is independent of dimension $n$, this probability converges to 0 exponentially with  $n$.
\end{theorem}

This probability bound is compared against empirically sampled data in Figure~\ref{fig:perturbationProbability}.
While the bound is not particularly sharp in low dimensions, it accurately describes the key phenomenon which is the convergence of the probability to 0 in high dimensions.
This phenomenon is startlingly persistent, even when the magnitude of the sampled perturbations is 50 times greater than the distance between the centres of the spheres (when $\delta = 2.5$).

We note that some care needs to be taken when considering perturbations with fixed $\ell^\infty$ norms.
The corresponding $\ell^2$ norm of these perturbations scales as $\sqrt{n}$, affecting convergence to $0$ of the probability of destabilisation (see Theorem \ref{thm:undetectability}).

Even though randomly sampled perturbations are unlikely to affect the classifier, it is often straightforward to construct special adversarial perturbations which will affect a specific data point.
Common algorithms for constructing adversarial attacks work by perturbing the target input in such a way as to increase an appropriate loss function.
Gradient-based methods for this, such as the Fast Gradient Sign Method~\cite{goodfellow2014explaining}, compute the gradient of the loss function with respect to the components of the input, evaluated at the target input with its true class.
Perturbing the input in the direction of this gradient therefore moves it in the direction of steepest ascent of the loss function locally, thereby representing a good candidate for an adversarial direction.
The minimal scaling to be applied to this adversarial direction, required to form the final adversarial input, can then be determined via a line search in the adversarial direction.

In the case of this model setup, such an algorithm (with a standard choice of loss function) will successfully provide the optimal direction for an adversarial attack: the most direct path to move the input along in order to cross the decision surface.
To show this, we first observe that the classifier $f$ in~\eqref{eq:classifier} can be equivalently defined as $f(x) = H(g(x))$, where $H : \mathbb{R} \to \{0, 1\}$ denotes the (piecewise constant) Heaviside function, and the linear function $g(x) = \mathbf{e}_1 \cdot x -\frac{1}{2}$.
To construct gradient-based attacks, we use a differentiable version $\tilde{f}$ of $f$ constructed as $\tilde{f}(x) = \sigma(g(x))$, where $\sigma : \mathbb{R} \to (0, 1)$ is a continuously differentiable version of the Heaviside function which is monotonically increasing with $\sigma(0) = \frac{1}{2}$.
An example of such a function is the standard sigmoid function.
Then, the following result, proved in Section~\ref{sec:gradientAttackProof}, shows that gradient-based attacks on this classifier will always return the optimal attack direction.

\begin{theorem}[Gradient-based methods find the optimal adversarial attack]\label{thm:gradientBasedAttack}
    Let $L : \mathbb{R}_{> 0} \to \mathbb{R}$ denote any differentiable, monotonically increasing loss function, and let $(x, \ell) \sim \mathcal{D}_{\epsilon}$.
    Then, with probability 1 with respect to the sample $(x, \ell)$, the gradient of the loss $L(|\tilde{f}(x) - \ell|)$ with respect to the components of $x$ corresponds to a positive multiple of the optimal attack direction $(1 - 2\ell) \mathbf{e}_1$.
\end{theorem}

A further aspect of this model problem is that successful adversarial attacks are universal in high dimensions.
To state this property mathematically, we define the \emph{destabilisation margin} to be the distance by which a destabilising perturbation pushes a data point across the decision threshold of the classifier~\eqref{eq:classifier}.
This is measured by the functions $d_\ell : \mathbb{R}^n \times \mathbb{R}^n \to \mathbb{R}$ associated with each class $\ell = 0, 1$, where, for a data point $x$ and a perturbation $s$,
\begin{align*}
    d_0(x, s) = \max\{x_1 + s_1, 0\}, 
\end{align*}
and
\begin{align*}
    d_1(x, s) = \max\{- x_1 - s_1, 0\}.
\end{align*}
The following result then holds, as proven in Section~\ref{sec:universalityProof}.

\begin{theorem}[Universality of adversarial attacks]\label{thm:universality}
    Let $\epsilon \geq 0$ and suppose that $x, z \sim \mathcal{D}_{\epsilon}$ are independently sampled points with the same class label $\ell$.
    For any $\gamma \in (0, 1]$, the probability that $x$ is destabilised by all perturbations $s \in \mathbb{R}^n$ which destabilise $z$ with destabilisation margin $d_{\ell}(z, s) > \gamma$ converges exponentially to 1 as the dimension $n$ increases.
    Specifically, for $\ell \in \{0, 1\}$ and $z \in \Real^n$, let $S_z = \{s \in \Real^n \suchthat d_{\ell}(z, s) > \gamma \}$.
    Then,
    \begin{align*}
        P( x, z &\sim \mathcal{D}_\ell \suchthat f(x + s) \neq \ell \text{ for all } s \in S_z)
        \\&
        \geq
        \Big(1 - A  \Big(1 - \frac{\gamma^2}{4} \Big)^{\frac{n}{2}} \Big)^2
    \end{align*}
\end{theorem}

This bound shows that in high dimensions we may expect pairs of sampled points to share their sets of adversarial perturbations.
The dependence on the margin $\gamma$ by which the perturbation destabilises $z$ is an interesting feature.
Roughly speaking, the result suggests that in low dimensions only severe perturbations which push points a long way past the decision threshold may be regarded as universal in the sense of having a high probability of destabilising other sampled points.
As the dimension $n$ increases, however, perturbations which produce smaller and smaller margins on individual points become universal in the sense that they have a constant probability of destabilising other sampled points.

\subsection{A generalised theoretical model}\label{sec:general-model}
We now show that the simple case presented in Section~\ref{sec:two-balls-model} extends to more general cases in which the classification surface is not assumed to be flat, and the data are sampled from more general distributions.
To demonstrate that these abstract results are true generalisations of the results proven in Section~\ref{sec:two-balls-model}, we derive corollaries to each result for a general SmAC distribution with a flat decision surface.
These corollaries are therefore directly comparable with the results in Section~\ref{sec:two-balls-model} for specific indicated values of the parameters.

Let $\planenormal, w \in \mathbb{R}^n$ with $\|\planenormal\| = 1$, and define the plane
\begin{align*}
    \plane = \{ x \in \mathbb{R}^n \suchthat (x - w) \cdot \planenormal = 0 \} \subset \mathbb{R}^n,
\end{align*}
which passes through $w$ and is normal to the vector $\planenormal$.
Denote by $\Pi : \mathbb{R}^n \to \plane$ the orthogonal projection operator onto $\plane$ in the Euclidean inner product, given by
\begin{align*}
    \Pi x = x - ((x - w) \cdot \planenormal) \planenormal.
\end{align*}
Let $\sfunc : \plane \to \mathbb{R}$ be continuous, and define the surface 
\begin{align*}
    S = \{x \in \mathbb{R}^n \suchthat x - \sfunc(\proj x) \planenormal \in \plane \} \subset \mathbb{R}^n.
\end{align*}
A projector $\sproj : \mathbb{R}^n \to S$ onto the surface $S$ (along the vector $\planenormal$) can be defined by
\begin{align*}
    \sproj x = \proj x + \sfunc (\proj x) \planenormal.
\end{align*}

We also introduce the signed directed distance function $d_\plane : \mathbb{R}^n \to \mathbb{R}$ measuring the signed distance from a point $x$ to the plane $\plane$ along the normal vector $\planenormal$, given by
\begin{align*}
    d_{\plane}(x) = (x - \proj x) \cdot \planenormal = (x - w) \cdot \planenormal,
\end{align*}
and $d_S : \mathbb{R}^n \to \mathbb{R}$ measuring the signed distance from a point $x$ to the surface $S$ along the vector $\planenormal$, given by
\begin{align*}
    d_S(x) &= (x - \sproj(x)) \cdot \planenormal = (x - \proj x) \cdot \planenormal - \sfunc(\proj x) 
    \\&
    = d_{\plane}(x) - \sfunc(\proj x).
\end{align*}
Finally, we can define the distance from a point $x$ to the surface $S$ by
\begin{align*}
    \sigma(x) = \inf_{\hat{y} \in S} \|x - \hat{y}\|,
\end{align*}
noting the trivial inequality
\begin{align}\label{eq:generalisation:sBound}
    \sigma(x) \leq |d_S(x)|,
\end{align}
for any $x \in \mathbb{R}^n$, since $d_S$ only measures distance to $S$ in the direction of $\planenormal$ while $\sigma$ measures the shortest distance to $S$ in any direction.

With these constructions, we can define a binary classifier with decision surface $S$ as the function $f : \mathbb{R}^n \to \{0, 1\}$ given by
\begin{align}\label{eq:generalisation:classifier}
    f(x) =
    \begin{cases}
        0 &\text{ if } d_S(x) \leq 0,
        \\
        1 &\text{ otherwise}.
    \end{cases}
\end{align}

To show how our previous results extend into this more general case, suppose that data points of class $0$ are sampled from a distribution $\distribution$ on $\mathbb{R}^n$, and that data points of class $1$ are sampled from the distribution $\mathcal{D}^{\prime}$ on $\mathbb{R}^n$.
In the interests of simplicity, we only study the behaviour of the classifier for data from the class $0$, as the result for the class $1$ is analogous.
We study this more general model in parallel with the results of Section~\ref{sec:two-balls-model}.

We first observe that the accuracy of the classifier may be controlled in an analogous way to the simple case in Section~\ref{sec:two-balls-model}.
The supremum in this result (and the suprema and infima in subsequent results) is simply present to ensure an optimal balancing for the two terms; a valid (though possibly sub-optimal) result may be obtained by selecting any value of $\alpha \geq 0$.

\begin{theorem}[Accuracy of the classifier $f$]\label{thm:supplementary:generalisation:accuracy}
    Let $x \sim \mathcal{D}$. Then, the probability that $x$ is correctly classified as class $0$ by the classifier $f$ is at least
    \begin{align*}
        &P(x \sim \distribution \suchthat f(x) = 0)
        \\&\qquad
        \geq 
        \sup_{\alpha \geq 0}
        \big[
            P(x \sim \distribution \suchthat |\sfunc(\proj x)| \leq \alpha)
            \\&\qquad\qquad\qquad
            -
            P(x \sim \distribution \suchthat d_{\plane}(x) > -\alpha)
        \big].
    \end{align*}
\end{theorem}

The proof of this result is given in Section~\ref{sec:supplementary:generalisation:accuracyProof}.
The first term appearing on the right hand side controls how far the surface $S$ may be expected to deviate from the plane $\plane$ (and is therefore simply 1 in the case when $\sfunc \equiv 0$ and so $S = \plane$; in this case the optimal balancing of the terms will be obtained when $\alpha = 0$).
The second term, on the other hand, estimates the probability that a point is correctly classified by the plane placed parallel to $\plane$, but offset by distance $\alpha$ to account for the variability of $\sfunc$.

We demonstrate this result in the setting of a linear classifier with a distribution $\mathcal{E}$ which satisfies the SmAC condition of Definition~\ref{def:smac} with radius $r > 0$ and centre $c$ such that $d_{\plane}(c) = - \eta$ for some $\eta \in [0, r)$. 
Then, Theorem~\ref{thm:supplementary:generalisation:accuracy} takes the following form, from which we obtain Theorem~\ref{thm:accuracy} when $r = 1$ and $\eta = \epsilon$.

\begin{corollary}[Accuracy for SmAC distributions]\label{corr:supplementary:generalisation:accuracy}
    Suppose that points with label 0 are sampled from the distribution $\mathcal{E}$, and suppose that $\sfunc \equiv 0$.
    Then, for $x \sim \mathcal{E}$, the probability that the classifier $f$ correctly assigns $x$ class $0$ is at least
    \begin{align*}
        P(x \sim \mathcal{E} \suchthat f(x) = 0)
        \geq 
        1
        -
        \frac{1}{2} A \Big(1 - \Big(\frac{\eta}{r}\Big)^2\Big)^{\frac{n}{2}}.
    \end{align*}
\end{corollary}

We may also prove a generalised version of the susceptibility result of Theorem~\ref{thm:susceptibility} in our abstract setting in Section~\ref{sec:supplementary:generalisation:susceptibilityProof}.
The probability of sampling a data point which is susceptible to an adversarial attack of size $\delta$ may be bounded from below as in the following result.
The form of this result is similar to that of Theorem~\ref{thm:supplementary:generalisation:accuracy}, although we note the crucial difference in the second term.

\begin{theorem}[Susceptibility to adversarial perturbations]\label{thm:supplementary:generalisation:susceptibility}
    Suppose that points with label 0 are sampled from the distribution $\mathcal{D}$.
    Then, for any $\delta > 0$, the probability that a point sampled at random from the class $0$ is susceptible to an adversarial attack with Euclidean norm $\delta$ is at least
    \begin{align*}
        &P(x \sim \distribution \suchthat \text{ there exists } s \in \mathbb{B}^n_\delta \text{ with } f(x + s) \neq 0)
        \\&\qquad
        \geq
        \sup_{\alpha \geq 0}
        \big[
            P(x \sim \distribution \suchthat |\sfunc(\proj x)| \leq \alpha) 
            \\&\qquad\qquad\qquad
            - 
            P(x \sim \distribution \suchthat d_{\pi}(x) \leq \alpha - \delta)
        \big].
    \end{align*}
\end{theorem}

When applied to the SmAC distribution $\mathcal{E}$, this result takes the following form, from which we obtain Theorem~\ref{thm:susceptibility}.

\begin{corollary}[Susceptibility for SmAC distributions]\label{corr:supplementary:generalisation:susceptibility}
    Suppose that points with label 0 are sampled from the distribution $\mathcal{E}$, and suppose that $\sfunc \equiv 0$.
    Then, for any $\delta \in [\eta, r]$, the probability that a point sampled at random from the class $0$ is susceptible to an adversarial attack with Euclidean norm $\delta$ is at least
    \begin{align*}
        &P(x \sim \mathcal{E} \suchthat \text{ there exists } s \in \mathbb{B}^n_\delta \text{ with } f(x + s) \neq 0)
        \\&\qquad
        \geq
        1 
        - 
        \frac{1}{2} A \Big(1 - \Big(\frac{\delta - \eta}{r}\Big)^2\Big)^{\frac{n}{2}}.
    \end{align*}
\end{corollary}

We next derive a generalised version of Theorem~\ref{thm:undetectability}, which bounds the probability of sampling a random perturbation which is adversarial for $f$.
For this result, we assume that the surface $S$ has some regularity, in the sense that the function $\sfunc$ is Lipschitz with constant $L \geq 0$; i.e. for any $\hat{x}, \hat{y} \in \plane$ we have $|\sfunc(\hat{x}) - \sfunc(\hat{y})| \leq L \| \hat{x} - \hat{y} \|$.
Geometrically, for any $x \in \Real^n$ this defines a cone of points containing $x$ in which $f$ is guaranteed to be constant.
This property allows us to prove the following lower bound on $\sigma$ by $d_S$ in Section~\ref{sec:supplementary:generalisation:undetectabilityProof}, which may be viewed as a companion to~\eqref{eq:generalisation:sBound}

\begin{lemma}[Lipschitz regularity gives control of $\sigma$]\label{lem:supplementary:generalisation:sLowerBound}
    Suppose that $\sfunc$ is Lipschitz with parameter $L$. Then, for any $x \in \mathbb{R}^n$,
    \begin{align}\label{eq:supplementary:generalisation:sLowerBound}
        \sigma(x) \geq |d_S(x)| \sin \theta,
    \end{align}
    where $\theta = \arctan(L^{-1})$.
\end{lemma}

This crucial property allows us to prove the following generalisation of Theorem~\ref{thm:undetectability} in Section~\ref{sec:supplementary:generalisation:undetectabilityProof}, indicating that destabilising random perturbations may be expected to be rare.

\begin{theorem}[Probability of sampling misclassifying random perturbations]\label{thm:supplementary:generalisation:undetectability}
    Suppose that points with label 0 are sampled from the distribution $\mathcal{D}$, and suppose that $\sfunc$ is Lipschitz with parameter $L$.
    Then, for any $\delta > 0$, the probability that a point sampled at random from the class $0$ will be misclassified after the application of a perturbation randomly sampled uniformly from $\mathbb{B}^n_{\delta}$ is bounded by
    \begin{align*}
        &P(x \sim \distribution, s \sim \mathbb{B}^n_{\delta} \suchthat f(x + s) \neq 0)
        \\&\,
        \leq
        \inf_{\substack{\alpha, \gamma \geq 0 \\ t \in T(L)}}
        \Big[
            P( x \sim \mathcal{D} \suchthat |\sfunc(\proj x)| \geq \alpha)
            \\&\qquad\qquad\quad
            + 
            P\Big( x \sim \mathcal{D} \suchthat d_{\plane}(x) \geq - \alpha - \frac{t}{\sin\theta} \Big)
            \\&\qquad\qquad\quad
            +
            \Delta(L)
            \frac{1}{2} \Big(1 - \Big( \frac{t}{\delta} - L \Big)^2 \Big)^{\frac{n}{2}} \cdot
            \\&\qquad\qquad\qquad\quad
            \cdot
            \big(
                P(x \sim \mathcal{D} \suchthat d_{\plane}(x) \leq \gamma - t)
                \\&\qquad\qquad\qquad\qquad\quad
                +
                P(x \sim \mathcal{D} \suchthat |\sfunc(\proj x)| > \gamma) 
            \big)
        \Big],
    \end{align*}
    where $\Delta(L) = 1$ for $L \leq 1$ and 0 for $L > 1$, and the set $T(L) = [\min\{L, 1\}\delta, \delta]$.
\end{theorem}

For the SmAC distribution $\mathcal{E}$, Theorem~\ref{thm:supplementary:generalisation:undetectability} produces the following corollary (proved in Section~\ref{sec:supplementary:generalisation:undetectabilityProof}) from which Theorem~\ref{thm:undetectability} follows when $r = 1$ and $\eta = \epsilon$.
In this case, we have $L = 0$ and so $\theta = \frac{\pi}{2}$ and $\sin\theta = 1$.

\begin{corollary}[Destabilising random perturbations are rare for SmAC distributions]\label{corr:supplementary:generalisation:undetectability}
    Suppose that points with label 0 are sampled from the distribution $\mathcal{E}$, and suppose that $\sfunc \equiv 0$.
    Then, for any $\delta \in [\eta, r]$, the probability that a point sampled a random from the class $0$ is misclassified after the application of a perturbation sampled uniformly from the ball $\mathbb{B}^n_\delta$ is bounded by
    \begin{align*}
        &P(x \sim \mathcal{E}, s \sim \mathbb{B}^n_{\delta} \suchthat f(x + s) \neq 0)
        \leq
        A \Big(1 - \Big(\frac{\eta}{r + \delta}\Big)^2\Big)^{\frac{n}{2}}.
    \end{align*}
\end{corollary}

Finally, we also obtain a generalised analogue of the universality result of Theorem~\ref{thm:universality}.
We define the notion of the \emph{destabilisation margin} in this setting to be the distance by which a perturbation pushes a data point across the decision threshold of the classifier~\eqref{eq:generalisation:classifier}.
This is measured for class 0 by the function $d_0 : \mathbb{R}^n \times \mathbb{R}^n \to \mathbb{R}$, where, for a data point $x$ and a perturbation $s$,
\begin{align*}
    d_0(x, s) = \max\{d_S(x + s), 0\}.
\end{align*}
The following result then holds, as proven in Section~\ref{sec:supplementary:generalisation:universalityProof}.

\begin{theorem}[Universality of adversarial attacks]\label{thm:supplementary:generalisation:universality}
    Suppose that $x, z \sim \mathcal{D}$ are independently sampled points with label 0, and suppose that $\sfunc$ is Lipschitz with parameter $L$.
    For any $\delta, \gamma \in \Real$, the probability that $x$ is destabilised by all perturbations $s \in \mathbb{B}^n_\delta$ which destabilise $z$ with destabilisation margin $d_{0}(z, s) > \gamma$ is bounded from below by
    \begin{align*}
        &P( x, z \sim \mathcal{D} \suchthat f(x + s) \neq 0 \text{ for all } s \in S_z(\delta))
        \\&
        \geq
        \sup_{\alpha \geq 0, t \in \Real}
        \Big[
            \Big( 
                P(z \sim \mathcal{D} \suchthat |\phi(\Pi z)| \leq \alpha)
                \\&\qquad\qquad\quad
                - P(z \sim \mathcal{D} \suchthat d_\pi(z) > t + \chi)
            \Big)
            \cdot
            \\&\qquad\qquad
            \cdot
            \Big( 
                P(x \sim \mathcal{D} \suchthat |\phi(\Pi x)| \leq \alpha)
                \\&\qquad\qquad\quad
                - P(x \sim \mathcal{D} \suchthat d_\pi(x) \leq t - \chi)
            \Big)
        \Big],
    \end{align*}
    where $\chi = \frac{1}{2}\gamma  - L\delta - \alpha$, and for $z \in \Real^n$ and $\delta \in \Real$, we define $S_z(\delta) = \{s \in \mathbb{B}^n_\delta \suchthat d_{0}(z, s) > \gamma \}$.
\end{theorem}

For the SmAC distribution $\mathcal{E}$, this result takes the form shown in Corollary~\ref{corr:supplementary:generalisation:universality}. Theorem~\ref{thm:universality} follows from this result in the case when $r = 1$ and $\eta = \epsilon$.
Interestingly, this result does not depend on the perturbation size $\delta$, due to the fact that the decision surface is assumed to be flat.

\begin{corollary}[Universality of adversarial perturbations for SmAC distributions]\label{corr:supplementary:generalisation:universality}
    Suppose that points with label 0 are sampled from the distribution $\mathcal{E}$, and suppose that $\sfunc \equiv 0$.
    For any $\gamma \in \Real$, the probability that $x$ is destabilised by all perturbations $s \in \mathbb{B}^n$ which destabilise $z$ with destabilisation margin $d_{0}(z, s) > \gamma$ is bounded from below by
    \begin{align*}
        &P( x, z \sim \mathcal{E} \suchthat f(x + s) \neq 0 \text{ for all } s \in S_z)
        \\&\qquad
        \geq
        \Big(
            1 - A \Big( 1 - \frac{\gamma^2}{4r^2} \Big)^{\frac{n}{2}}
        \Big)^2
    \end{align*}
\end{corollary}

\begin{figure*}[h!]
    \centering
    \includegraphics[width=0.8\textwidth] {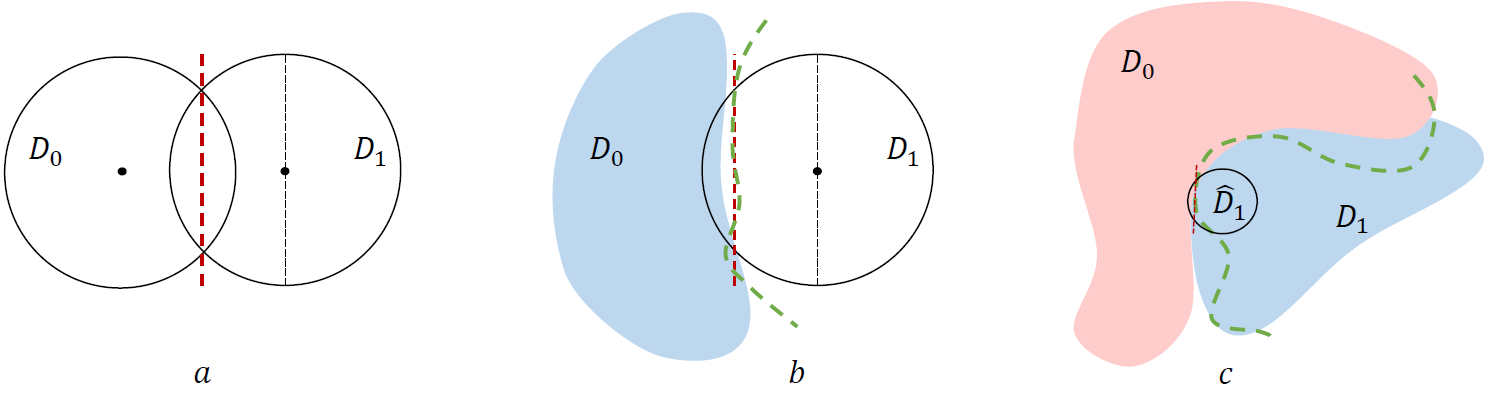}
    \caption{Different scenarios to which the simple two ball model may be generalised.}
    \label{fig:problemExplained}
\end{figure*}

\subsection{Further generalisations}
\label{sec:commentsOnGeneralisbility}

Despite their simplicity, the models presented above cover a wide variety of settings.
The results in Section~\ref{sec:two-balls-model} include data sampled from many common distributions such as uniform distributions and truncated Gaussian distributions.
This setup is depicted in Figure~\ref{fig:problemExplained}a.
The results may be naturally extended to a case where only one of the data classes is sampled from a distribution satisfying the SmAC property, or where the classifier's decision surface is only locally linear (such as ReLU networks), as illustrated in Figure~\ref{fig:problemExplained}b.
Furthermore, the results may be applied in a fully local sense, to locally SmAC distributions, and locally linear classifiers.
This generalisation is shown in Figure~\ref{fig:problemExplained}c.

The generalised setup introduced in Section~\ref{sec:general-model} already incorporates general data distributions, and only assumes that the classifier's decision boundary is a Lipschitz warping of a plane in its normal direction.
This setup can also be directly extended to incorporate other `wiggly' decision surfaces.
For instance, if $S$ cannot be expressed as a modification of a hyperplane in its normal direction, one could instead consider the surfaces defined by the upper and lower graphs of $S$ with respect to $\plane$.
For example, if $S$ is given by a multi-valued function, one could instead just take the maximum or minimum values, and where necessary work with a Lipschitz extension of these surfaces.
Our results extend to this case, albeit with some additional looseness reflecting the `uncertainty' this imposes on the location of the decision surface of the classifier.

Our results also naturally extend to standard multiclass classification problems.
In this case, the decision boundary separating any pair of classes may be viewed locally as a binary classifier, and our results therefore apply locally (as in Figure~\ref{fig:problemExplained}c, for example).
In regions of data space where a small number of classes meet (relative to the data dimension), analogous versions of the results will hold.
This is because in these regions we can apply our result to the boundary between the sampled data point and each other class separately, and collect them together.
The exponential nature of our bounds in the data dimension will therefore dwarf the additional looseness introduced by considering the class boundaries separately.
To treat the situation when the number of classes meeting near a sampled data point is large relative to the data dimension, additional theoretical developments would be required.
However, standard geometric arguments would suggest that these regions of data space would have only a very small measure, implying that data points from non-degenerate distributions are unlikely to be sampled from them.

We do not attempt to treat all these generalised scenarios here, in order to present the main ideas in a simple framework.

\section{Class separation margins hide adversarial susceptibility}\label{sec:hemispheres_model}

A further intriguing component of the paradox of apparent stability is that it may no longer occur when the two data classes are separable, but have no margin separating them\footnote{We note that the two balls model of Section~\ref{sec:two-balls-model} is unable to capture this scenario  since when $\epsilon = 0$ the two balls overlap and the classifier is just 50\% accurate}.
To model this situation, we introduce the \emph{two half-balls model}.
The model comprises two data classes, with binary labels $\{0, 1\}$, each sampled uniformly from a half-ball in dimension $n > 0$.
These half-balls have their flat face parallel to each other and are separated by distance $2\epsilon \geq 0$.
Data of class $0$ are sampled uniformly from the half-ball $D_0 = \{ x \in \mathbb{R}^n \suchthat x + \epsilon \, \mathbf{e}_1 \in H^{-}\mathbb{B}^n \}$, where $\mathbf{e}_1 = (1, 0, \dots, 0)^{\top} \in \mathbb{R}$, while data from class $1$ are sampled uniformly from $D_1 = \{ x \in \mathbb{R}^n \suchthat x - \epsilon \, \mathbf{e}_1 \in H^{+}\mathbb{B}^n \}$.
Here, we use the notation $H^{-}\mathbb{B}^n = \{x \in \mathbb{B}^n \suchthat x \cdot \mathbf{e}_1 < 0 \}$ and $H^{+}\mathbb{B}^n = \{x \in \mathbb{B}^n \suchthat x \cdot \mathbf{e}_1 > 0 \}$.
Any pair of data points $x, y$ sampled with opposite classes therefore satisfy $\|x - y\| \geq 2 \epsilon$.
We denote the combined distribution by $\mathcal{D}_\epsilon = \mathcal{U}(D_0 \cup D_1)$.

A classification function which correctly labels this data for any $\epsilon \geq 0$ can be defined by
\begin{equation}\label{eq:hemisphere:classifier}
    f(x) = 
    \begin{cases}
        0 &\text{ if } x_1 < 0,
        \\
        1 &\text{ otherwise}.
    \end{cases}
\end{equation}
Data points sampled from either class are separated from the decision surface of this classifier by distance at least $\epsilon$.
On the other hand, for any $\delta > \epsilon$, there are clearly points sampled from near the boundary of each class susceptible to perturbations $s \in \mathbb{B}^n_\delta$ such that $f(x + s) \neq f(x)$.
In high dimensions, concentration effects ensure that data points sampled from either class concentrate close to the flat surface of their respective half ball, and therefore close to the decision surface.
This means that the probability of sampling a point which is susceptible to an adversarial attack is high, as encapsulated in the following result, which is proved in Section~\ref{sec:supplementary:hemisphere:susceptibility-proof}.

\begin{theorem}[Susceptible data points are typical]\label{thm:hemisphere:susceptibility}
    For any $\epsilon \geq 0$ and $\delta \in [\epsilon, 1 + \epsilon]$,  the probability that a randomly sampled data point is susceptible to an adversarial attack with Euclidean norm $\delta$ grows exponentially to 1 with the dimension $n$, specifically
    \begin{align*}
        P\big( x &\sim \mathcal{D}_\epsilon \suchthat \text{there exists } s \in \mathbb{B}^n_\delta \text{ such that } f(x + s) \neq f(x) \big)
        \\&
        \geq
        1 - (1 - (\delta - \epsilon)^2)^{n/2}.
    \end{align*}
\end{theorem}

Analogously to Theorem~\ref{thm:undetectability}, we also derive the following bound on the probability of a random perturbation destabilising a sampled data point, the proof of which is in Section~\ref{sec:supplementary:hemisphere:undetectability-proof}.

\begin{theorem}[Destabilising perturbations are rare]\label{thm:hemisphere:undetectability}
    For any $\delta > \epsilon \geq 0$, the probability that a randomly selected perturbation with Euclidean norm $\delta$ causes a randomly sampled data point to be misclassified is bounded from above as
    \begin{align*}
        &P\big(x \sim \mathcal{D}_\epsilon, s \sim \mathcal{U}(\mathbb{B}^n_\delta) \suchthat f(x + s) \neq f(x)\big)
        \\&\qquad\qquad\qquad
        \leq
        \frac{1}{4} \Big(1 - \Big( \frac{\epsilon}{\delta} \Big)^2 \Big)^{n/2}.
    \end{align*}
\end{theorem}

\begin{figure}
\centering
\includegraphics[width=\linewidth]{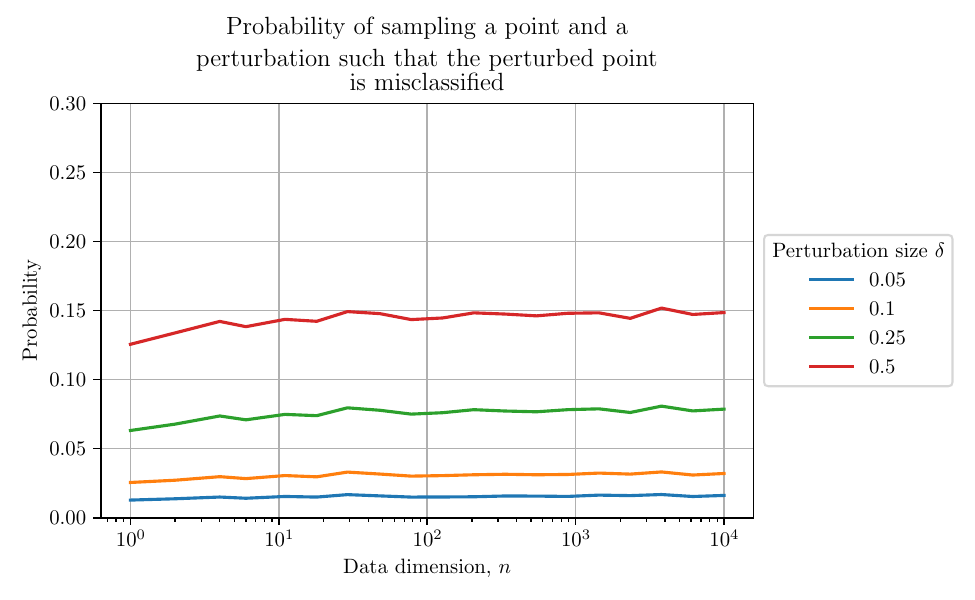}
\caption{Two half balls model with $\epsilon = 0$ --- The empirical probability of sampling a point and perturbation of size $\delta$ such that the perturbed data point is misclassified. This empirical data was computed by sampling 10,000 points from the half-ball distribution and 10,000 perturbations from $\mathcal{U}(\mathbb{B}^n_{\delta})$.}
\label{fig:NoMargins}
\end{figure}

Surprisingly, for $\epsilon = 0$ (when the two half balls meet along their flat faces) this probability does not converge to zero with increasing dimension $n$.
To illustrate that this is not simply a looseness in the bound, we present empirical data in Figure~\ref{fig:NoMargins} demonstrating that the probability of sampling a destabilising perturbation at random remains approximately constant, even in high dimensions.

A deeper theoretical analysis reveals that the probability of a label swap in the model with $\epsilon=0$ is always separated away from zero for all dimensions $n>1$. In particular, the following result holds, the proof of which is provided in Section~\ref{sec:supplementary:hemisphere:nohide-proof}.

\begin{theorem}[No place to hide when margins are zero]\label{thm:hemisphere:no_hiding} Consider the two half-balls model with $\epsilon=0$, $n>1$, and let $\delta>0$. Then
\[
\begin{split}
\lim_{n\rightarrow\infty} P(x\sim\mathcal{D}_\epsilon, \ s\sim\mathcal{U}(\mathbb{B}_\delta^n): \ f(x+s)\neq f(x)) \geq \\
\sup_{p\in(0,1)}2 p \left(1-\Phi\left(\frac{\sqrt{2|\log(1-p)|}}{\delta}\right)\right),
\end{split}
\]
where $\Phi$ is the standard cumulative distribution function.
\end{theorem}

According to Theorem \ref{thm:hemisphere:no_hiding}, the probability of label swaps due to additive and independent random perturbations sampled from $\mathcal{U}(\mathbb{B}_\delta^n)$ does not converge to zero when $n$ grows arbitrarily large in this model when  $\epsilon=0$.
This is in stark contrast with the case when the separation margin $\epsilon$ is non-zero, where the analogous upper bound from Theorem~\ref{thm:hemisphere:undetectability} goes to zero with increasing dimension $n$.
We conclude from these results that it is the presence of a non-zero margin $\epsilon > 0$ separating pairs of typical data points that is responsible for `hiding' the adversarial susceptibility of the classifier such that it cannot be { efficiently} detected using random perturbations.

\section{Discussion and relation to prior work}\label{sec:discussion}

\subsection{Existence of adversarial examples}
Since the seminal work \cite{szegedy2013intriguing} reporting the discovery of adversarial examples in deep neural networks, the topic of adversarial examples as well as their origins and the mechanisms behind their occurrence have been the focus of significant attention in theoretical and computational machine learning communities.
One hypothesis, expressed in \cite{szegedy2013intriguing} was that the existence of the adversarial examples could be attributed to the inherent instabilities -- i.e., large Jacobian norms leading to large Lipschitz constants for the classification maps.
Theorems~\ref{thm:susceptibility},  \ref{thm:undetectability} (see also Theorems~\ref{thm:hemisphere:susceptibility} and~\ref{thm:hemisphere:undetectability} in Section~\ref{sec:hemispheres_model})
show that whilst the latter mechanism may indeed constitute a feasible route for adversarial examples to occur, our presented framework reveals a simple pathway for adversarial data to emerge naturally in systems without large Jacobian norms.

\subsection{Fragility of adversarial examples}
It has been empirically observed in~\cite{kurakin2018adversarial, gupta2020applicability} that the capability of adversarial examples to fool the classifiers for which they have been designed can be hindered by perturbations and transformations which are naturally present in real-world environments.
Here we show and prove (Theorems~\ref{thm:undetectability} and~\ref{thm:hemisphere:undetectability}) that in the vicinity of the target images, adversarial examples may indeed occupy sets whose Lebesgue measure is exponentially small.
Hence, the addition of a small but appropriate perturbation to an example of that type will have the capability to make it non-adversarial.
Our results also show that simply adding random noise to an adversarially attacked image is very unlikely to produce something which would be correctly classified.
Taken together, these two observations suggest that random image perturbations have a significantly different effect on standard image classification models from natural environmental changes to images.

\subsection{Certifying robustness of classifiers to adversarial perturbations}\label{sec:discussion:adversarialDefence}
There is a body of work in the literature dedicated to detecting, mitigating, and defending against adversarial attacks using randomly sampled noise; see, for example, the algorithms discussed in~\cite{cohen2019certified, li2019certified, ye2024unit} amongst many others.
If many such randomly sampled perturbations are used, our results suggest that only a small fraction of them would change the classification of an image.
Indeed, this fraction is exponentially small (in the data dimension $n$) when the classifier's decision surface is locally linear around the perturbed data point (Theorems~\ref{thm:undetectability} and~\ref{thm:supplementary:generalisation:undetectability}).
Equivalently, this suggests that such algorithms would need to take exponentially many (in $n$) samples to find even one which changes the classification.
This implies that, to reliably detect or defend against an adversarial attack, algorithms based on this approach require an exponentially large computational complexity.

\subsection{Universal adversarial perturbations}
Another striking feature of adversarial examples is {the existence of seemingly universal adversarial perturbations.}
{These are small image-agnostic perturbations which can be applied to most, if not all, images in a dataset to cause the image to be misclassified by a given model.}
The phenomenon {of universal adversarial perturbations} was first reported in~\cite{moosavi2017universal} and since then observed in a wide range of tasks and architectures~\cite{chaubey2020universal}.
Several explanations justifying the existence of universal adversarial perturbations have been proposed in the literature.
This includes the view that universal perturbations may exploit correlated lower-dimensional structures in the classifier's decision boundaries.
It has been less clear how to explain the simultaneous existence, fragility, typicality, and universality of adversarial perturbations.
Theorems~\ref{thm:susceptibility}, \ref{thm:undetectability}, and \ref{thm:universality} show that the combination of these correlations with the high dimensionality of data may explain the co-existence of the typicality of adversarial examples, their fragility, and at the same time universality.

\subsection{Notions of stability}
Our results reveals a new unexplored relationship between stability and the existence of adversarial data.
We show that the ubiquitous presence of adversarial perturbations which destabilise the classifier is not contradictory to the robustness of the classifier to random perturbations of the data.
If we view the former as a form of \emph{deterministic instability} (i.e. there exist small, 
and potentially arbitrarily small,
destabilising perturbations which can be constructed by an attacker), and the latter as a form of \emph{probabilistic stability} (destabilising perturbations are unlikely to be sampled at random), it becomes apparent that the probabilistic stability is in fact masking the underlying instability.
Since these two notions of stability are clearly not equivalent, it is imperative to understand the difference between the two.
To clarify this intriguing relationship, let us first recall two relevant definitions of stability (cf.~\cite{huang2022}).  

\begin{definition}[$\epsilon$-stability]\label{def:epsilon-stability}
    The classification map $f:\Real^n\rightarrow \{0,1\}$ is $\epsilon$-stable at  $x$ if
    \begin{align*}
        f(x+s)=f(x) \ \text{ for all } \ s\in\mathbb{B}_\epsilon^n.
    \end{align*}
   {
    Otherwise, if there is an $s\in\mathbb{B}_\epsilon^n$ for which $f(x+s)\neq f(x)$, we say that the classification map $f$ is not $\epsilon$-stable at $x$, or that $f$ is $\epsilon$-unstable at $x$.
    }
\end{definition}

\begin{definition}[$\epsilon$-stability with confidence $\upsilon$]\label{def:epsilon-stablility-confidence}
    Let $\mu$ be a probability distribution on $\mathbb{B}_\epsilon^n$. The classification map  $f:\Real^n\rightarrow \{0,1\}$ is $\epsilon$-stable at $x$ with confidence $\upsilon$ w.r.t. the distribution $\mu$
    if
    \begin{align*}
        P( s \sim \mu \suchthat f(x+s) = f(x)) \geq \upsilon.
    \end{align*}
\end{definition}

\begin{figure}
\centering
\includegraphics[width=0.6\linewidth]{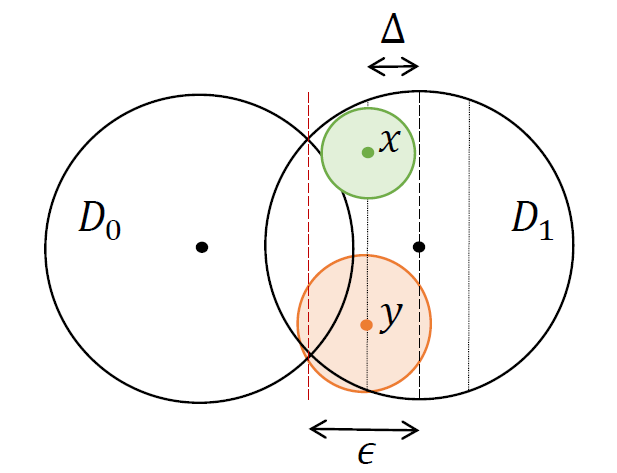}
\caption{Adversarial susceptibility of seemingly stable classifiers. Points $x$ and $y$ are in the $\Delta$ thickening of disc intersecting the ball $\mathcal{D}_1$ along one of its largest equators. For $n$ sufficiently large, most points sampled from $\mathcal{U}(D_1)$ belong to this domain. Both $x$ and $y$ are $(\epsilon-\Delta)$-stable. At the same time, they are also $\delta$-stable with confidence $\upsilon \eqsim 1$.}\label{fig:AdversarialStability}
\end{figure}

At the core of the phenomenon explored in Theorems \ref{thm:susceptibility} and \ref{thm:undetectability} is the fact that a ``typical'' point $x$ is $\delta$-stable with confidence $\upsilon$ with respect to perturbations sampled from $\mathcal{U}(\mathbb{B}^n_{\delta})$, where $\upsilon$ approaches $1$ exponentially in $n$.
This makes the finding of adversarial perturbations by adding random samples $s\sim \mathcal{U}(\mathbb{B}_\delta^n)$ difficult and unlikely.

At the same time, for $n$ sufficiently large, typical points are located in some $\Delta<\epsilon$ vicinity of the equators of the $n$-dimensional unit balls supporting $\mathcal{D}_0$ and $\mathcal{D}_1$.
This implies that these typical points are $\epsilon-\Delta$-stable in the sense of Definition~\ref{def:epsilon-stability}.
This is visualised in the diagram shown in Figure~\ref{fig:AdversarialStability}.
In the absence of the margin $\epsilon$ separating the centres of $\mathcal{D}_0$ and $\mathcal{D}_1$, there is no room to ``hide''  adversarial examples among random perturbations. 
This leads to the intriguing observation that, for some appropriate value of $\epsilon$:
\begin{quote}
\it
    The existence and prevalence of adversarial examples which are undetectable via random perturbations can be enabled by the $\epsilon$-stability of `typical' data samples.
\end{quote}

This is further illustrated through the two half-balls model investigated in Section~\ref{sec:hemispheres_model}.
The choice of sampling two data classes were sampled from complementary half-balls separated by margin $\epsilon \geq 0$ was motivated by its ability to represent two separable classes without any margin or overlap.
As shown numerically in Figure~\ref{fig:NoMargins}, in the absence of margins (which is an admissible case in the setup adopted in~\cite{shafahi2018adversarial}) the probability of registering misclassifications due to random perturbation is significant and does not change much with dimension.
This is confirmed theoretically by Theorem~\ref{thm:hemisphere:no_hiding}.

\subsection{Other theoretical frameworks explaining the phenomenon of adversarial examples}

Several works have presented feasible mechanisms explaining some elements of the paradox considered in this work. For example
in~\cite{shafahi2018adversarial}, \cite{tyukin2020adversarial} the authors exploited concentration of measure arguments to determine conditions when small destabilising perturbations can be typical in high dimensional settings. In \cite{fawzi2016robustness} the authors looked at the relationships between the relative sizes of class-altering perturbations in random directions and their worst-case counterparts (adversarial).  Sample-inefficiency of robust training with random noise as well as the impact of the choice of norms have been discussed in~\cite{khoury2018geometry}. Relevant geometric concepts explaining the feasibility of the expected emergence of adversarial examples have been suggested in  \cite{shamir2021dimpled} (the dimpled manifold hypothesis) and \cite{tanay2016boundary} (the boundary tilting mechanism).

In our work, we focused on presenting a single simple theoretical framework that could holistically explain the simultaneous rarity of destabilising random perturbations, the typicality of adversarial examples (see Figure~\ref{fig:NoMargins} and the discussion below), their universality, their potential fragility, and the relationship between the presence of non-zero independent on dimension separation margins (i.e. stability) and the possibility to successfully hide vulnerability to adversarial perturbations in the apparent robustness to random perturbations. Revealing the connection between all these phenomena within a single setting is a key feature of our framework.

The typicality of such coexistence in a broad class of problems distinguishes our work from other relevant theories and explanations focussing on showing the existence of tasks in which instabilities are expected in otherwise accurate classifiers (see e.g. \cite{bastounis2021mathematics}, \cite{Bastounis2023530}).

\section{Experimental investigation of the paradox of apparent stability}\label{sec:full-experiments}

We experimentally explored the paradox of apparent stability using several standard benchmark image classification datasets.
We first describe the experimental methodology in Section~\ref{sec:experimentalSetup}, and the results are reported in Section~\ref{sec:experimentalResults}.
The results for each benchmark are reported separately in the following sections:
\begin{itemize}
    \item results for the CIFAR-10 dataset~\cite{cifar10} are in Section~\ref{sec:experimentResults:cifar10}
    \item results for the Fashion MNIST dataset~\cite{fashionMNIST} are in Section~\ref{sec:experimentResults:fashionmnist}
    \item results for the German Traffic Sign Recognition Benchmark (GTSRB)~\cite{GTSRB} are in Section~\ref{sec:experimentResults:gtsrb}
    \item {results for the ImageNet benchmark~\cite{imagenet} are in Section~\ref{sec:experimentResults:imagenet}}
\end{itemize}
To present the phenomenon in the simplest possible setting, {for CIFAR-10, Fashion MNIST and GTSRB,} we arranged the $n$ classes of each benchmark dataset into $\frac{1}{2}n(n-1)$ binary classification problems.
A convolutional neural network was trained and assessed for each problem using a standardised protocol described below.
To complement these experiments, we also investigated two pre-trained foundation models on the 1000-class image classification ImageNet benchmark, as described in Section~\ref{sec:experimentSetup:imagenet}.

The results here are presented normalised to the setting of images with pixel values in $[0, 1]$, regardless of the native scaling of the datasets or pre-trained models.
This enables us to conveniently and comparably discuss the sizes of individual adversarial or random perturbations.

These results were computed using the CREATE HPC facilities at King's College London~\cite{CREATE}.

\subsection{Experimental setup}
\label{sec:experimentalSetup}

\subsubsection{Network architecture}\label{sec:supplementary:networkSetup}
Convolutional neural networks were trained on each of these problems, using a similar architecture and training regime for each problem.
Here, we describe the default settings, and any variations made for specific datasets are documented in the section describing the results computed on that dataset.
We used a simplification of the VGG architecture~\cite{simonyan2015deep}, the details of which are given in Table~\ref{tbl:architecture}.

For each pair-wise binary classification problem, the classes were assigned the labels 0 and 1, for compatibility with a standard sigmoid function on the output node of the network.
A mean square error loss function was used to train the network in Tensorflow~\cite{tensorflow} using stochastic gradient descent using a batch size of 128 for 100 epochs with Nesterov momentum parameter 0.9 and an initial learning rate of 0.1, which was halved every 20 epochs.
Dropout was used on the convolutional layers during training, with a parameter of 0.4.

For the binary classification problem of distinguishing class $i$ from class $j$, we denote the training set by $\mathcal{X}_{i,j}$, and the test set by $\mathcal{Y}_{i,j}$.
The subsets of training and test images which were correctly classified by the network are then denoted by $\mathcal{X}^{\operatorname{corr}}_{i,j} \subset \mathcal{X}_{i,j}$ and $\mathcal{Y}^{\operatorname{corr}}_{i,j} \subset \mathcal{Y}_{i,j}$ respectively.

We are therefore able to compute the training and test accuracy of the network for the binary classification problem involving class $i$ and class $j$ as the percentages
\begin{align}
    100 \frac{\operatorname{card}(\mathcal{X}^{\operatorname{corr}}_{i,j})}{\operatorname{card}(\mathcal{X}_{i,j})}
    \,\text{ and }\,
    100 \frac{\operatorname{card}(\mathcal{X}^{\operatorname{corr}}_{i,j})}{\operatorname{card}(\mathcal{X}_{i,j})},
\end{align}
respectively, where we use $\operatorname{card}$ to denote the cardinality of a set.

\begin{table}
    \centering
    \scalebox{0.75}{
    \begin{tabular}{c|c|c|c}
        Layer & Size & Output channels & Number of trained parameters
        \\
        \midrule
        Conv-1 & $3\times3$ & 64  & 1,792
        \\
        Conv-2 & $3\times3$ & 64 & 36,928
        \\
        Max pool & $2\times2$ & 
        \\
        Conv-3 & $3\times3$ & 128 & 73,856
        \\
        Conv-4 & $3\times3$ & 128 & 147,584
        \\
        Max pool & $2\times2$ & 
        \\
        Conv-5 & $3\times3$ & 256 & 295,168
        \\
        Conv-6 & $3\times3$ & 256 & 590,080
        \\
        Global max pool & &
        \\
        Dense && 512 & 131,584
        \\
        Dense && 1 & 513
    \end{tabular}
    }
    \caption{Architecture used for the binary classification problems. All convolutional layers do not pad their output, and are followed by a leaky ReLU activation function with leakiness parameter 0.1. The final dense layer has a standard sigmoid activation function. The number of trainable parameters depends on the size of the input data, and we use CIFAR-10 as an example.}
    \label{tbl:architecture}
\end{table}

\subsubsection{Adversarial attacks}\label{sec:supplementary:advAttackSetup}
To investigate the susceptibility of the networks to adversarial attacks, we used a standard gradient-based algorithm on a loss function, which can be viewed as an Euclidean version of the Fast Gradient Sign Method (FGSM) introduced in~\cite{goodfellow2014explaining}.
Specifically, if $L(x, y, N)$ denotes the mean square error loss function evaluated on the neural network $N$ at the target image $x$ with label $\ell$, we compute the adversarial attack direction as
\begin{align*}
    a(x) = \frac{\nabla_x L(x, \ell, N)}{\|\nabla_x L(x, \ell, N)\|},
\end{align*}
where $\|\cdot\|$ denotes the Euclidean norm.
We then tested 256 equally-spaced scalings $\epsilon \in [0, 5]$ to determine the smallest value such that $|\ell - N(x + \epsilon a(x))| > \frac{1}{2}$.
This value of $\epsilon$ therefore gives the Euclidean norm of the smallest perturbation (among those tested) in the direction of $a(x)$ such that the network therefore predicts the wrong class for the attacked image.
The value of $\epsilon$ therefore provides an upper bound on the minimal Euclidean distance of the image $x$ from the decision surface of the neural network $N$.

For the class $i$ vs class $j$ binary classification problem, we use $\mathcal{X}_{i,j}^{\operatorname{adv}} \subset \mathcal{X}^{\operatorname{corr}}_{i,j}$ to denote the set of training images $x \in \mathcal{X}^{\operatorname{corr}}_{i,j}$ such that $x$ was correctly classified by the network, but $x + \epsilon a(x)$ was misclassified for at least one of our tested values of $\epsilon$.
The equivalent subset of the test set is denoted by $\mathcal{Y}_{i,j}^{\operatorname{adv}} \subset \mathcal{Y}^{\operatorname{corr}}_{i,j}$.
We may then define the \emph{adversarial susceptibility} of the network for the training and test sets as the percentages
\begin{align}
    100 \frac{\operatorname{card} (\mathcal{X}_{i,j}^{\operatorname{adv}})}{\operatorname{card} (\mathcal{X}^{\operatorname{corr}}_{i,j})},
    \,\text{ and }\,
    100 \frac{\operatorname{card} (\mathcal{Y}_{i,j}^{\operatorname{adv}})}{\operatorname{card} (\mathcal{Y}^{\operatorname{corr}}_{i,j})},
    \label{eq:supplementary:adversarialSusceptibility}
\end{align}
respectively, where we use $\operatorname{card}$ to denote the cardinality of a set.

\subsubsection{Random perturbations}\label{sec:supplementary:randomPerturbationsSetup}
To assess the effect on the network of random perturbations to the images, we sampled a set $P$ of 2000 random perturbations from a uniform distribution on the $d$-dimensional ball with Euclidean norm $\leq 1$, where $d$ denotes the number of individual attributes of an image from the dataset.
Then, for each pair $i, j$ of classes, we performed the following experiment.
For each image $x$ in the subsets $\mathcal{X}_{i,j}^{\operatorname{adv}}$ and $\mathcal{Y}_{i,j}^{\operatorname{adv}}$ of the training and test sets which were susceptible to an adversarial attack, we constructed the perturbed image $x + \delta \epsilon s$ for each $s \in P$, where $\epsilon$ denotes the Euclidean norm of the smallest successful adversarial attack on $x$, scaled by each value of $\delta \in \{1, 2, 5, 10\}$ sequentially.
In other words, we evaluated the network on an image which was perturbed by a random perturbation with Euclidean norm scaled by a fixed multiple of that of the (known successful) adversarial attack.

For the class $i$ vs class $j$ binary classification problem, we define the set $\mathcal{X}^{\operatorname{rand},\delta}_{i,j} \subset \mathcal{X}^{\operatorname{adv}}_{i,j}$ as the set of images which were susceptible to one or more random perturbations with scaling factor $\delta$, as described above.
The set $\mathcal{Y}^{\operatorname{rand},\delta}_{i,j} \subset \mathcal{Y}^{\operatorname{adv}}_{i,j}$ is defined analogously on the test set of images.

This enables us to define the \emph{random perturbation susceptibility} of each network  for the training and test sets as the percentages
\begin{align}
    100 \frac{\operatorname{card} (\mathcal{X}_{i,j}^{\operatorname{rand}, \delta})}{\operatorname{card} (\mathcal{X}^{\operatorname{adv}}_{i,j})},
    \,\text{ and }\,
    100 \frac{\operatorname{card} (\mathcal{Y}_{i,j}^{\operatorname{rand}, \delta})}{\operatorname{card} (\mathcal{Y}^{\operatorname{adv}}_{i,j})},
    \label{eq:supplementary:randomSusceptibility}
\end{align}
respectively for each tested value of $\delta$, where we use $\operatorname{card}$ to denote the cardinality of a set.

\subsubsection{Training with random perturbations}\label{sec:supplementary:randomTraining}
We explored the effect of applying additive random noise to images during training on adversarial robustness.
For simplicity, we only explored this using the CIFAR-10 benchmark.
To do this we inserted a layer at the beginning of the network architecture described in Table~\ref{tbl:architecture} which sampled noise from a prescribed distribution independently for each input and added it to the input.
The precise random perturbation added to each image is therefore different each time the image is presented to the network during training.
The random perturbation layer is only active during training, so does not affect how the trained network is assessed at test time.
We experimented with noise sampled uniformly from the cube $[-a, a]^n$ (i.e. with maximum $\ell^{\infty}$ norm $a > 0$ and with noise sampled from the ball $\mathbb{B}^n_b$ (i.e. with maximum Euclidean norm $b > 0$), with $a \in \{0.1, 0.5, 1.0\}$, $b \in \{3.2, 16, 32\}$, where $n$ is the dimension of a single image in the dataset.
These values of $a$ and $b$ were selected to ensure that for each pair of $a$ and $b$ values the samples from each distribution would have approximately the same Euclidean norm on average.
This enables us to observe whether the sampling distribution makes a significant impact on the results, independently of the magnitude.
Each network was otherwise trained exactly as described in Section~\ref{sec:supplementary:networkSetup}.

\subsubsection{ImageNet experimental setup}\label{sec:experimentSetup:imagenet}
Experiments using the ImageNet image classification benchmark~\cite{imagenet} were performed using the pretrained VGG 19~\cite{simonyan2015deep} and ResNet50~\cite{he2016deep} neural networks available from Tensorflow~\cite{tensorflow}.
These architectures were selected because the VGG-19 network resembles the smaller networks we trained for the other datasets, while the ResNet50 architecture enables us to compare how our findings translate to a significantly different family of models.
For these experiments, we sampled 20,480 images from the standard validation split of the ImageNet dataset, and assessed the accuracy, adversarial susceptibility and random susceptibility of each network as described above.
Since ImageNet has 1,000 classes, we ensured that every class was represented in the sampled data, although did not require the same number of images from each class.
Our notions of adversarial and random susceptibility in this setting are `one-vs-all': an adversarial attack or random perturbation is considered to cause a misclassification if it causes the predicted class label to change to any other class.
Since our aim is simply to understand the relationship between random and worst-case perturbations, this treatment does not account for the widely-reported close semantic similarity between various pairs of ImageNet classes (see ~\cite{beyer2020imagenet}, for example).

\subsection{Experimental results}\label{sec:experimentalResults}

\subsubsection{Experimental results on CIFAR-10}\label{sec:experimentResults:cifar10}

\begin{table}
    \centering{
    \scalebox{0.8}{
    \begin{tabular}{cc}
        Index & Name
        \\
        \midrule
        0 & Aeroplane
        \\
        1 & Automobile
        \\
        2 & Bird
        \\
        3 & Cat
        \\
        4 & Deer
        \\
        5 & Dog
        \\
        6 & Frog
        \\
        7 & Horse
        \\
        8 & Ship
        \\
        9 & Truck
        \\
    \end{tabular}
    }
    \caption{CIFAR-10 --- Class names associated with each class index.}
    \label{tbl:classNames}
    }
\end{table}

The English names associated with each of the 10 classes are provided in Table~\ref{tbl:classNames}.

\paragraph{Network performance.}
The training and test accuracy of the networks trained on each of the binary classification problems is shown in Table~\ref{tbl:accuracy}.
The mean accuracy on the training set of the networks trained for all of the binary classification problems was 99.57\% (standard deviation 0.24), with a minimum of 98.74\%.
In comparison, the mean accuracy on the test set was 94.09\% (standard deviation 3.78), with a minimum of 82.6\%.
These figures indicate that the networks were generally quite capable to learning these binary classification problems, despite the fact they were trained using the same regime for only 100 training epochs each, and no specific tweaks were applied to improve the performance of any network.

\begin{table*}
    \centering
    \centering
    \scalebox{0.8}{
    \begin{tabular}{c|ccccccccc}
    \toprule
    {} &             1 &             2 &             3 &             4 &             5 &             6 &             7 &             8 &             9 \\
    \midrule
    0 &  99.88, 96.45 &  99.31, 91.70 &  99.40, 95.20 &  99.36, 94.65 &  99.67, 95.45 &  99.25, 96.05 &  99.49, 96.45 &  99.73, 94.10 &  99.77, 95.40 \\
    1 &               &  99.67, 96.65 &  99.46, 95.95 &  99.78, 98.10 &  99.75, 97.55 &  99.17, 96.85 &  99.91, 98.80 &  99.72, 96.85 &  99.77, 93.65 \\
    2 &               &               &  99.08, 85.20 &  99.70, 87.25 &  99.35, 87.05 &  99.68, 91.05 &  99.63, 92.90 &  99.58, 95.30 &  99.42, 95.80 \\
    3 &               &               &               &  98.74, 86.70 &  99.77, 82.60 &  99.09, 88.80 &  99.68, 91.35 &  99.53, 96.05 &  99.47, 95.05 \\
    4 &               &               &               &               &  99.59, 90.60 &  99.84, 94.90 &  99.78, 90.30 &  99.58, 97.20 &  99.42, 96.75 \\
    5 &               &               &               &               &               &  99.79, 93.95 &  99.72, 90.40 &  99.60, 96.85 &  99.41, 96.05 \\
    6 &               &               &               &               &               &               &  99.85, 96.70 &  99.40, 97.10 &  99.72, 97.20 \\
    7 &               &               &               &               &               &               &               &  99.70, 97.65 &  99.85, 97.60 \\
    8 &               &               &               &               &               &               &               &               &  99.78, 95.80 \\
    \bottomrule
    \end{tabular}
    }
    \caption{CIFAR-10 --- Accuracy of the networks on the binary classification problems, reported in the form `train accuracy, test accuracy', where accuracy is calculated as the percentage of images which were correctly classified. The row and column headers indicate the classes used in each binary classification problem.}
    \label{tbl:accuracy}
\end{table*}

\paragraph{Adversarial attacks.}
We report the adversarial susceptibility of each network (as defined in Section~\ref{sec:supplementary:advAttackSetup}) in Table~\ref{tbl:adversarialSusceptibility}.
On average over all the binary classification problems, 85.0\% of the training images were susceptible to an adversarial attack (standard deviation 9.71) with a minimum of 70.28\%, while the average on the test set was 79.48\% (standard deviation 7.91) with a minimum of 69.82\%.
We note that both minima were attained on the same task `frog-vs-ship' (6-vs-8).
In the vast majority of the binary classification problems, over 80\% of images in the training and test sets could be adversarially attacked in such a way that they would be misclassified by the network.
This demonstrates the susceptibility of all of the networks to adversarial attacks, implying that the decision surface in each case passes close to most of the points in the training and test sets.

To measure just how close the decision surface passes to each data point, we also explore the sizes of the computed adversarial perturbations measured in several norms.
In Table~\ref{tbl:advAttacks:l2Norms} we show the mean and standard deviations over each training and test set of the Euclidean norms of the smallest computed adversarial attack on each image.
Similarly, Table~\ref{tbl:advAttacks:l1Norms} shows the mean and standard deviations over each training and test set of the $\ell^1$ norms of the adversarial attacks, while Table~\ref{tbl:advAttacks:linftyNorms} shows the same information for the $\ell^{\infty}$ norms.

\begin{figure}
    \centering
    \includegraphics[width=\linewidth]{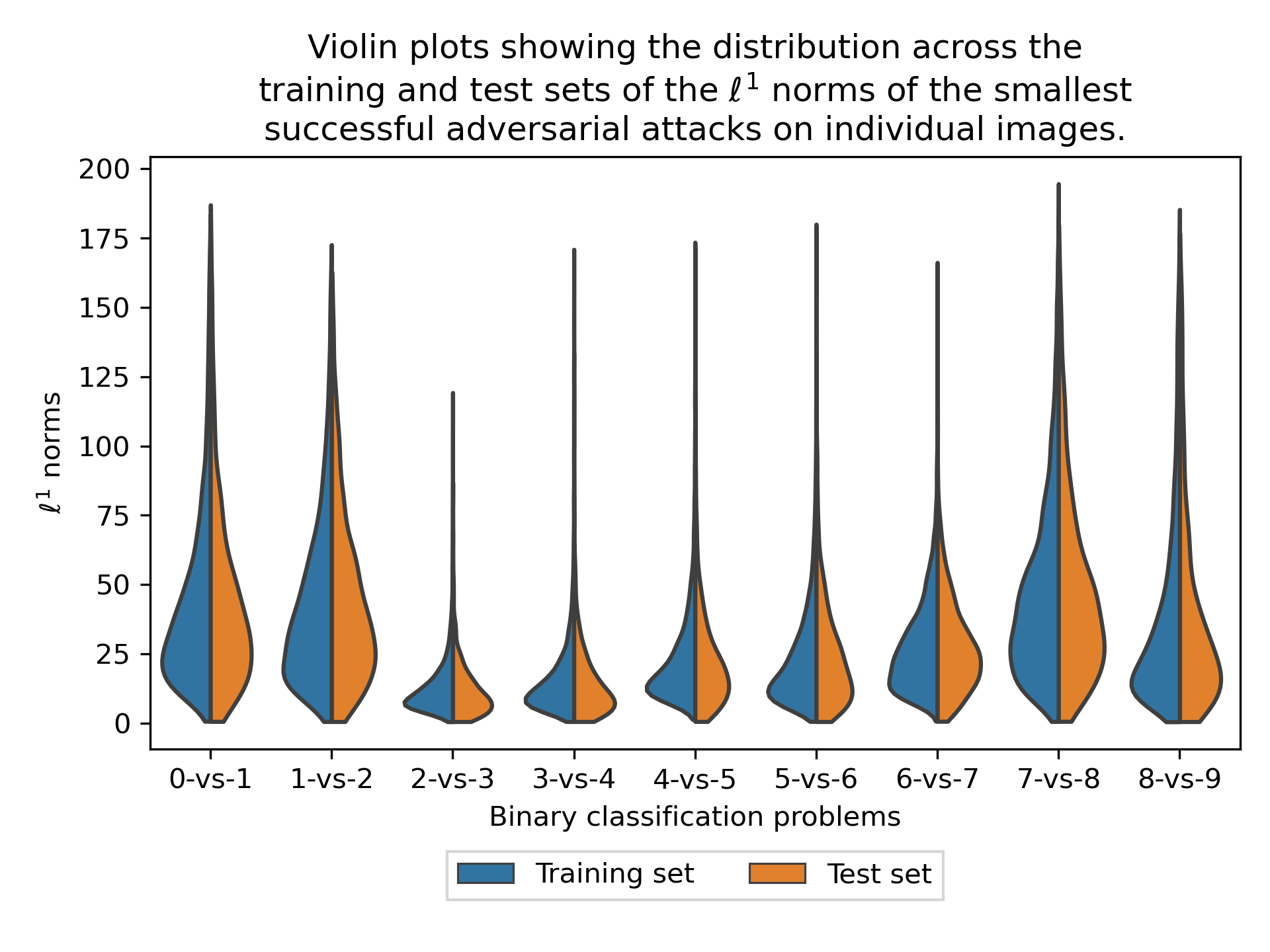}
    \caption{CIFAR-10 --- The distribution of the $\ell^1$ norms of the successful adversarial attacks found for each image using the algorithm in Section~\ref{sec:supplementary:advAttackSetup}, shown for a representative sample of the binary classification problems.
    The plotted distributions were fitted to the data using a standard Kernel Density Estimation algorithm and therefore only provide an approximation of the true empirical distribution.}
    \label{fig:adversarial:violin:l1}
\end{figure}

\begin{figure}
    \centering
    \includegraphics[width=\linewidth]{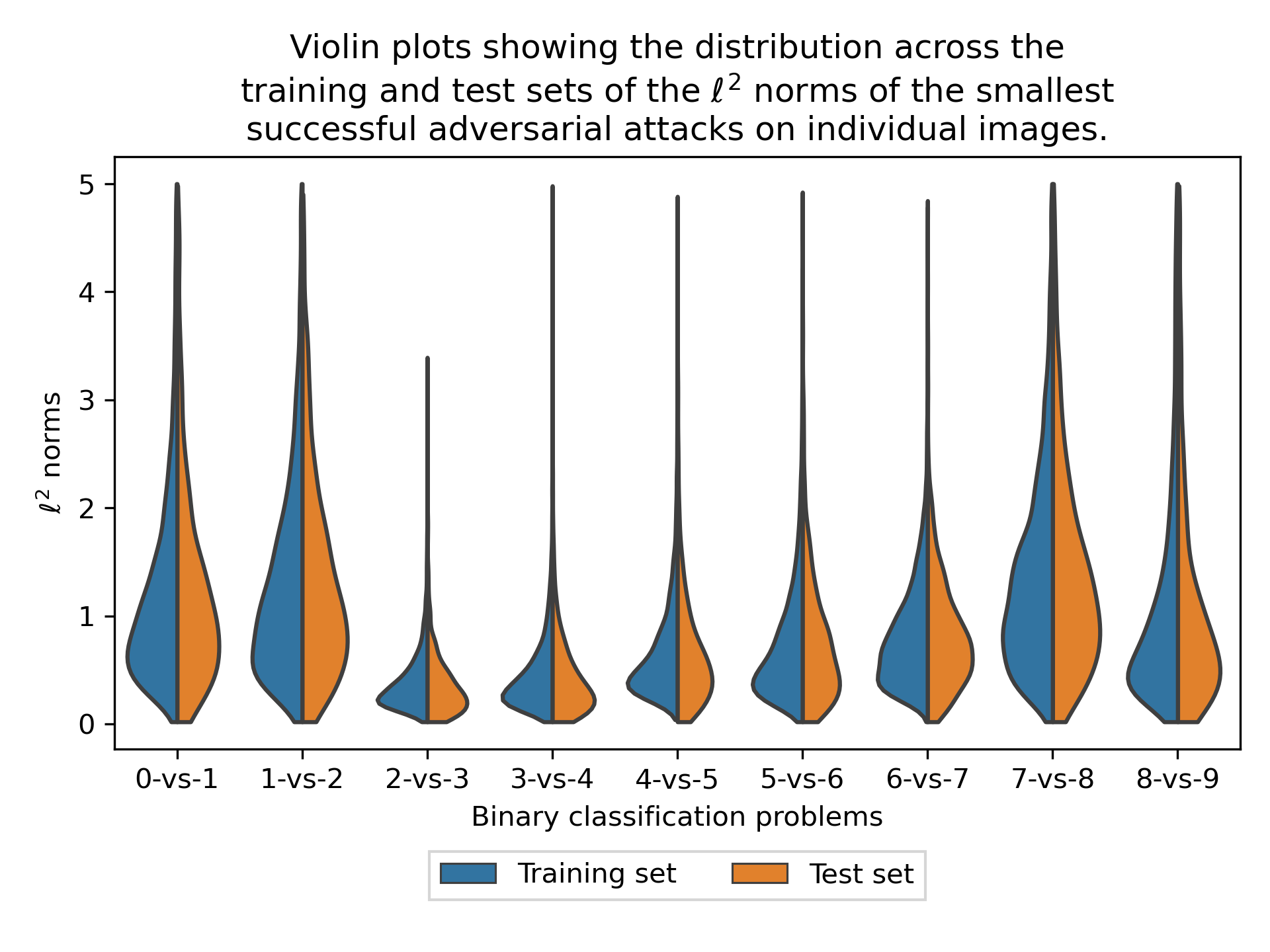}
    \caption{CIFAR-10 --- The distribution of the Euclidean norms of the successful adversarial attacks found for each image using the algorithm in Section~\ref{sec:supplementary:advAttackSetup}, shown for a representative sample of the binary classification problems.
    The plotted distributions were fitted to the data using a standard Kernel Density Estimation algorithm and therefore only provide an approximation of the true empirical distribution.}
    \label{fig:adversarial:violin:l2}
\end{figure}

\begin{figure}
    \centering
    \includegraphics[width=\linewidth]{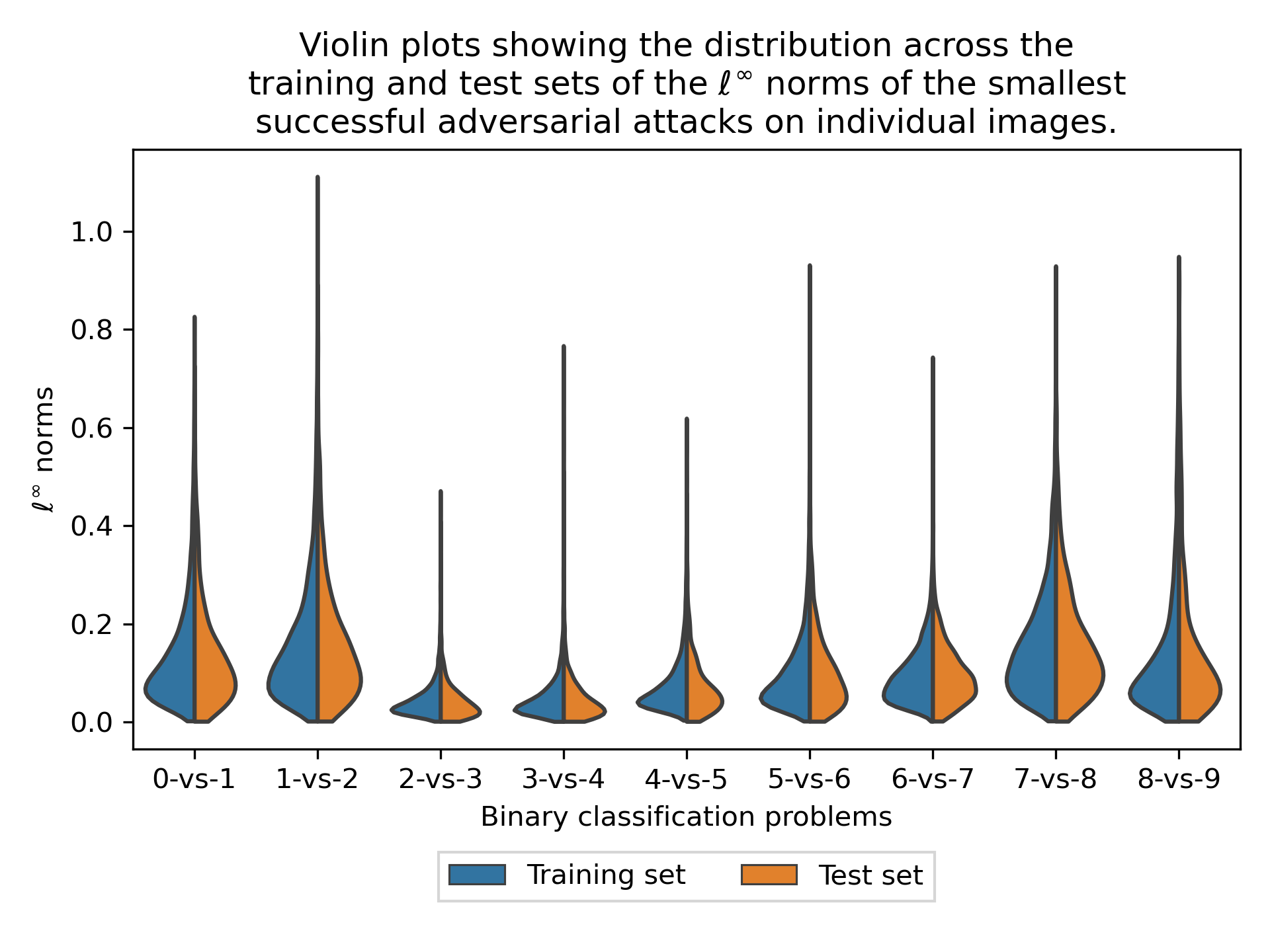}
    \caption{CIFAR-10 --- The distribution of the $\ell^\infty$ norms of the successful adversarial attacks found for each image using the algorithm in Section~\ref{sec:supplementary:advAttackSetup}, shown for a representative sample of the binary classification problems.
    The plotted distributions were fitted to the data using a standard Kernel Density Estimation algorithm and therefore only provide an approximation of the true empirical distribution.}
    \label{fig:adversarial:violin:linfty}
\end{figure}

The summary statistics reported in these tables are broken down in violin plots for a representative sample of the binary classification problems (selected, for simplicity, as the `$i$-vs-$(i + 1)$' problems).
These show an approximation of the distribution of the Euclidean norms (Figure~\ref{fig:adversarial:violin:l2}), $\ell^{\infty}$ norms (Figure~\ref{fig:adversarial:violin:linfty}) and $\ell^1$ norms (Figure~\ref{fig:adversarial:violin:l1}) of adversarial attacks.
In each case, this is the distribution across the whole training or test set of the norm of the smallest misclassifying adversarial attack found for each image using the algorithm described in Section~\ref{sec:supplementary:advAttackSetup}.
It is clear from these plots that for the majority of the adversarial attacks
the largest change to any individual pixel value is comparatively small: the $\ell^{\infty}$ norm is less than 0.2 for most of the images across all tasks.
The $\ell^1$ norms, on the other hand, compute the sum of the absolute values of all changes to all pixels, so are expected to be a much larger value. 
Scaling these $\ell^1$ norms by the number of pixel channels ($32\times32\times3 = 3,072$), we obtain the mean absolute change to a single pixel. 
Taking 100 as a representative maximum value for the $\ell^1$ norm across the majority of cases, we can therefore observe that this would correspond to a mean absolute change of approximately 0.03.
Comparing this value to a similarly representative value of less than 0.5 for the $\ell^{\infty}$ norm of the adversarial attack, it is clear that this implies that the attacks are typically very localised since most of the change must be focused in just a few pixels.

The plots also indicate that the networks trained on certain tasks (such as `bird-vs-cat' (2-vs-3) and `cat-vs-deer' (3-vs-4) seem to be much more susceptible to small adversarial attacks.
We mean this in the sense that while the overall attack susceptibility (Table~\ref{tbl:adversarialSusceptibility}) is quite typical, the attacks themselves on these classes appear to have much smaller norms.

The conclusion from these experiments is that most points in all of the training and test sets lie very close to the decision surface of the neural network, implying that the networks are susceptible to small perturbations to most of their training and test data.

\begin{table*}
    \centering
    \scalebox{0.8}{
    \begin{tabular}{c|ccccccccc}
    \toprule
    {} &             1 &             2 &             3 &             4 &             5 &             6 &             7 &             8 &             9 \\
    \midrule
    0 &  88.33, 86.88 &  95.69, 96.24 &  86.74, 86.08 &  95.19, 94.35 &  85.27, 83.97 &  83.43, 81.68 &  89.47, 88.02 &  92.78, 91.87 &  74.23, 73.69 \\
    1 &               &  86.25, 86.08 &  84.52, 83.79 &  93.59, 92.92 &  89.94, 88.83 &  85.23, 85.91 &  97.17, 97.17 &  89.62, 88.80 &  87.85, 88.04 \\
    2 &               &               &  91.88, 89.96 &  99.54, 99.66 &  96.93, 96.50 &  96.75, 96.05 &  92.97, 93.16 &  85.14, 83.89 &  92.73, 92.75 \\
    3 &               &               &               &  93.85, 92.85 &  99.86, 99.82 &  98.49, 98.65 &  98.95, 98.63 &  91.93, 92.04 &  78.84, 78.12 \\
    4 &               &               &               &               &  98.97, 98.34 &  98.99, 98.79 &  99.77, 99.56 &  80.96, 80.45 &  90.02, 90.08 \\
    5 &               &               &               &               &               &  96.19, 95.48 &  99.61, 99.61 &  84.14, 83.32 &  72.53, 71.63 \\
    6 &               &               &               &               &               &               &  98.63, 98.76 &  70.28, 69.82 &  87.35, 88.22 \\
    7 &               &               &               &               &               &               &               &  82.60, 83.56 &  96.85, 96.82 \\
    8 &               &               &               &               &               &               &               &               &  79.41, 78.03 \\
    \bottomrule
    \end{tabular}
    }
    \caption{CIFAR-10 --- Susceptibility of the networks to adversarial attacks, reported in the form `train susceptibility, test susceptibility', where susceptibility is calculated as in~\eqref{eq:supplementary:adversarialSusceptibility} as the percentage of images from the training set and test set which were misclassified after an adversarial attack using the algorithm described in Section~\ref{sec:supplementary:advAttackSetup}. 
    The row and column headers indicate the classes used in each binary classification problem.}
    \label{tbl:adversarialSusceptibility}
\end{table*}

\begin{table*}
    \centering
    \scalebox{0.8}{
    \begin{tabular}{r|ccccccccc}
    \toprule
    {} &            1 &            2 &            3 &            4 &            5 &            6 &            7 &            8 &            9 \\
    \midrule
    0 train &  1.25 (0.92) &  0.76 (0.60) &  1.19 (0.86) &  0.96 (0.84) &  1.32 (0.94) &  1.21 (0.83) &  1.24 (0.92) &  0.84 (0.81) &  1.20 (1.01) \\
     test  &  1.25 (0.94) &  0.79 (0.66) &  1.21 (0.92) &  0.97 (0.81) &  1.36 (0.97) &  1.24 (0.83) &  1.26 (0.94) &  0.83 (0.83) &  1.25 (1.03) \\
    \midrule
    1 train &              &  1.32 (0.93) &  1.00 (0.58) &  1.13 (0.82) &  1.27 (0.71) &  0.96 (0.63) &  1.50 (0.92) &  1.08 (0.86) &  0.75 (0.74) \\
     test  &              &  1.35 (0.96) &  1.00 (0.58) &  1.14 (0.86) &  1.28 (0.70) &  0.99 (0.62) &  1.47 (0.91) &  1.06 (0.87) &  0.74 (0.75) \\
    \midrule
    2 train &              &              &  0.40 (0.31) &  0.48 (0.37) &  0.64 (0.53) &  0.56 (0.45) &  0.86 (0.70) &  1.05 (0.72) &  1.21 (0.89) \\
     test  &              &              &  0.37 (0.32) &  0.47 (0.41) &  0.63 (0.58) &  0.57 (0.51) &  0.86 (0.75) &  1.06 (0.75) &  1.22 (0.90) \\
    \midrule
    3 train &              &              &              &  0.47 (0.41) &  0.40 (0.26) &  0.74 (0.55) &  0.58 (0.45) &  1.27 (0.87) &  0.72 (0.45) \\
     test  &              &              &              &  0.44 (0.40) &  0.35 (0.31) &  0.73 (0.58) &  0.57 (0.46) &  1.29 (0.89) &  0.72 (0.46) \\
    \midrule
    4 train &              &              &              &              &  0.68 (0.50) &  0.54 (0.40) &  0.65 (0.44) &  1.17 (0.90) &  1.03 (0.75) \\
     test  &              &              &              &              &  0.67 (0.53) &  0.53 (0.43) &  0.63 (0.50) &  1.21 (0.91) &  1.07 (0.76) \\
    \midrule
    5 train &              &              &              &              &              &  0.75 (0.58) &  0.73 (0.50) &  1.33 (0.90) &  0.94 (0.68) \\
     test  &              &              &              &              &              &  0.77 (0.59) &  0.70 (0.53) &  1.32 (0.90) &  0.98 (0.71) \\
    \midrule
    6 train &              &              &              &              &              &              &  0.82 (0.49) &  1.38 (1.06) &  0.94 (0.56) \\
     test  &              &              &              &              &              &              &  0.83 (0.51) &  1.42 (1.08) &  0.97 (0.58) \\
    \midrule
    7 train &              &              &              &              &              &              &              &  1.39 (0.93) &  1.05 (0.69) \\
     test  &              &              &              &              &              &              &              &  1.41 (0.97) &  1.05 (0.66) \\
    \midrule
    8 train &              &              &              &              &              &              &              &              &  1.14 (1.02) \\
     test  &              &              &              &              &              &              &              &              &  1.13 (1.05) \\
    \bottomrule
    \end{tabular}
    }
    \caption{CIFAR-10 --- Means and standard deviations of the Euclidean norms norms of the smallest successful adversarial attack on each image in the training and test set, reported in the form `mean (standard deviation)'. The numbers in the row and column headers indicate the classes used in each binary classification problem. The `train' row shows the values computed over the training set, while the `test' row shows the values computed over the test set.}
    \label{tbl:advAttacks:l2Norms}
\end{table*}

\begin{table*}
    \centering
    \scalebox{0.7}{
    \begin{tabular}{r|ccccccccc}
    \toprule
    {} &            1 &            2 &            3 &            4 &            5 &            6 &            7 &            8 &            9 \\
    \midrule
    0 train &  43.58 (31.63) &  24.81 (19.52) &  40.77 (30.04) &  32.93 (29.12) &  43.40 (31.07) &  40.08 (27.55) &  43.21 (32.21) &  27.75 (26.27) &  41.17 (34.03) \\
    test  &  43.75 (32.53) &  25.80 (21.35) &  41.29 (31.76) &  33.23 (28.07) &  44.58 (31.90) &  41.01 (27.42) &  43.99 (32.61) &  27.31 (26.69) &  42.73 (35.09) \\
    \midrule
    1 train &                &  43.67 (30.41) &  31.26 (18.66) &  37.92 (26.88) &  41.80 (23.68) &  30.91 (20.83) &  50.20 (30.37) &  36.27 (28.58) &  23.48 (22.71) \\
     test  &                &  44.62 (31.28) &  31.42 (18.83) &  38.38 (28.13) &  42.21 (23.44) &  31.85 (20.46) &  49.17 (29.95) &  35.49 (28.86) &  23.24 (23.52) \\
     \midrule
    2 train &                &                &  13.21 (10.15) &  15.77 (12.02) &  20.75 (17.32) &  18.44 (14.49) &  28.16 (23.22) &  34.41 (23.75) &  39.90 (29.16) \\
     test  &                &                &  12.27 (10.59) &  15.37 (13.53) &  20.27 (18.76) &  19.03 (16.61) &  28.24 (24.51) &  34.76 (24.91) &  40.46 (29.26) \\
     \midrule
    3 train &                &                &                &  15.75 (13.50) &   12.88 (8.26) &  23.30 (17.50) &  19.21 (14.39) &  43.05 (29.57) &  23.97 (15.48) \\
     test  &                &                &                &  14.86 (13.22) &   11.22 (9.67) &  23.05 (18.58) &  18.90 (15.04) &  43.74 (30.26) &  24.07 (15.69) \\
    \midrule
    4 train &                &                &                &                &  23.18 (17.02) &  18.52 (13.60) &  19.96 (13.64) &  38.80 (30.40) &  33.26 (23.93) \\
     test  &                &                &                &                &  22.81 (18.14) &  18.22 (14.59) &  19.39 (15.33) &  40.01 (30.97) &  34.66 (24.06) \\
    \midrule
    5 train &                &                &                &                &                &  23.89 (18.77) &  23.76 (15.43) &  45.22 (30.63) &  30.60 (22.05) \\
      test  &                &                &                &                &                &  24.53 (19.03) &  22.78 (16.78) &  45.22 (30.79) &  31.86 (23.00) \\
    \midrule
    6 train &                &                &                &                &                &                &  27.77 (16.80) &  46.02 (35.74) &  30.55 (18.78) \\
      test  &                &                &                &                &                &                &  27.97 (17.57) &  47.44 (36.49) &  31.66 (19.39) \\
    \midrule
    7 train &                &                &                &                &                &                &                &  47.05 (31.64) &  35.33 (22.86) \\
     test  &                &                &                &                &                &                &                &  47.78 (33.38) &  35.27 (22.17) \\
    \midrule
    8 train &                &                &                &                &                &                &                &                &  38.53 (34.28) \\
     test  &                &                &                &                &                &                &                &                &  38.29 (35.17) \\
    \bottomrule
    \end{tabular}
    }
    \caption{CIFAR-10 --- Means and standard deviations of the $\ell^1$ norms of the successful adversarial attacks on each training and test set, reported in the form `mean (standard deviation)'. The numbers in the row and column headers indicate the classes used in each binary classification problem. The `train' row shows the values computed over the training set, while the `test' row shows the values computed over the test set.}
    \label{tbl:advAttacks:l1Norms}
\end{table*}

\begin{table*}
    \centering
    \scalebox{0.8}{
    \begin{tabular}{r|ccccccccc}
    \toprule
    {} &            1 &            2 &            3 &            4 &            5 &            6 &            7 &            8 &            9 \\
    \midrule
    0 train &  0.14 (0.11) &  0.09 (0.08) &  0.13 (0.09) &  0.11 (0.10) &  0.16 (0.12) &  0.14 (0.10) &  0.14 (0.11) &  0.10 (0.10) &  0.14 (0.12) \\
    test  &  0.14 (0.11) &  0.10 (0.09) &  0.13 (0.10) &  0.12 (0.10) &  0.17 (0.13) &  0.14 (0.10) &  0.15 (0.12) &  0.10 (0.10) &  0.14 (0.12) \\
    \midrule
    1 train &              &  0.16 (0.12) &  0.13 (0.08) &  0.13 (0.10) &  0.15 (0.09) &  0.11 (0.07) &  0.18 (0.12) &  0.13 (0.11) &  0.10 (0.10) \\
    test  &              &  0.16 (0.13) &  0.13 (0.08) &  0.13 (0.11) &  0.16 (0.09) &  0.11 (0.07) &  0.18 (0.12) &  0.13 (0.11) &  0.09 (0.10) \\
    \midrule
    2 train &              &              &  0.05 (0.04) &  0.06 (0.04) &  0.08 (0.07) &  0.07 (0.06) &  0.11 (0.10) &  0.13 (0.09) &  0.15 (0.12) \\
    test  &              &              &  0.04 (0.04) &  0.05 (0.05) &  0.08 (0.07) &  0.07 (0.06) &  0.11 (0.11) &  0.13 (0.10) &  0.15 (0.12) \\
    \midrule
    3 train &              &              &              &  0.05 (0.05) &  0.06 (0.04) &  0.10 (0.07) &  0.08 (0.06) &  0.15 (0.11) &  0.09 (0.06) \\
    test  &              &              &              &  0.05 (0.05) &  0.05 (0.05) &  0.10 (0.08) &  0.07 (0.07) &  0.15 (0.11) &  0.09 (0.06) \\
    \midrule
    4 train &              &              &              &              &  0.08 (0.06) &  0.06 (0.05) &  0.09 (0.07) &  0.15 (0.11) &  0.13 (0.10) \\
    test  &              &              &              &              &  0.07 (0.06) &  0.06 (0.06) &  0.09 (0.07) &  0.15 (0.11) &  0.13 (0.10) \\
    \midrule
    5 train &              &              &              &              &              &  0.10 (0.08) &  0.10 (0.08) &  0.15 (0.11) &  0.12 (0.09) \\
    test  &              &              &              &              &              &  0.10 (0.08) &  0.10 (0.08) &  0.15 (0.11) &  0.12 (0.10) \\
    \midrule
    6 train &              &              &              &              &              &              &  0.10 (0.06) &  0.17 (0.13) &  0.11 (0.07) \\
    test  &              &              &              &              &              &              &  0.10 (0.06) &  0.17 (0.14) &  0.12 (0.07) \\
    \midrule
    7 train &              &              &              &              &              &              &              &  0.16 (0.11) &  0.13 (0.09) \\
    test  &              &              &              &              &              &              &              &  0.16 (0.12) &  0.13 (0.08) \\
    \midrule
    8 train &              &              &              &              &              &              &              &              &  0.15 (0.14) \\
    test  &              &              &              &              &              &              &              &              &  0.15 (0.15) \\
    \bottomrule
    \end{tabular}
    }
    \caption{CIFAR-10 --- Means and standard deviations of the $\ell^{\infty}$ norms of the smallest successful adversarial attack on each image in the training and test set, reported in the form `mean (standard deviation)'. The numbers in the row and column headers indicate the classes used in each binary classification problem. The `train' row shows the values computed over the training set, while the `test' row shows the values computed over the test set.}
    \label{tbl:advAttacks:linftyNorms}
\end{table*}

\paragraph{Random perturbations}
To explore whether the adversarial sensitivities described above could be triggered by random perturbations to the input data, we used the approach outlined in Section~\ref{sec:supplementary:randomPerturbationsSetup}.
We report the random perturbation susceptibility of each network (as defined in Section~\ref{sec:supplementary:randomPerturbationsSetup}) in
Table~\ref{tbl:randomPerturbationSuccessDelta1} for $\delta = 1$, Table~\ref{tbl:randomPerturbationSuccessDelta2} for $\delta = 2$, Table~\ref{tbl:randomPerturbationSuccessDelta5} for $\delta = 5$, and Table~\ref{tbl:randomPerturbationSuccessDelta10} for $\delta = 10$.

\begin{figure}
    \begin{subfigure}[t]{0.23\textwidth}
        \includegraphics[width=\linewidth]{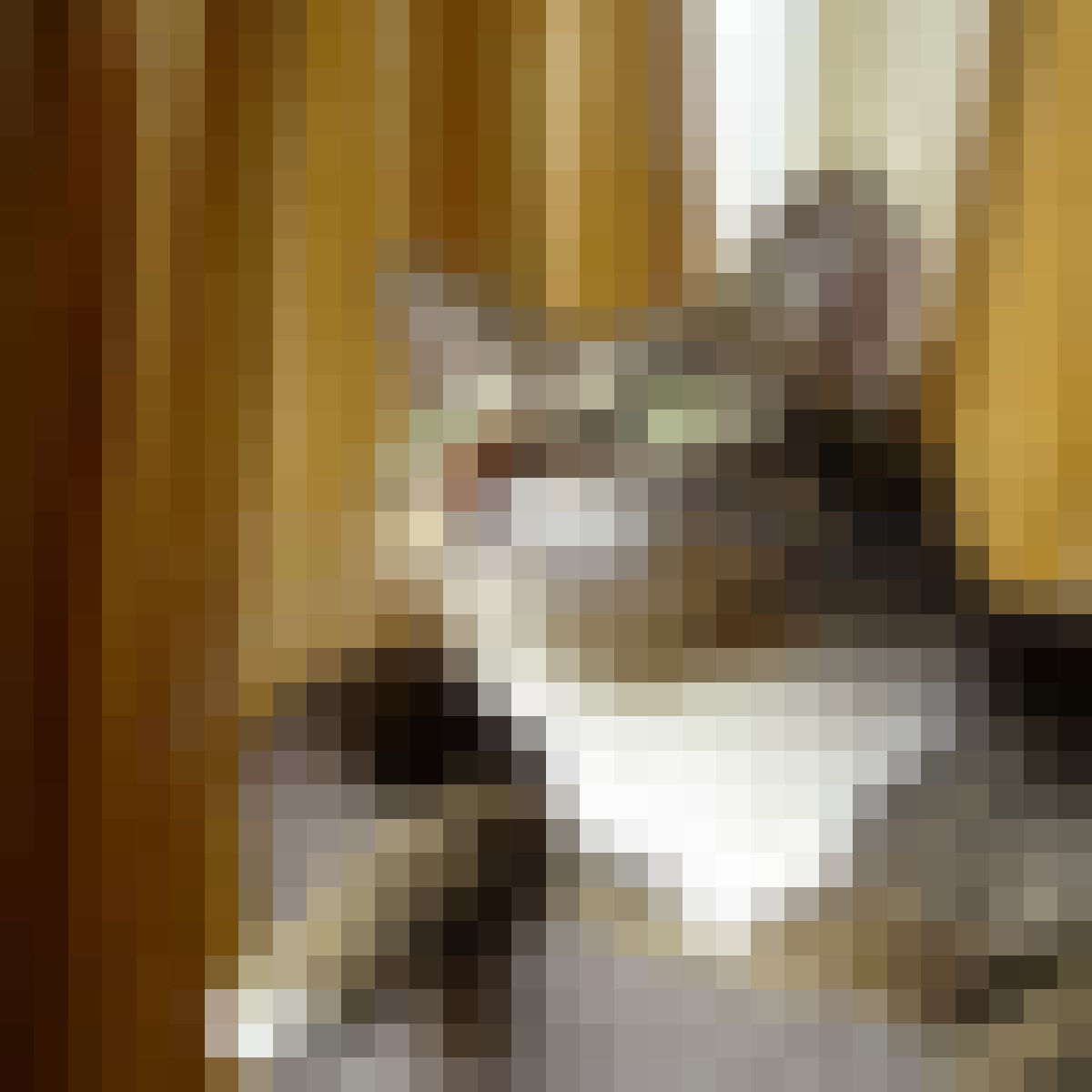}
        \caption{Original}
    \end{subfigure}
    \hfill
    \begin{subfigure}[t]{0.23\textwidth}
        \includegraphics[width=\linewidth]{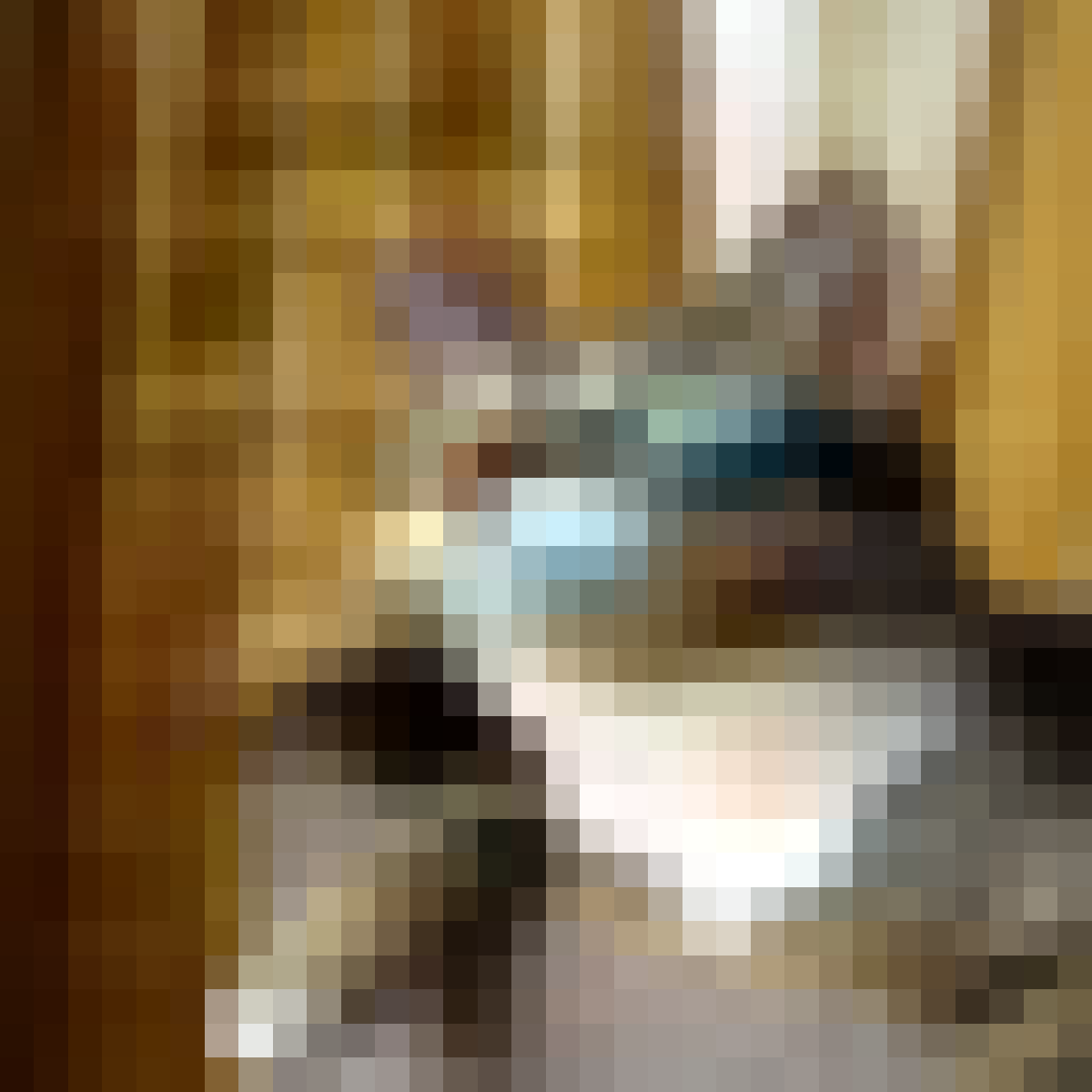}
        \caption{Misclassified adversarially attacked cat (attack \\
        $\ell^1$ norm: 67.98, \\
        Euclidean norm: 1.94, \\
        $\ell^{\infty}$ norm: 0.19)}
    \end{subfigure}
    \hfill
    \begin{subfigure}[t]{0.23\textwidth}
        \includegraphics[width=\linewidth]{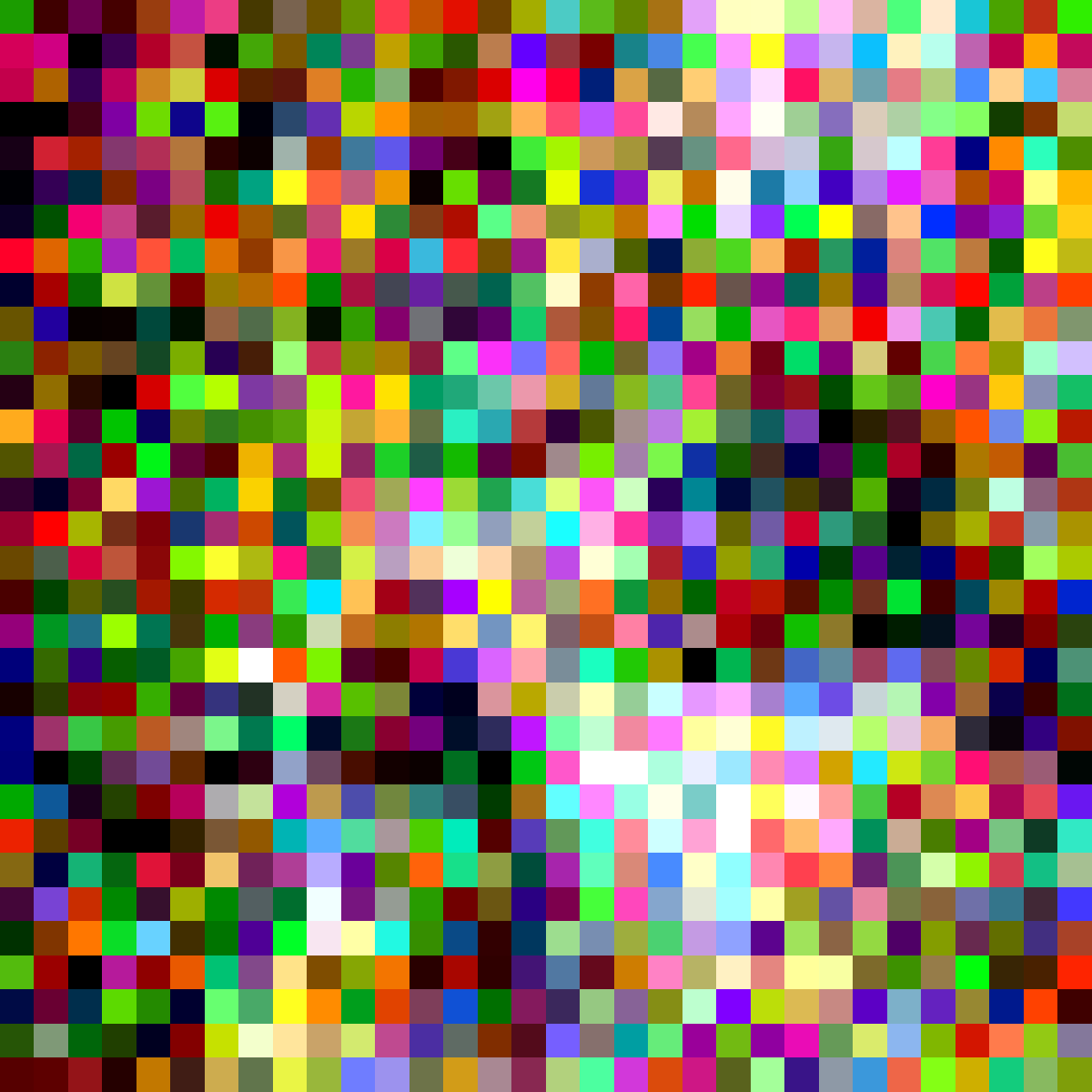}
        \caption{Correctly classified randomly perturbed cat ($\delta = 10$, perturbation \\
        $\ell^1$ norm: 856.28, \\
        Euclidean norm: 19.41, \\
        $\ell^{\infty}$ norm: 1.83)
        }
    \end{subfigure}
    \hfill
    \begin{subfigure}[t]{0.23\textwidth}
        \includegraphics[width=\linewidth]{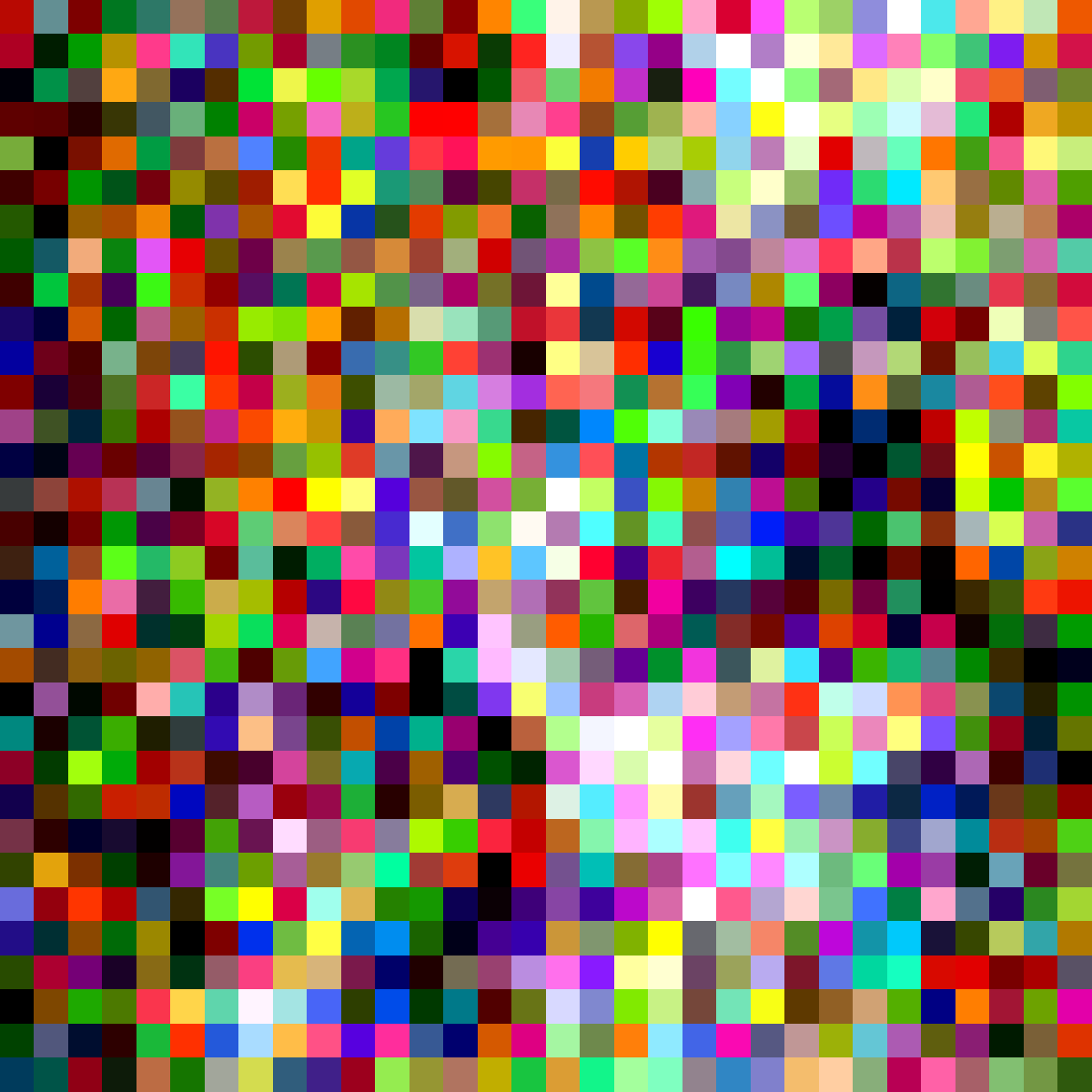}
        \caption{
        Misclassified randomly perturbed cat ($\delta = 10$, perturbation \\
        $\ell^1$ norm: 860.64, \\
        Euclidean norm: 19.41, \\
        $\ell^{\infty}$ norm: 1.65)
        }
    \end{subfigure}
    \caption{CIFAR-10 --- An example of an adversarially attacked image of a cat, taken from the `cat-vs-aeroplane' binary classification problem (0-vs-3), alongside examples of large random perturbations to the same image which did and did not cause the network to misclassify the image. Of the 2,000 sampled random perturbations, 83 (4.15\%) caused this image to be misclassified. Components of the modified image which were outside the range $[0, 1]$ have been clipped into the range for plotting, although not for the classification.}
    \label{fig:catAttacks}
\end{figure}

\begin{figure}
    \begin{subfigure}[t]{0.23\textwidth}
        \includegraphics[width=\linewidth]{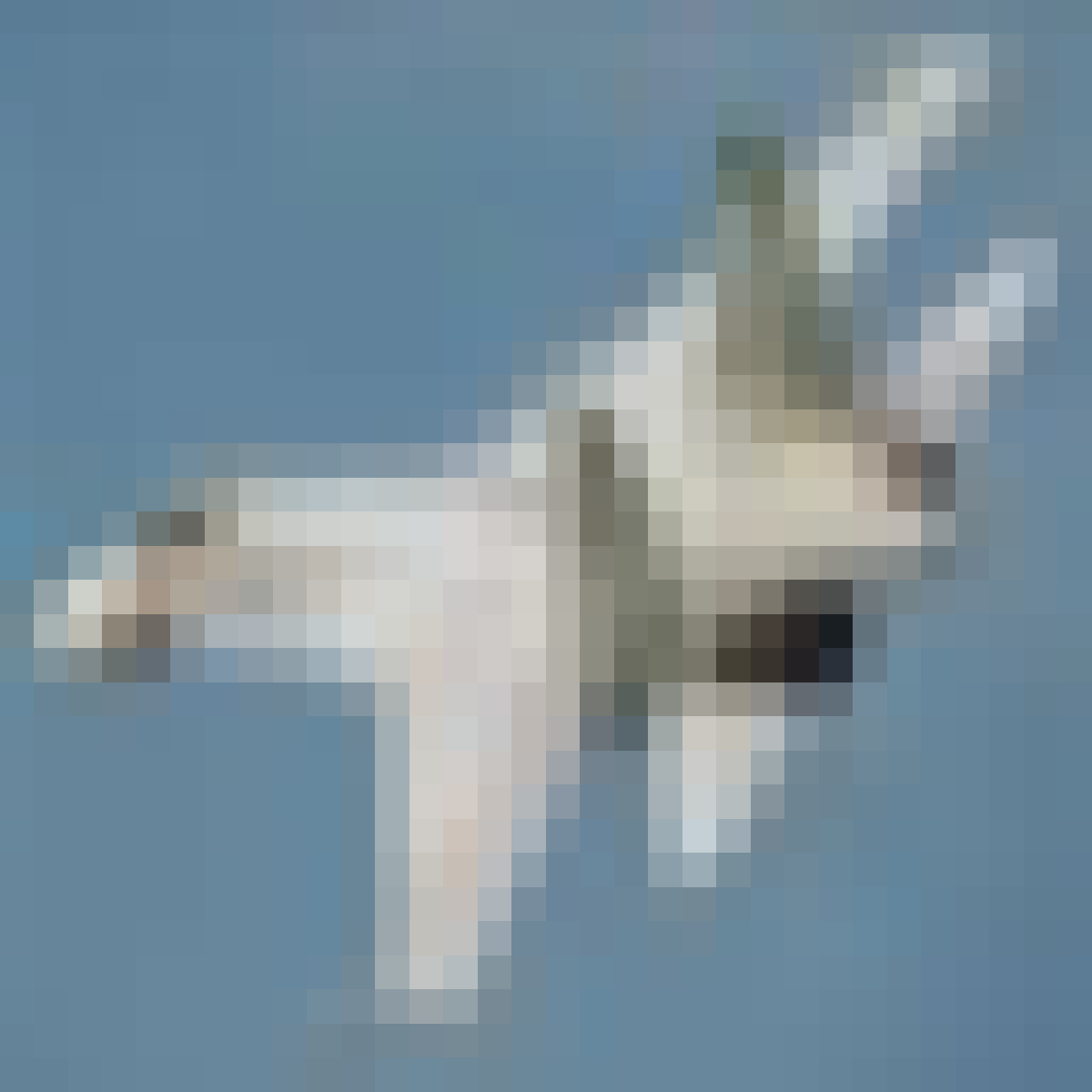}
        \caption{Original}
    \end{subfigure}
    \hfill
    \begin{subfigure}[t]{0.23\textwidth}
        \includegraphics[width=\linewidth]{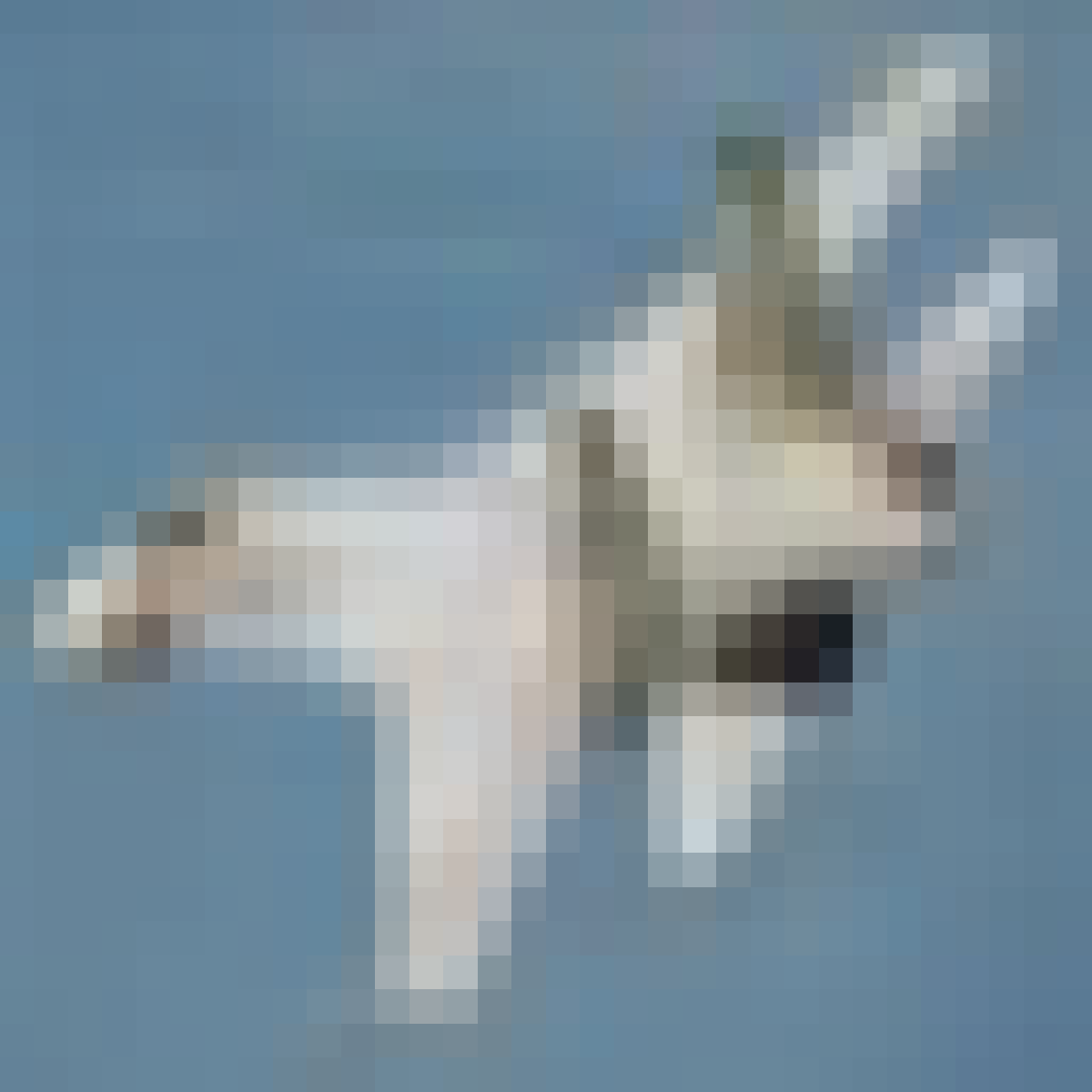}
        \caption{Misclassified adversarially attacked aeroplane (attack \\
        $\ell^1$ norm: 13.53, \\
        Euclidean norm: 0.39, \\
        $\ell^{\infty}$ norm: 0.05)}
    \end{subfigure}
    \hfill
    \begin{subfigure}[t]{0.23\textwidth}
        \includegraphics[width=\linewidth]{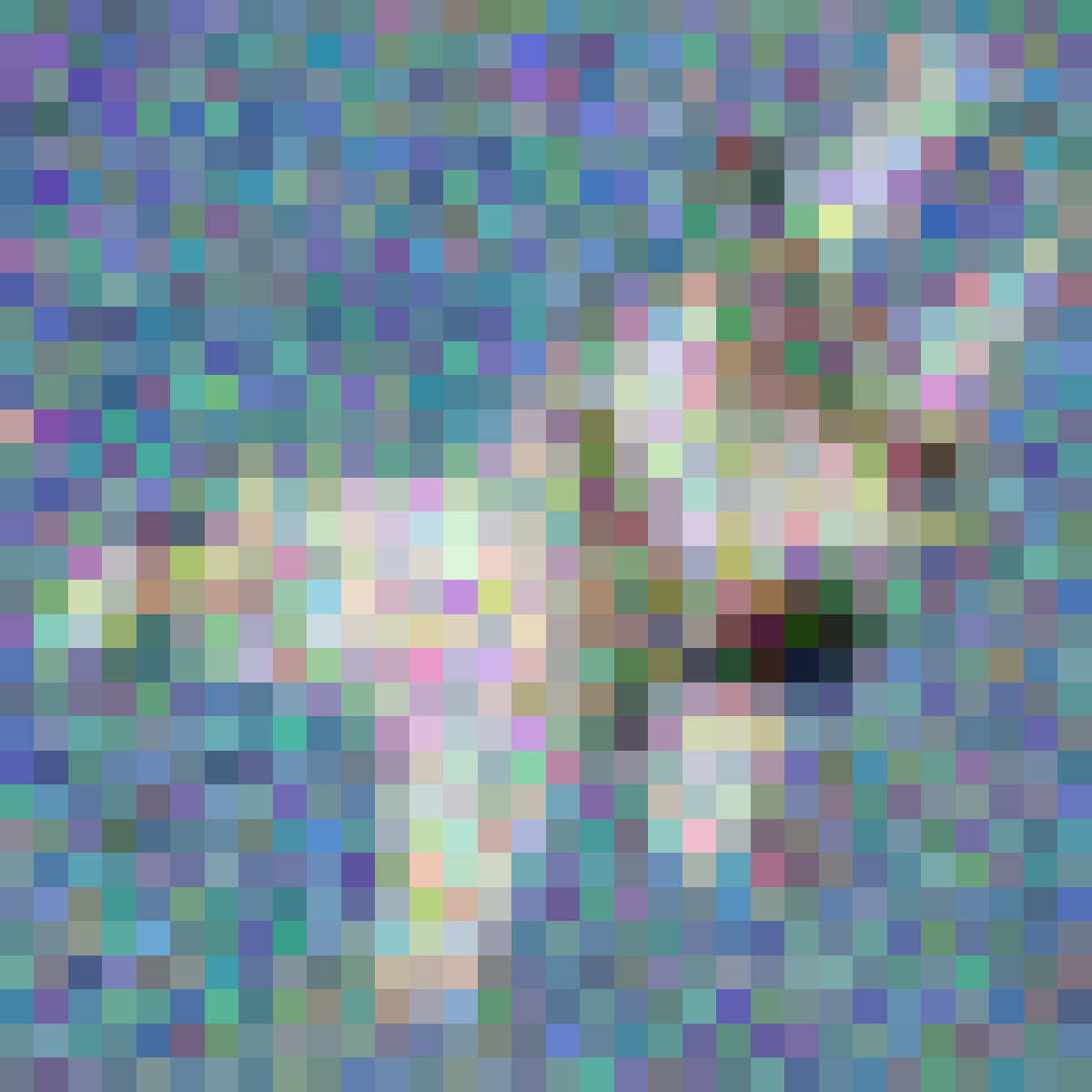}
        \caption{Correctly classified randomly perturbed aeroplane ($\delta = 10$, perturbation \\
        $\ell^1$ norm: 172.99, \\
        Euclidean norm: 3.92, \\
        $\ell^{\infty}$ norm: 0.37)
        }
    \end{subfigure}
    \hfill
    \begin{subfigure}[t]{0.23\textwidth}
        \includegraphics[width=\linewidth]{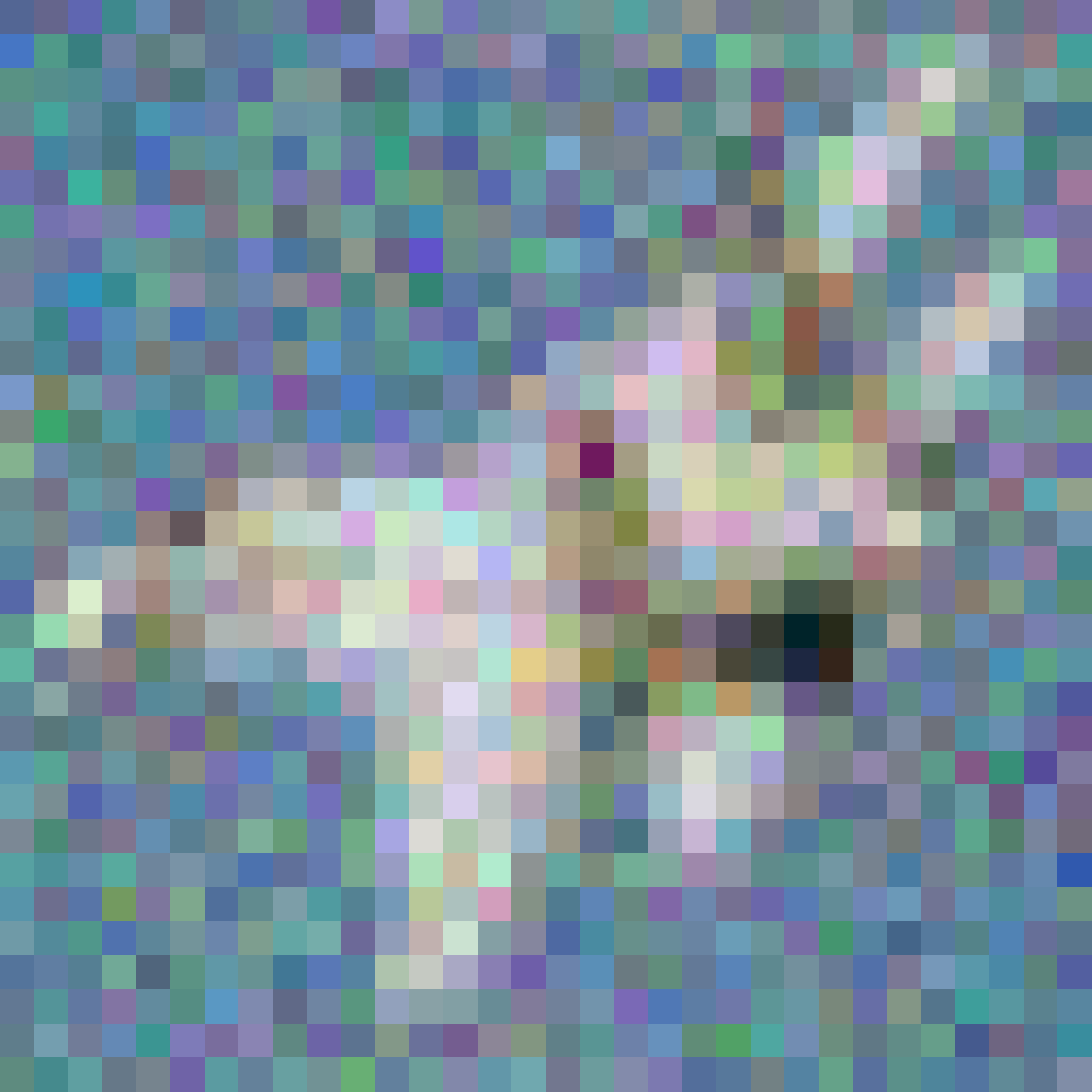}
        \caption{
        Misclassified randomly perturbed aeroplane ($\delta = 10$, perturbation \\
        $\ell^1$ norm: 174.23, \\
        Euclidean norm: 3.92, \\
        $\ell^{\infty}$ norm: 0.32)
        }
    \end{subfigure}
    \caption{CIFAR-10 --- An example of an adversarially attacked image of an aeroplane, taken from the `cat-vs-aeroplane' binary classification problem (0-vs-3), alongside examples of large random perturbations to the same image which did and did not cause the network to misclassify the image. Of the 2,000 sampled random perturbations, 4 (0.2\%) caused this image to be misclassified. Components of the modified image which were outside the range $[0, 1]$ have been clipped into the range for plotting, although not for the classification.}
    \label{fig:airplaneAttacks}
\end{figure}

The remarkable story shown by this data is that the networks are almost universally insensitive to random perturbations to the images, even when those perturbations become quite drastic.
This puzzling feature is demonstrated in Figures~\ref{fig:catAttacks} and~\ref{fig:airplaneAttacks}, where we show examples from the `aeroplane-vs-cat' binary classification problem (0-vs-3), of images of a cat and an aeroplane (original image in panel (a)) which were correctly classified by the network, alongside the same image after an adversarial attack which successfully changed the network's predicted class in panel (b).
This adversarial attack makes only a small change to the image (the largest change to any single pixel channel is 0.19 for the cat and 0.05 for the aeroplane).
When the image is perturbed using a large random perturbation, as shown in panel (c), however, the network still produces the correct classification. 
For comparison, in panel (d) we show a random perturbation which caused the network to misclassify the image.
Both of these random perturbations were obtained using $\delta = 10$ as in Section~\ref{sec:supplementary:randomPerturbationsSetup} and therefore have similar norms.
To the human eye, however, there is no significant difference between the two randomly perturbed images, or between the original image and the adversarially attacked image.
Even in these cases where a random perturbation was found which caused a misclassification, it is to be noted that only a small fraction of the 2,000 sampled random perturbations did so (4.15\% for the cat and 0.2\% for the aeroplane).

The data from Tables~\ref{tbl:randomPerturbationSuccessDelta1}--\ref{tbl:randomPerturbationSuccessDelta10} shows that as the scale of the random perturbations increases (as controlled by $\delta$), so too does the probability of causing a perturbed image to be misclassified.
In itself, this observation is unsurprising, but the data in Table~\ref{tbl:randomPerturbationSuccessDelta5} shows that, even when the random perturbations are scaled to be 5 times larger than the known adversarial attack (measured in the Euclidean norm, corresponding to $\delta = 5$), typically fewer than 5\% of images were misclassified after applying any of the random perturbations.

\begin{figure}
    \centering
    \begin{subfigure}[t]{0.9\linewidth}
        \includegraphics[width=\linewidth]{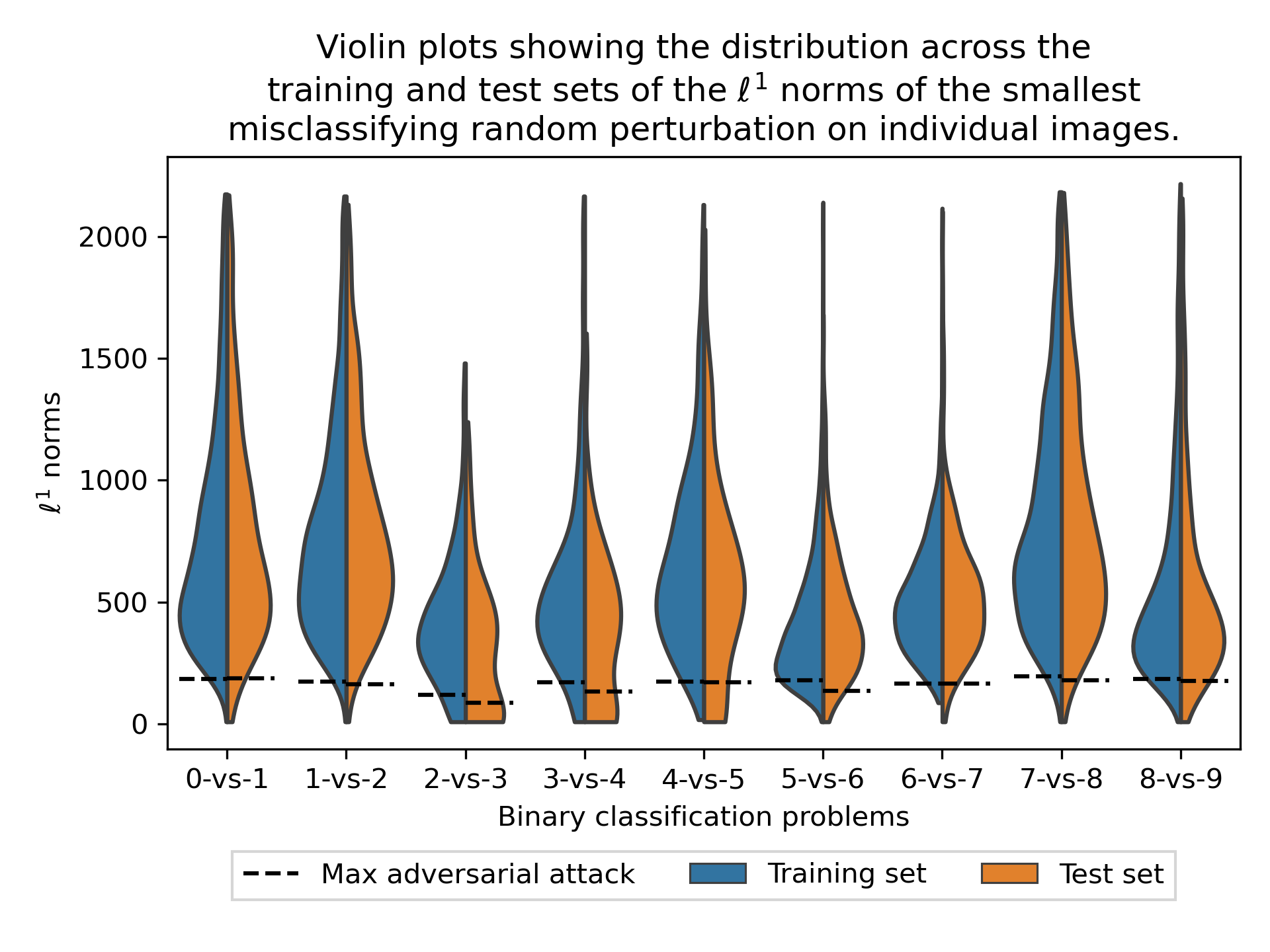}
        \caption{$\ell^1$ norms}
        \label{fig:rdpt:violin:l1}
    \end{subfigure}
    \begin{subfigure}[t]{0.9\linewidth}
        \includegraphics[width=\linewidth]{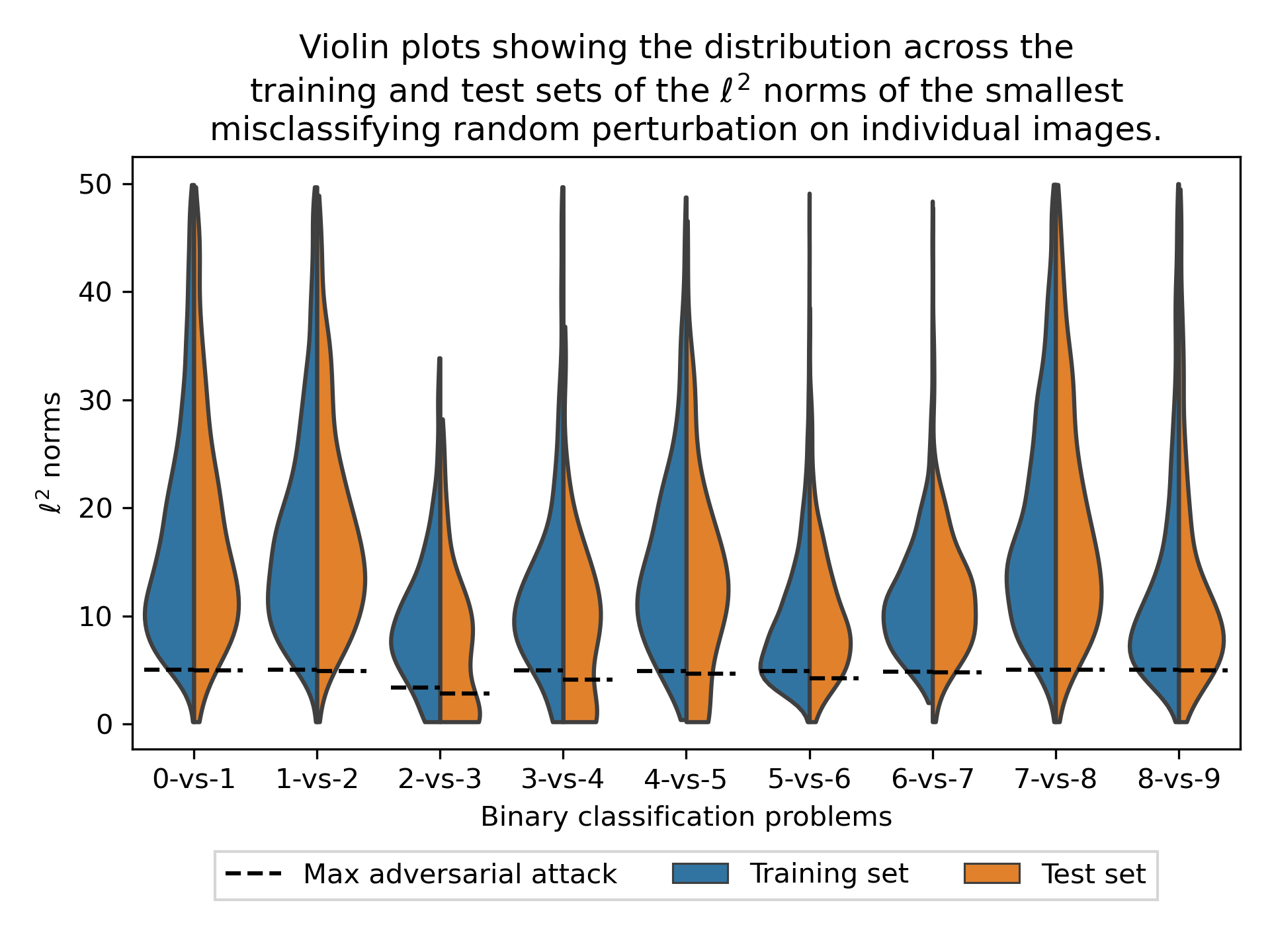}
        \caption{$\ell^2$ norms}
        \label{fig:rdpt:violin:l2}
    \end{subfigure}
    \begin{subfigure}[t]{0.9\linewidth}
        \includegraphics[width=\linewidth]{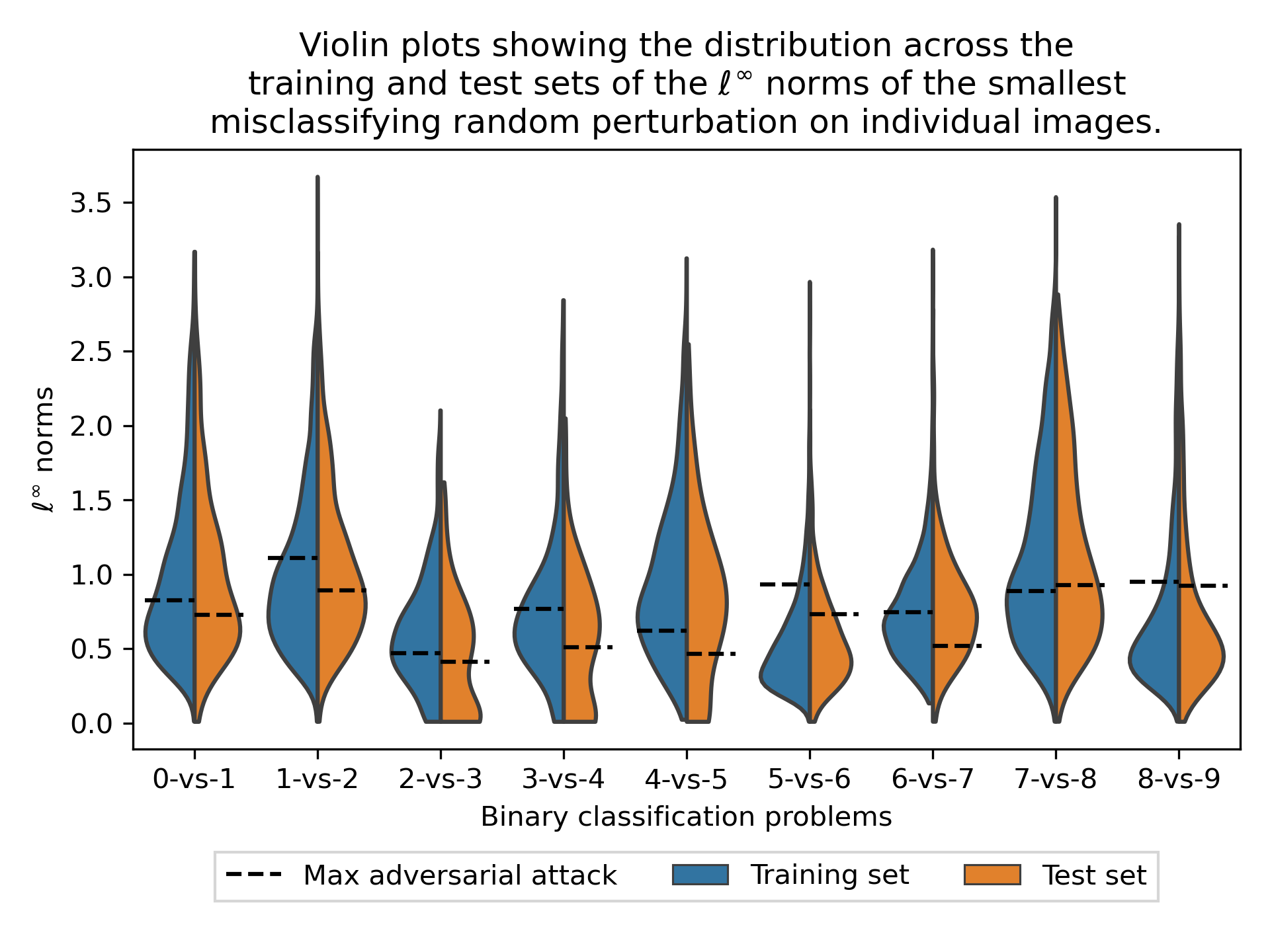}
        \caption{$\ell^\infty$ norms}
        \label{fig:rdpt:violin:linfty}
    \end{subfigure}
    \caption{CIFAR-10 --- The distribution over all images in $\mathcal{X}^{\operatorname{rand}, 10}_{i,j}$ (from the training set, see Section~\ref{sec:supplementary:randomPerturbationsSetup}) and $\mathcal{Y}^{\operatorname{rand}, 10}_{i,j}$ (from the test set) of the smallest norm of a random perturbation which caused the network to misclassify the image.
    Black dashed lines show the size of the largest adversarial attack required on each data set.
    These were fitted to the data using a standard Kernel Density Estimation algorithm and therefore only provide an approximation of the distribution.}
\end{figure}

\begin{figure}
    \centering
    \includegraphics[width=\linewidth]{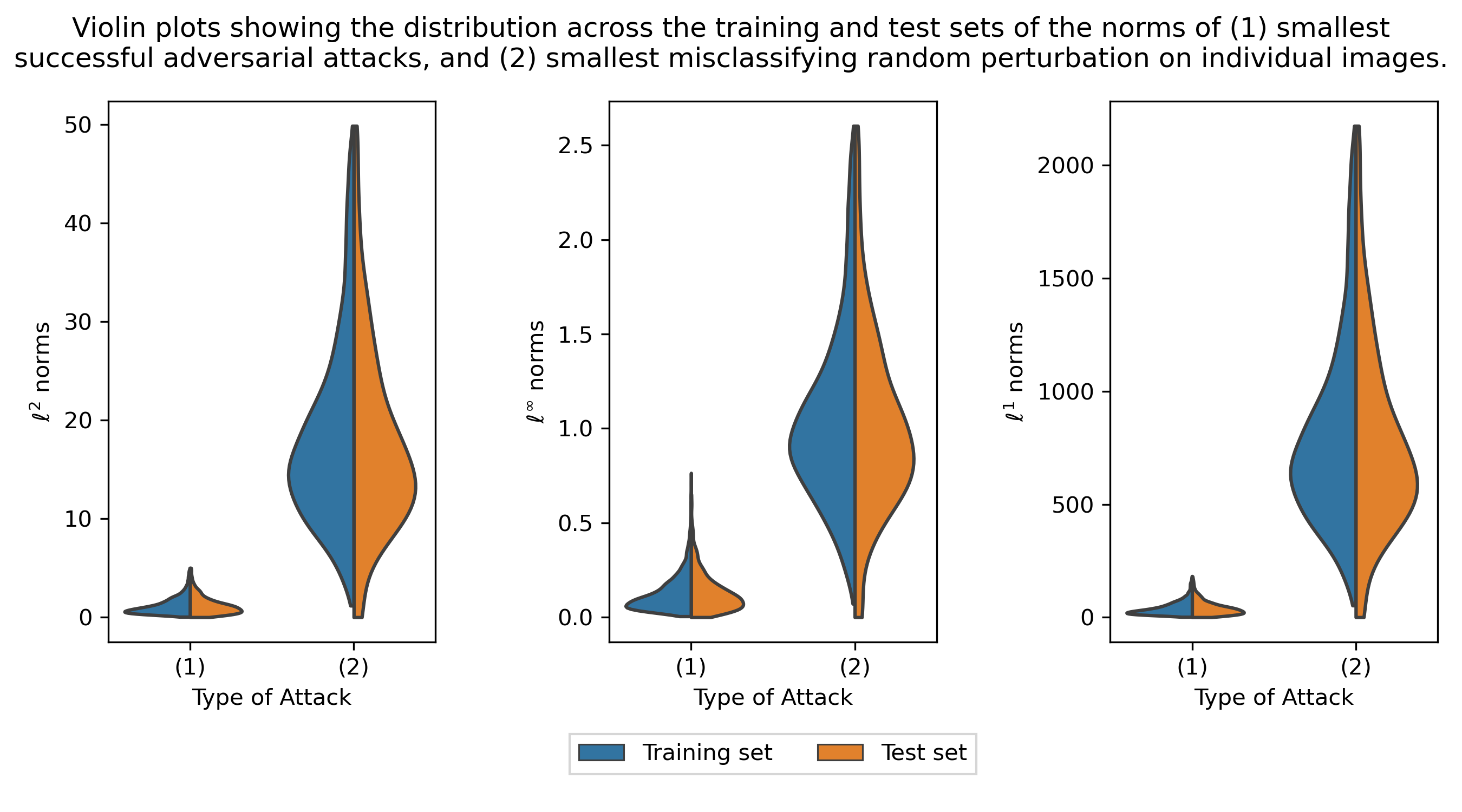}
    \caption{CIFAR-10 --- A direct comparison of the size distributions over all attackable images in the training and test sets of the smallest successful adversarial attack and smallest misclassifying random perturbation for the `cat-vs-aeroplane' problem (0-vs-3), as measured in various norms.
    The plotted distributions were fitted to the data using a standard Kernel Density Estimation algorithm and therefore only provide an approximation of the true empirical distribution.}
    \label{fig:violinComparison:0vs3}
\end{figure}

In Figures~\ref{fig:rdpt:violin:l2}--\ref{fig:rdpt:violin:l1} we show the distributions of the smallest random perturbation found to cause an image to be misclassified, as measured in the Euclidean, $\ell^{\infty}$ and $\ell^1$ norms respectively, for $\delta=10$ on a representative sample of the binary classification problems (the `$i$-vs-$(i + 1)$' problems, cf. Figures~\ref{fig:adversarial:violin:l2}--\ref{fig:adversarial:violin:l1}).
Recall that the random perturbations were sampled from a ball with Euclidean norm less than or equal to $1$ (although high dimensional concentration phenomena ensure that the all of the random perturbations have Euclidean norm very close to 1), and were scaled by $\delta$ times the Euclidean norm of the smallest successful adversarial attack when used to attack each image.
This explains the underlying similarity between these distributions and those in Figures~\ref{fig:adversarial:violin:l2}--\ref{fig:adversarial:violin:l1}, which show the size distributions of the adversarial attacks.
However, it is readily apparent here once again that significantly larger random perturbations are required as compared to adversarial perturbations.
This is visible (and shown in more detail for the `cat-vs-aeroplane' problem (0-vs-3) in Figure~\ref{fig:violinComparison:0vs3} from the fact that the random perturbation distributions appear to have much thicker tails than those for the adversarial perturbations; if simply a fixed fraction of all random perturbations were successful in causing an image to be misclassified then the distributions would shrink by a constant factor along their length.

Together, this evidence indicates that the decision surface does not pass close to the image in all directions, but rather only in one or a few specific adversarial directions.

\begin{table*}
    \centering
    \scalebox{0.8}{
    \begin{tabular}{c|ccccccccc}
    \toprule
    {} &           1 &           2 &           3 &           4 &           5 &           6 &           7 &           8 &           9 \\
    \midrule
    0 &  0, 0.06 &  0, 0.06 &  0, 0.06 &     0, 0.11 &  0, 0.12 &     0.01, 0 &  0.01, 0.12 &  0.10, 0.12 &     0.01, 0 \\
    1 &          &  0, 0.06 &     0, 0 &  0.01, 0.11 &     0, 0 &  0.04, 0.06 &        0, 0 &     0, 0.06 &     0.01, 0 \\
    2 &          &          &  0, 0.26 &  0.01, 0.06 &  0, 0.12 &        0, 0 &        0, 0 &     0, 0.06 &        0, 0 \\
    3 &          &          &          &  0.04, 0.12 &  0, 0.18 &        0, 0 &     0, 0.06 &        0, 0 &        0, 0 \\
    4 &          &          &          &             &  0, 0.06 &     0, 0.05 &        0, 0 &     0, 0.06 &  0.02, 0.06 \\
    5 &          &          &          &             &          &     0, 0.06 &     0, 0.06 &        0, 0 &        0, 0 \\
    6 &          &          &          &             &          &             &        0, 0 &        0, 0 &  0.01, 0.06 \\
    7 &          &          &          &             &          &             &             &        0, 0 &        0, 0 \\
    8 &          &          &          &             &          &             &             &             &     0.05, 0 \\
    \bottomrule
    \end{tabular}
    }
    \caption{CIFAR-10 --- Susceptibility of the networks to random perturbations, as described in Section~\ref{sec:supplementary:randomPerturbationsSetup} for $\delta=1$. This is reported in the form `train susceptibility, test susceptibility', where susceptibility is calculated as in~\eqref{eq:supplementary:randomSusceptibility} as the percentage of adversarially attackable images from each set which were misclassified after applying any of the 2,000 random perturbations. The row and column headers indicate the classes used in each binary classification problem. Here, we use 0 without any trailing decimal places to indicate a value which was actually zero, and not simply rounded to zero when rounding to two decimal places.}
    \label{tbl:randomPerturbationSuccessDelta1}
\end{table*}

\begin{table*}
    \centering
    \scalebox{0.8}{
    \begin{tabular}{c|ccccccccc}
    \toprule
    {} &           1 &           2 &           3 &           4 &           5 &           6 &           7 &           8 &           9 \\
    \midrule
    0 &  0.17, 0.42 &  0.02, 0.06 &     0, 0.06 &  0.07, 0.17 &  0.04, 0.31 &     0.01, 0 &  0.04, 0.29 &  1.11, 1.33 &  0.89, 0.78 \\
1 &             &  0.02, 0.06 &     0.01, 0 &  0.03, 0.27 &        0, 0 &  0.05, 0.06 &        0, 0 &  0.59, 0.81 &  0.06, 0.36 \\
2 &             &             &  0.02, 0.26 &  0.02, 0.12 &  0.06, 0.24 &  0.07, 0.06 &  0.01, 0.12 &     0, 0.19 &        0, 0 \\
3 &             &             &             &  0.11, 0.31 &  0.01, 0.24 &     0.01, 0 &     0, 0.06 &     0.01, 0 &        0, 0 \\
4 &             &             &             &             &     0, 0.17 &  0.03, 0.05 &     0, 0.06 &  0.05, 0.06 &  0.03, 0.11 \\
5 &             &             &             &             &             &  0.03, 0.06 &     0, 0.11 &     0.01, 0 &  0.03, 0.07 \\
6 &             &             &             &             &             &             &        0, 0 &     0.03, 0 &  0.01, 0.06 \\
7 &             &             &             &             &             &             &             &        0, 0 &     0.04, 0 \\
8 &             &             &             &             &             &             &             &             &  0.97, 1.47 \\
    \bottomrule
    \end{tabular}
    }
    \caption{CIFAR-10 --- Susceptibility of the networks to random perturbations, as described in Section~\ref{sec:supplementary:randomPerturbationsSetup} for $\delta=2$. This is reported in the form `train susceptibility, test susceptibility', where susceptibility is calculated as in~\eqref{eq:supplementary:randomSusceptibility} as the percentage of adversarially attackable images from each set which were misclassified after applying any of the 2,000 random perturbations. The row and column headers indicate the classes used in each binary classification problem. Here, we use 0 without any trailing decimal places to indicate a value which was actually zero, and not simply rounded to zero when rounding to two decimal places.}
    \label{tbl:randomPerturbationSuccessDelta2}
\end{table*}

\begin{table*}
    \centering
    \scalebox{0.8}{
    \begin{tabular}{c|ccccccccc}
    \toprule
    {} &           1 &           2 &           3 &           4 &           5 &           6 &           7 &           8 &           9 \\
    \midrule
    0 &  9.67, 10.56 &  1.55, 2.38 &  3.87, 4.58 &  9.94, 10.30 &  4.06, 4.93 &  1.75, 1.85 &  7.52, 8.42 &    8.93, 9.43 &  14.85, 16.57 \\
    1 &              &  7.61, 7.33 &  0.58, 0.25 &   5.99, 6.20 &  1.04, 1.27 &  1.11, 1.14 &  3.67, 3.12 &  11.06, 11.57 &    0.53, 0.75 \\
    2 &              &             &  0.12, 0.85 &   0.27, 0.40 &  0.61, 1.31 &  2.02, 2.46 &  1.65, 1.56 &    0.32, 0.56 &    2.30, 2.14 \\
    3 &              &             &             &   0.53, 0.99 &  0.25, 1.03 &  0.63, 0.63 &  0.23, 0.33 &    2.42, 2.94 &    0.34, 0.40 \\
    4 &              &             &             &              &  0.11, 0.56 &  0.44, 0.59 &  0.18, 0.56 &    3.46, 3.39 &    3.83, 3.96 \\
    5 &              &             &             &              &             &  2.97, 4.91 &  0.11, 0.33 &    2.36, 2.66 &    9.51, 9.74 \\
    6 &              &             &             &              &             &             &  0, 0.10 &    2.86, 3.24 &    0.07, 0.23 \\
    7 &              &             &             &              &             &             &             &    2.65, 2.57 &    1.40, 1.48 \\
    8 &              &             &             &              &             &             &             &               &  10.37, 10.37 \\
    \bottomrule
    \end{tabular}
    }
    \caption{CIFAR-10 --- Susceptibility of the networks to random perturbations, as described in Section~\ref{sec:supplementary:randomPerturbationsSetup} for $\delta=5$. This is reported in the form `train susceptibility, test susceptibility', where susceptibility is calculated as in~\eqref{eq:supplementary:randomSusceptibility} as the percentage of adversarially attackable images from each set which were misclassified after applying any of the 2,000 random perturbations. The row and column headers indicate the classes used in each binary classification problem. Here, we use 0 without any trailing decimal places to indicate a value which was actually zero, and not simply rounded to zero when rounding to two decimal places.}
    \label{tbl:randomPerturbationSuccessDelta5}
\end{table*}

\begin{table*}
    \centering
    \scalebox{0.8}{
    \begin{tabular}{c|ccccccccc}
    \toprule
    {} &           1 &           2 &           3 &           4 &           5 &           6 &           7 &           8 &           9 \\
    \midrule
    0 &  41.49, 40.57 &  14.02, 16.71 &  29.34, 32.15 &  47.37, 49.55 &  27.49, 30.07 &  27.92, 28.11 &  50.40, 52.41 &  41.44, 42.16 &  50.82, 49.72 \\
    1 &               &  41.82, 42.67 &  29.38, 28.98 &  27.18, 28.96 &  41.01, 43.16 &  15.90, 17.79 &  30.90, 29.43 &  36.98, 36.92 &  29.42, 31.09 \\
    2 &               &               &    4.76, 6.78 &  16.74, 18.75 &   9.12, 10.48 &  34.10, 34.42 &  22.70, 23.11 &  19.20, 20.08 &  32.67, 31.80 \\
    3 &               &               &               &    5.54, 6.58 &   9.57, 11.04 &  19.86, 19.24 &    6.99, 8.49 &  28.58, 29.07 &  26.54, 28.01 \\
    4 &               &               &               &               &    5.59, 6.40 &  14.37, 14.77 &    3.87, 4.73 &  38.19, 39.58 &  31.07, 33.56 \\
    5 &               &               &               &               &               &  43.20, 42.81 &    6.99, 9.00 &  22.66, 23.23 &  56.10, 58.79 \\
    6 &               &               &               &               &               &               &  16.45, 16.70 &  26.87, 30.53 &    8.64, 8.92 \\
    7 &               &               &               &               &               &               &               &  42.90, 44.36 &  26.44, 26.14 \\
    8 &               &               &               &               &               &               &               &               &  45.04, 45.08 \\
    \bottomrule
    \end{tabular}
    }
    \caption{CIFAR-10 --- Susceptibility of the networks to random perturbations, as described in Section~\ref{sec:supplementary:randomPerturbationsSetup} for $\delta=10$. This is reported in the form `train susceptibility, test susceptibility', where susceptibility is calculated as in~\eqref{eq:supplementary:randomSusceptibility} as the percentage of adversarially attackable images from each set which were misclassified after applying any of the 2,000 random perturbations. The row and column headers indicate the classes used in each binary classification problem.}
    \label{tbl:randomPerturbationSuccessDelta10}
\end{table*}

\begin{figure*}
    \includegraphics[width=\textwidth]{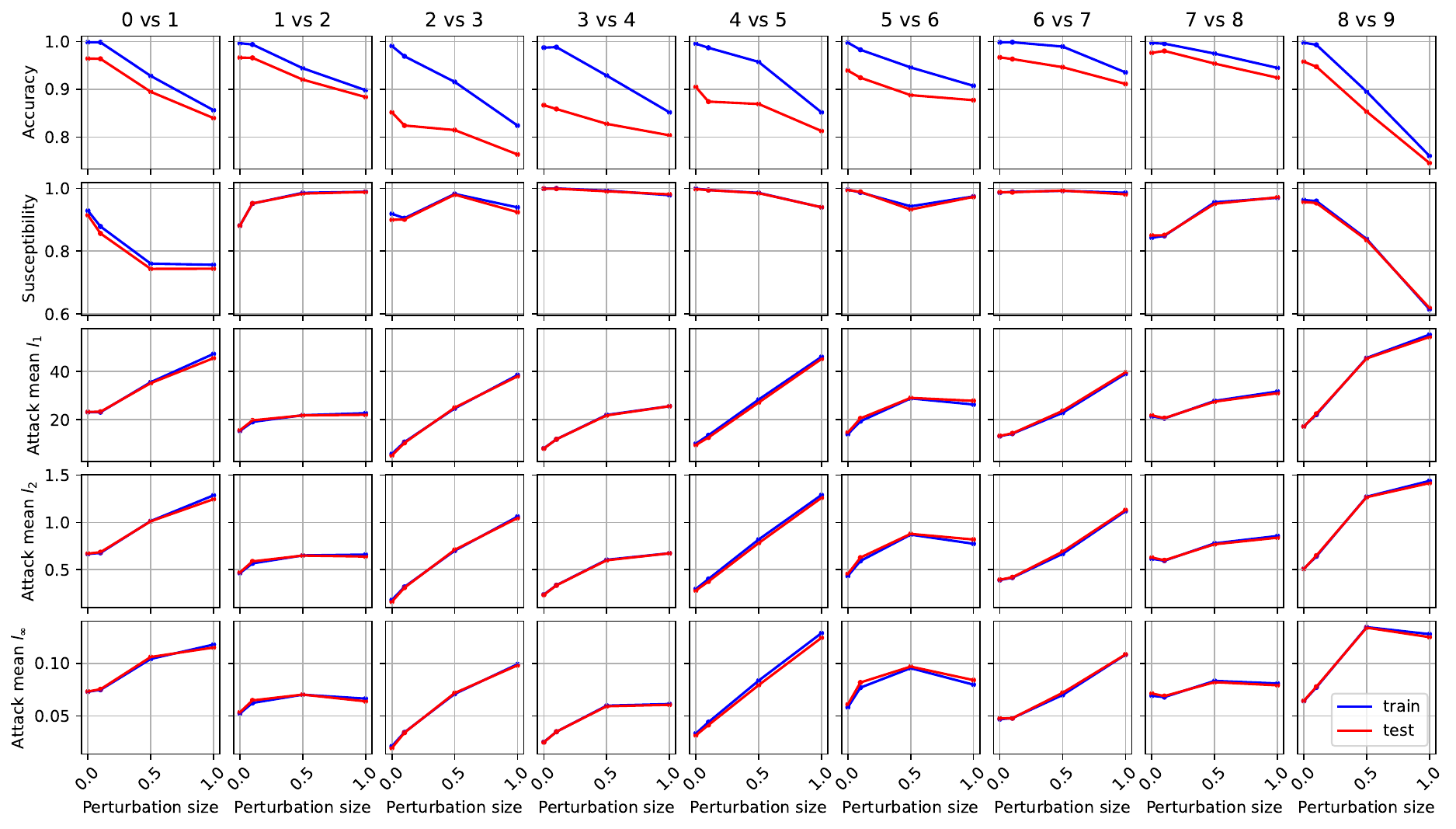}
    \caption{CIFAR-10 --- Plots showing how the performance of the network is affected by various magnitudes of random perturbations added to the images during training. This figure shows the results for random perturbations sampled from the cube $[-a, a]^n$ for $a \in \{0, 0.1, 0.5, 1\}$.
    This visualises the results in in Tables~\ref{tbl:randomTraining:linf:0.1}, \ref{tbl:randomTraining:linf:0.5} and \ref{tbl:randomTraining:linf:1.0} compared to the previous data computed with no random perturbations (corresponding to $a = 0$). The data is plotted as separate lines for the training and test sets. `Susceptibility' here refers to the adversarial susceptibility reported in the tables, and `Attack mean $\ell_p$' indicates the mean across each data set of the $\ell^p$ norm of the smallest adversarial perturbation affecting each image. The perturbation size plotted on the $x$ axis is the size of $a$.}
    \label{fig:randomTraining:cubePerturbations}
\end{figure*}

\begin{figure*}
    \includegraphics[width=\textwidth]{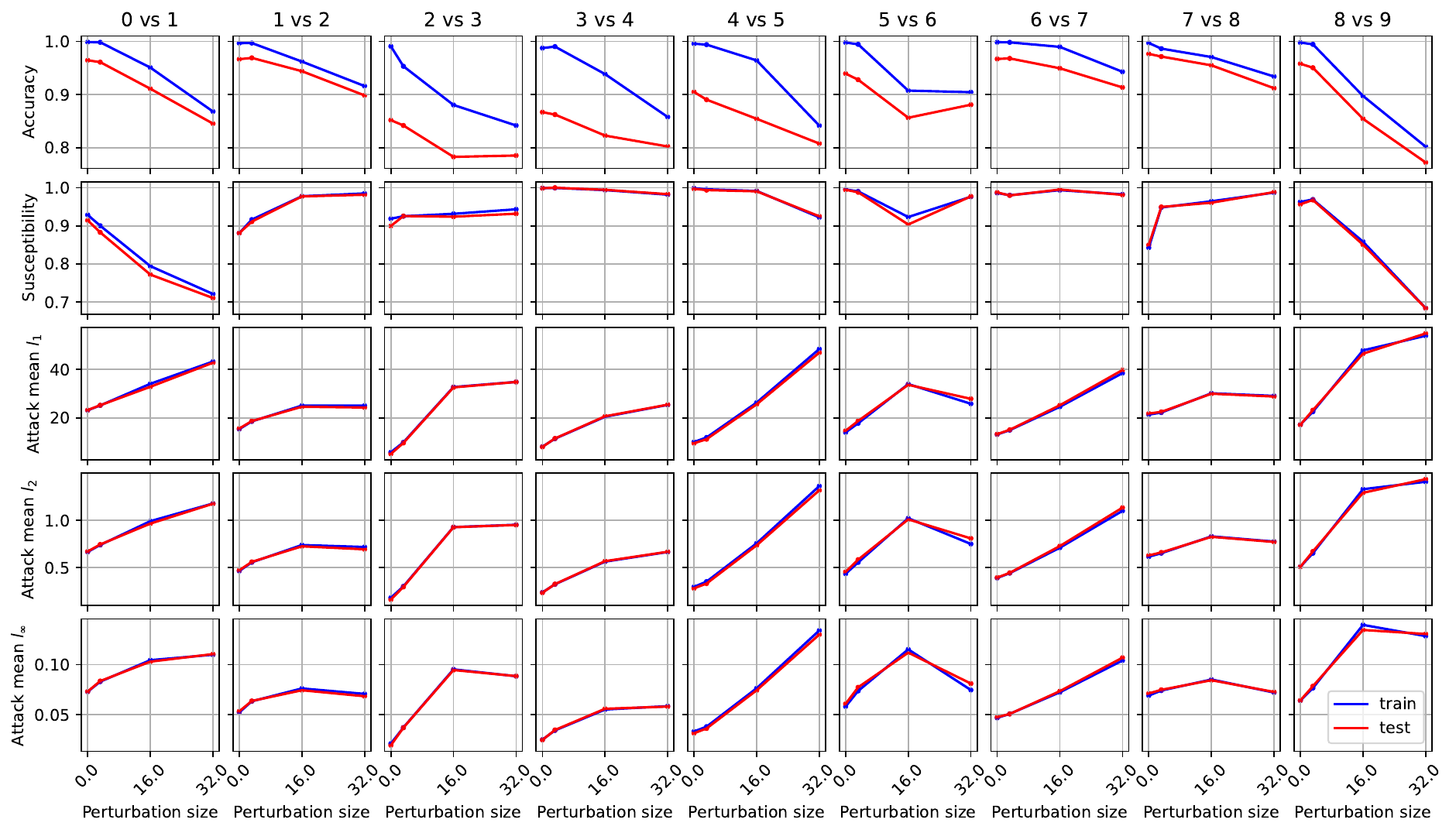}
    \caption{CIFAR-10 --- Plots showing how the performance of the network is affected by various magnitudes of random perturbations added to the images during training. This figure shows the results for random perturbations sampled from the ball $\mathbb{B}^n_b$ for $b \in \{0, 3.2, 16, 32\}$.
    This visualises the results in in Tables~\ref{tbl:randomTraining:linf:0.1}, \ref{tbl:randomTraining:linf:0.5} and \ref{tbl:randomTraining:linf:1.0} compared to the previous data computed with no random perturbations (corresponding to $b = 0$). The data is plotted as separate lines for the training and test sets. `Susceptibility' here refers to the adversarial susceptibility reported in the tables, and `Attack mean $\ell_p$' indicates the mean across each data set of the $\ell^p$ norm of the smallest adversarial perturbation affecting each image. The perturbation size plotted on the $x$ axis is the size of $a$. }
    \label{fig:randomTraining:ballPerturbations}
\end{figure*}

\paragraph{Training with random perturbations.}
For brevity, we only report the results of these experiments for the representative subset of `class $i$-vs-class $i+1$' binary classification problems.
These are given in Tables~\ref{tbl:randomTraining:linf:0.1}, \ref{tbl:randomTraining:linf:0.5} and \ref{tbl:randomTraining:linf:1.0} for random perturbations sampled uniformly from the cube $[-a, a]^n$ with $a \in \{0.1, 0.5, 1.0\}$ respectively, and Tables~\ref{tbl:randomTraining:l2:3.2}, \ref{tbl:randomTraining:l2:16} and \ref{tbl:randomTraining:l2:32} for noise sampled uniformly from the ball $\mathbb{B}^n_b$ with $b \in \{3.2, 16, 32\}$ respectively (see Section~\ref{sec:supplementary:randomTraining} for details of the experimental setup).
These results are also plotted against the size of the sampled perturbations in Figures~\ref{fig:randomTraining:cubePerturbations} and \ref{fig:randomTraining:ballPerturbations} for perturbations sampled from the cube and ball respectively.
From these results, it is clear that additive random noise has little impact on the adversarial susceptibility of the networks, and large perturbations cause the networks' accuracy to decrease.

It should be stressed that additive noise sampled from the cube $[-1, 1]$ (the largest cube we tested) represents a significant modification to an image where each pixel value is in $[0, 1]$.
We also recall that the adversarial susceptibility is only calculated as the fraction of correctly classified images which are susceptible to adversarial attacks, meaning that the drop in accuracy of the classifier is implicitly decreasing the pool of images which were tested for adversarial attacks.
Interestingly, the average norm of the successful adversarial attacks does seem to increase with the size of the random perturbations applied during training.
However, this could once again be due to the observed drop in accuracy: training and test points which were near the decision boundary of the original classifier trained without perturbations would be those which were susceptible to the smallest adversarial attacks.
These would also be the points which would be most likely to be misclassified by the less accurate classifiers trained with randomly perturbed data, so would not be included when the adversarial attacks were computed.
Consequently, while large additive random noise may eliminate some of the smallest adversarial attacks, it does so at the expense of a significant drop in accuracy.

\begin{table*}
    \centering
    \scalebox{0.675}{
    \begin{tabular}{rr|ccccccccc}
    \toprule
    {} &    {} &         0 vs 1 &         1 vs 2 &         2 vs 3 &         3 vs 4 &         4 vs 5 &         5 vs 6 &         6 vs 7 &         7 vs 8 &         8 vs 9 \\
    \midrule
    Accuracy                   &  train &          99.85 &          99.39 &          96.93 &          98.85 &          98.71 &          98.29 &          99.87 &          99.57 &          99.33 \\
                               &   test &          96.40 &          96.60 &          82.45 &          85.90 &          87.45 &          92.45 &          96.35 &          98.05 &          94.75 \\
    Adv. susceptibility             &  train &          87.85 &          95.14 &          90.49 &          99.96 &          99.46 &          98.67 &          98.87 &          84.85 &          95.94 \\
                               &   test &          85.63 &          95.19 &          90.12 &          99.88 &          99.43 &          98.86 &          98.75 &          85.06 &          95.30 \\
    Adv. attack $\ell^1$ norm     &  train &  23.14 (35.17) &  19.11 (27.94) &  10.78 (17.35) &  11.94 (16.84) &  13.57 (18.69) &  19.38 (26.04) &  14.23 (19.03) &  20.55 (33.22) &  22.12 (32.41) \\
                               &   test &  23.33 (36.73) &  19.75 (29.04) &  10.42 (17.82) &  11.92 (18.32) &  12.66 (18.25) &  20.58 (27.78) &  14.39 (19.51) &  20.68 (33.56) &  22.46 (33.46) \\
    Adv. attack $\ell^2$ norm     &  train &    0.68 (1.04) &    0.57 (0.83) &    0.32 (0.52) &    0.34 (0.47) &    0.40 (0.56) &    0.59 (0.79) &    0.42 (0.55) &    0.60 (0.96) &    0.64 (0.94) \\
                               &   test &    0.69 (1.08) &    0.59 (0.87) &    0.31 (0.54) &    0.34 (0.51) &    0.38 (0.54) &    0.63 (0.84) &    0.42 (0.57) &    0.60 (0.97) &    0.65 (0.97) \\
    Adv. attack $\ell^\infty$ norm &  train &    0.07 (0.12) &    0.06 (0.09) &    0.03 (0.06) &    0.04 (0.05) &    0.04 (0.06) &    0.08 (0.11) &    0.05 (0.06) &    0.07 (0.11) &    0.08 (0.11) \\
                               &   test &    0.08 (0.12) &    0.06 (0.10) &    0.03 (0.06) &    0.04 (0.06) &    0.04 (0.06) &    0.08 (0.11) &    0.05 (0.06) &    0.07 (0.11) &    0.08 (0.12) \\
\bottomrule
    \end{tabular}
    }
    \caption{CIFAR-10 --- Performance results when images are randomly perturbed during training using additive random noise sampled from the cube $[-a, a]^n$ with $a = 0.1$. The abbreviation `Adv.' should be read as `Adversarial'. The quantities computed are defined in Section~\ref{sec:experimentalSetup}. Accuracy and susceptibility are reported as percentages. The norms of the adversarial attacks are reported in the form `mean (standard deviation)', calculated by averaging over all of the correctly classified and adversarially susceptible images in each of the training and test sets.}
    \label{tbl:randomTraining:linf:0.1}
\end{table*}

\begin{table*}
    \centering
    \scalebox{0.675}{
    \begin{tabular}{rr|ccccccccc}
    \toprule
    {} &    {} &         0 vs 1 &         1 vs 2 &         2 vs 3 &         3 vs 4 &         4 vs 5 &         5 vs 6 &         6 vs 7 &         7 vs 8 &         8 vs 9 \\
    \midrule
    Accuracy                   &  train &          99.83 &          99.69 &         95.33 &          99.02 &          99.39 &          99.45 &          99.82 &          98.64 &          99.45 \\
                               &   test &          96.10 &          96.90 &         84.20 &          86.25 &          89.05 &          92.80 &          96.80 &          97.15 &          95.05 \\
    Adv. susceptibility             &  train &          90.03 &          91.64 &         92.53 &          99.95 &          99.59 &          99.03 &          98.04 &          94.83 &          96.92 \\
                               &   test &          88.29 &          91.12 &         92.58 &         100.00 &          99.38 &          98.81 &          97.99 &          94.96 &          96.79 \\
    Adv. attack $\ell^1$ norm     &  train &  25.16 (36.97) &  18.58 (27.80) &  9.94 (17.17) &  11.45 (16.26) &  11.95 (16.51) &  17.81 (24.96) &  14.97 (20.01) &  22.23 (33.69) &  22.58 (32.08) \\
                               &   test &  25.23 (37.82) &  18.71 (28.24) &  9.69 (16.88) &  11.55 (18.15) &  11.25 (16.89) &  18.76 (25.94) &  15.12 (21.37) &  22.50 (34.12) &  23.19 (33.62) \\
    Adv. attack $\ell^2$ norm     &  train &    0.74 (1.10) &    0.56 (0.83) &   0.30 (0.53) &    0.32 (0.46) &    0.35 (0.49) &    0.56 (0.78) &    0.44 (0.59) &    0.65 (0.99) &    0.65 (0.93) \\
                               &   test &    0.74 (1.12) &    0.56 (0.84) &   0.30 (0.52) &    0.33 (0.51) &    0.33 (0.50) &    0.58 (0.80) &    0.44 (0.62) &    0.66 (1.00) &    0.67 (0.98) \\
    Adv. attack $\ell^\infty$ norm &  train &    0.08 (0.13) &    0.06 (0.10) &   0.04 (0.07) &    0.03 (0.05) &    0.04 (0.06) &    0.07 (0.11) &    0.05 (0.07) &    0.07 (0.12) &    0.08 (0.11) \\
                               &   test &    0.08 (0.13) &    0.06 (0.10) &   0.04 (0.07) &    0.03 (0.06) &    0.04 (0.06) &    0.08 (0.11) &    0.05 (0.07) &    0.07 (0.12) &    0.08 (0.12) \\
\bottomrule
    \end{tabular}
    }
    \caption{CIFAR-10 --- Performance results when images are randomly perturbed during training using additive random noise sampled from the ball $\mathbb{B}^n_b$ with $b = 3.2$. The abbreviation `Adv.' should be read as `Adversarial'. The quantities computed are defined in Section~\ref{sec:experimentalSetup}. Accuracy and susceptibility are reported as percentages. The norms of the adversarial attacks are reported in the form `mean (standard deviation)', calculated by averaging over all of the correctly classified and adversarially susceptible images in each of the training and test sets.}
    \label{tbl:randomTraining:l2:3.2}
\end{table*}

\begin{table*}
    \centering
    \scalebox{0.675}{
    \begin{tabular}{rr|ccccccccc}
    \toprule
    {} &    {} &         0 vs 1 &         1 vs 2 &         2 vs 3 &         3 vs 4 &         4 vs 5 &         5 vs 6 &         6 vs 7 &         7 vs 8 &         8 vs 9 \\
    \midrule
    Accuracy                   &  train &          92.82 &          94.43 &          91.57 &          92.89 &          95.75 &          94.60 &          98.96 &          97.50 &          89.54 \\
                               &   test &          89.50 &          92.05 &          81.50 &          82.80 &          86.95 &          88.80 &          94.65 &          95.40 &          85.35 \\
    Adv. susceptibility        &  train &          76.02 &          98.56 &          98.21 &          99.26 &          98.55 &          94.27 &          99.20 &          95.57 &          83.88 \\
                               &   test &          74.36 &          98.32 &          97.98 &          99.03 &          98.45 &          93.24 &          99.26 &          95.07 &          83.54 \\
    Adv. attack $\ell^1$ norm     &  train &  35.56 (50.17) &  21.88 (33.12) &  24.70 (34.77) &  22.03 (33.51) &  28.38 (35.91) &  28.85 (37.80) &  22.88 (29.34) &  27.83 (39.89) &  45.61 (55.12) \\
                               &   test &  35.17 (50.80) &  21.82 (32.95) &  25.02 (36.26) &  21.81 (33.94) &  27.13 (35.90) &  29.06 (38.87) &  23.70 (31.22) &  27.50 (39.63) &  45.39 (54.85) \\
    Adv. attack $\ell^2$ norm     &  train &    1.01 (1.44) &    0.65 (0.98) &    0.70 (1.00) &    0.61 (0.91) &    0.82 (1.04) &    0.87 (1.13) &    0.67 (0.85) &    0.78 (1.11) &    1.27 (1.53) \\
                               &   test &    1.01 (1.47) &    0.65 (0.98) &    0.71 (1.04) &    0.60 (0.93) &    0.78 (1.04) &    0.88 (1.17) &    0.69 (0.91) &    0.77 (1.10) &    1.27 (1.53) \\
    Adv. attack $\ell^\infty$ norm &  train &    0.10 (0.15) &    0.07 (0.11) &    0.07 (0.10) &    0.06 (0.09) &    0.08 (0.11) &    0.10 (0.13) &    0.07 (0.09) &    0.08 (0.12) &    0.13 (0.17) \\
                               &   test &    0.11 (0.16) &    0.07 (0.11) &    0.07 (0.11) &    0.06 (0.09) &    0.08 (0.11) &    0.10 (0.13) &    0.07 (0.10) &    0.08 (0.12) &    0.13 (0.17) \\
    \bottomrule
    \end{tabular}
    }
    \caption{CIFAR-10 --- Performance results when images are randomly perturbed during training using additive random noise sampled from the cube $[-a, a]^n$ with $a = 0.5$. The abbreviation `Adv.' should be read as `Adversarial'. The quantities computed are defined in Section~\ref{sec:experimentalSetup}. Accuracy and susceptibility are reported as percentages. The norms of the adversarial attacks are reported in the form `mean (standard deviation)', calculated by averaging over all of the correctly classified and adversarially susceptible images in each of the training and test sets.}
    \label{tbl:randomTraining:linf:0.5}
\end{table*}

\begin{table*}
    \centering
    \scalebox{0.675}{
    \begin{tabular}{rr|ccccccccc}
    \toprule
    {} &    {} &         0 vs 1 &         1 vs 2 &         2 vs 3 &         3 vs 4 &         4 vs 5 &         5 vs 6 &         6 vs 7 &         7 vs 8 &         8 vs 9 \\
    \midrule
    Accuracy                   &  train &          95.08 &          96.21 &          88.05 &          93.85 &          96.42 &          90.75 &          98.97 &          97.06 &          89.75 \\
                               &   test &          91.10 &          94.40 &          78.30 &          82.30 &          85.45 &          85.65 &          94.95 &          95.50 &          85.45 \\
    Adv. susceptibility        &  train &          79.44 &          97.78 &          93.16 &          99.38 &          99.16 &          92.32 &          99.36 &          96.46 &          85.86 \\
                               &   test &          77.22 &          97.72 &          92.40 &          99.45 &          99.06 &          90.37 &          99.53 &          96.07 &          85.08 \\
    Adv. attack $\ell^1$ norm     &  train &  34.08 (47.83) &  25.09 (36.56) &  32.75 (41.78) &  20.49 (31.84) &  26.35 (34.00) &  33.95 (41.52) &  24.59 (31.05) &  30.17 (42.56) &  47.84 (56.68) \\
                               &   test &  32.84 (46.83) &  24.62 (36.20) &  32.60 (42.07) &  20.69 (33.00) &  25.62 (34.86) &  33.66 (42.43) &  25.21 (32.42) &  29.99 (42.89) &  46.48 (55.40) \\
    Adv. attack $\ell^2$ norm     &  train &    0.99 (1.40) &    0.74 (1.07) &    0.93 (1.19) &    0.56 (0.87) &    0.76 (0.98) &    1.02 (1.24) &    0.71 (0.89) &    0.83 (1.17) &    1.33 (1.57) \\
                               &   test &    0.97 (1.39) &    0.73 (1.06) &    0.93 (1.20) &    0.57 (0.90) &    0.74 (1.00) &    1.01 (1.26) &    0.73 (0.93) &    0.83 (1.17) &    1.29 (1.54) \\
    Adv. attack $\ell^\infty$ norm &  train &    0.10 (0.15) &    0.08 (0.11) &    0.10 (0.13) &    0.06 (0.09) &    0.08 (0.10) &    0.12 (0.14) &    0.07 (0.09) &    0.09 (0.12) &    0.14 (0.17) \\
                               &   test &    0.10 (0.15) &    0.07 (0.11) &    0.09 (0.13) &    0.06 (0.09) &    0.07 (0.10) &    0.11 (0.14) &    0.07 (0.10) &    0.08 (0.12) &    0.13 (0.16) \\
    \bottomrule
    \end{tabular}
    }
    \caption{CIFAR-10 --- Performance results when images are randomly perturbed during training using additive random noise sampled from the ball $\mathbb{B}^n_b$ with $b = 16$. The abbreviation `Adv.' should be read as `Adversarial'. The quantities computed are defined in Section~\ref{sec:experimentalSetup}. Accuracy and susceptibility are reported as percentages. The norms of the adversarial attacks are reported in the form `mean (standard deviation)', calculated by averaging over all of the correctly classified and adversarially susceptible images in each of the training and test sets.}
    \label{tbl:randomTraining:l2:16}
\end{table*}

\begin{table*}
    \centering
    \scalebox{0.675}{
    \begin{tabular}{rr|ccccccccc}
    \toprule
    {} &    {} &         0 vs 1 &         1 vs 2 &         2 vs 3 &         3 vs 4 &         4 vs 5 &         5 vs 6 &         6 vs 7 &         7 vs 8 &         8 vs 9 \\
    \midrule
    Accuracy                   &  train &          85.65 &          89.84 &          82.44 &          85.20 &          85.20 &          90.75 &          93.56 &          94.50 &          76.09 \\
                               &   test &          84.00 &          88.40 &          76.40 &          80.40 &          81.30 &          87.75 &          91.15 &          92.45 &          74.60 \\
    Adv. susceptibility        &  train &          75.66 &          98.90 &          93.92 &          97.79 &          93.96 &          97.41 &          98.57 &          96.96 &          61.51 \\
                               &   test &          74.40 &          98.81 &          92.41 &          98.07 &          93.97 &          97.26 &          98.08 &          97.13 &          61.93 \\
    Adv. attack $\ell^1$ norm     &  train &  47.29 (60.24) &  22.74 (35.90) &  38.53 (49.56) &  25.62 (41.16) &  46.07 (52.80) &  26.32 (39.04) &  39.02 (45.31) &  31.71 (44.80) &  55.25 (68.60) \\
                               &   test &  45.52 (58.87) &  22.06 (35.82) &  38.01 (49.77) &  25.61 (41.41) &  45.18 (52.78) &  27.85 (41.27) &  39.61 (46.00) &  31.00 (44.60) &  54.28 (67.73) \\
    Adv. attack $\ell^2$ norm     &  train &    1.29 (1.64) &    0.66 (1.04) &    1.06 (1.36) &    0.67 (1.08) &    1.29 (1.48) &    0.77 (1.14) &    1.12 (1.30) &    0.86 (1.21) &    1.44 (1.78) \\
                               &   test &    1.25 (1.61) &    0.64 (1.03) &    1.05 (1.36) &    0.67 (1.08) &    1.26 (1.47) &    0.82 (1.21) &    1.13 (1.31) &    0.84 (1.21) &    1.41 (1.76) \\
    Adv. attack $\ell^\infty$ norm &  train &    0.12 (0.15) &    0.07 (0.11) &    0.10 (0.13) &    0.06 (0.10) &    0.13 (0.15) &    0.08 (0.12) &    0.11 (0.13) &    0.08 (0.12) &    0.13 (0.16) \\
                               &   test &    0.11 (0.15) &    0.06 (0.10) &    0.10 (0.13) &    0.06 (0.10) &    0.12 (0.15) &    0.08 (0.13) &    0.11 (0.13) &    0.08 (0.12) &    0.12 (0.16) \\
    \bottomrule
    \end{tabular}
    }
    \caption{CIFAR-10 --- Performance results when images are randomly perturbed during training using additive random noise sampled from the cube $[-a, a]^n$ with $a = 1$. The abbreviation `Adv.' should be read as `Adversarial'. The quantities computed are defined in Section~\ref{sec:experimentalSetup}. Accuracy and susceptibility are reported as percentages. The norms of the adversarial attacks are reported in the form `mean (standard deviation)', calculated by averaging over all of the correctly classified and adversarially susceptible images in each of the training and test sets.}
    \label{tbl:randomTraining:linf:1.0}
\end{table*}

\begin{table*}
    \centering
    \scalebox{0.675}{
    \begin{tabular}{rr|ccccccccc}
    \toprule
    {} &    {} &         0 vs 1 &         1 vs 2 &         2 vs 3 &         3 vs 4 &         4 vs 5 &         5 vs 6 &         6 vs 7 &         7 vs 8 &         8 vs 9 \\
    \midrule
    Accuracy                   &  train &          86.82 &          91.61 &          84.19 &          85.79 &          84.17 &          90.45 &          94.29 &          93.40 &          80.22 \\
                               &   test &          84.55 &          89.85 &          78.55 &          80.25 &          80.80 &          88.10 &          91.35 &          91.20 &          77.25 \\
    Adv. susceptibility        &  train &          72.14 &          98.50 &          94.35 &          98.18 &          92.23 &          97.63 &          98.26 &          98.73 &          68.44 \\
                               &   test &          71.08 &          98.16 &          93.19 &          98.32 &          92.51 &          97.73 &          98.14 &          98.85 &          68.41 \\
    Adv. attack $\ell^1$ norm     &  train &  43.25 (58.85) &  25.08 (38.16) &  34.84 (47.12) &  25.38 (40.59) &  48.40 (54.21) &  25.85 (39.18) &  38.47 (45.03) &  29.10 (42.32) &  53.87 (66.98) \\
                               &   test &  42.76 (58.44) &  24.30 (37.83) &  34.82 (48.19) &  25.47 (40.99) &  46.92 (53.64) &  27.88 (42.30) &  39.74 (46.60) &  28.83 (42.06) &  54.82 (67.20) \\
    Adv. attack $\ell^2$ norm     &  train &    1.18 (1.61) &    0.72 (1.09) &    0.95 (1.28) &    0.67 (1.06) &    1.36 (1.52) &    0.75 (1.12) &    1.10 (1.29) &    0.78 (1.13) &    1.41 (1.75) \\
                               &   test &    1.18 (1.61) &    0.69 (1.07) &    0.95 (1.31) &    0.67 (1.07) &    1.32 (1.50) &    0.81 (1.22) &    1.13 (1.33) &    0.77 (1.13) &    1.43 (1.76) \\
    Adv. attack $\ell^\infty$ norm &  train &    0.11 (0.15) &    0.07 (0.11) &    0.09 (0.12) &    0.06 (0.09) &    0.13 (0.16) &    0.07 (0.11) &    0.10 (0.13) &    0.07 (0.11) &    0.13 (0.16) \\
                               &   test &    0.11 (0.16) &    0.07 (0.11) &    0.09 (0.13) &    0.06 (0.09) &    0.13 (0.15) &    0.08 (0.12) &    0.11 (0.13) &    0.07 (0.11) &    0.13 (0.16) \\
    \bottomrule
    \end{tabular}
    }
    \caption{CIFAR-10 --- Performance results when images are randomly perturbed during training using additive random noise sampled from the ball $\mathbb{B}^n_b$ with $b = 32$. The abbreviation `Adv.' should be read as `Adversarial'. The quantities computed are defined in Section~\ref{sec:experimentalSetup}. Accuracy and susceptibility are reported as percentages. The norms of the adversarial attacks are reported in the form `mean (standard deviation)', calculated by averaging over all of the correctly classified and adversarially susceptible images in each of the training and test sets.}
    \label{tbl:randomTraining:l2:32}
\end{table*}

\subsubsection{Experimental results on the Fashion MNIST dataset}\label{sec:experimentResults:fashionmnist}

\begin{table}
    \centering{
    \scalebox{0.675}{
    \begin{tabular}{cc}
        Index & Name
        \\
        \midrule
        0 & T-shirt/top
        \\
        1 & Trouser
        \\
        2 & Pullover
        \\
        3 & Dress
        \\
        4 & Coat
        \\
        5 & Sandal
        \\
        6 & Shirt
        \\
        7 & Sneaker
        \\
        8 & Bag
        \\
        9 & Ankle boot
        \\
    \end{tabular}
    }
    \caption{Fashion MNIST --- Class names associated with each class index.}
    \label{tbl:fashionmnist:classNames}
    }
\end{table}

The Fashion MNIST dataset consists of $28\times 28$ pixel grayscale image (which we converted to RGB by simply duplicating the channels), separated into 10 classes.
The English names for these classes are given in Table~\ref{tbl:fashionmnist:classNames}
The network structure used in this case is similar to that described in Table~\ref{tbl:architecture}, but with the layers Conv-5 and Conv-6 removed.
The same training and evaluation procedures outlined in Section~\ref{sec:experimentalSetup} were applied, and the results are given below.
For brevity, we only present the results on the problems of the form `class $i$-vs-class $i+1$'.

Table~\ref{tbl:fmnist:sus} shows the accuracy and susceptibility to adversarial and random perturbations of the network trained on each binary classification problem. The random perturbations are categorised by their size in the Euclidean norm in the form of $\delta$, since they are uniformly sampled at random from the ball with radius $\delta\epsilon$, where $\epsilon$ denotes the Euclidean norm of the smallest adversarial perturbation identified on each image, while Table~\ref{tbl:fmnist:norms} shows the norms of the adversarial attack constructed with the smallest Euclidean norm on each image.

\begin{figure}
    \begin{subfigure}{\linewidth}
        \centering
        \includegraphics[width=\linewidth]{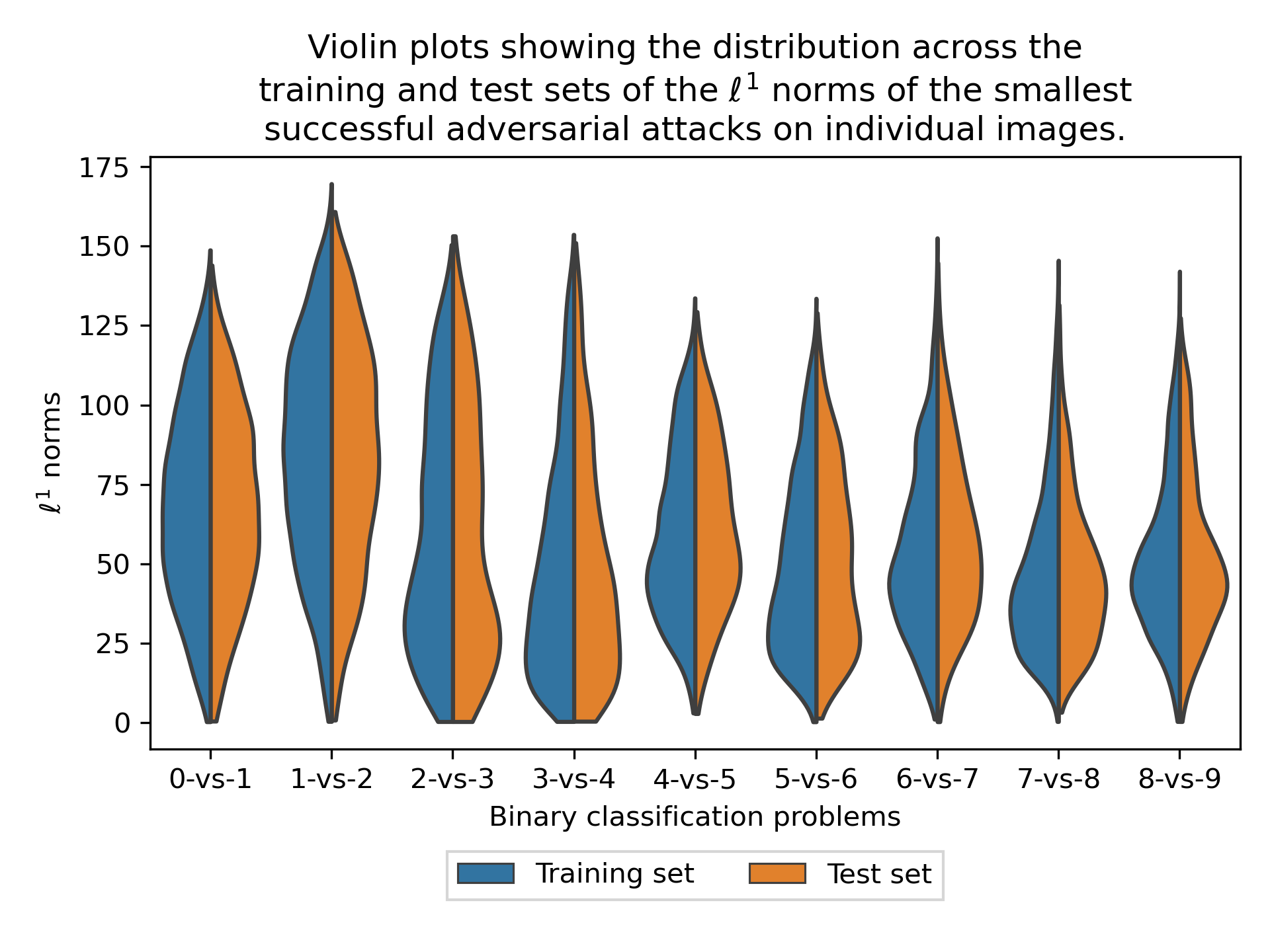}
        \caption{$\ell^1$ norms}
    \end{subfigure}

    \begin{subfigure}{\linewidth}
        \centering
        \includegraphics[width=\linewidth]{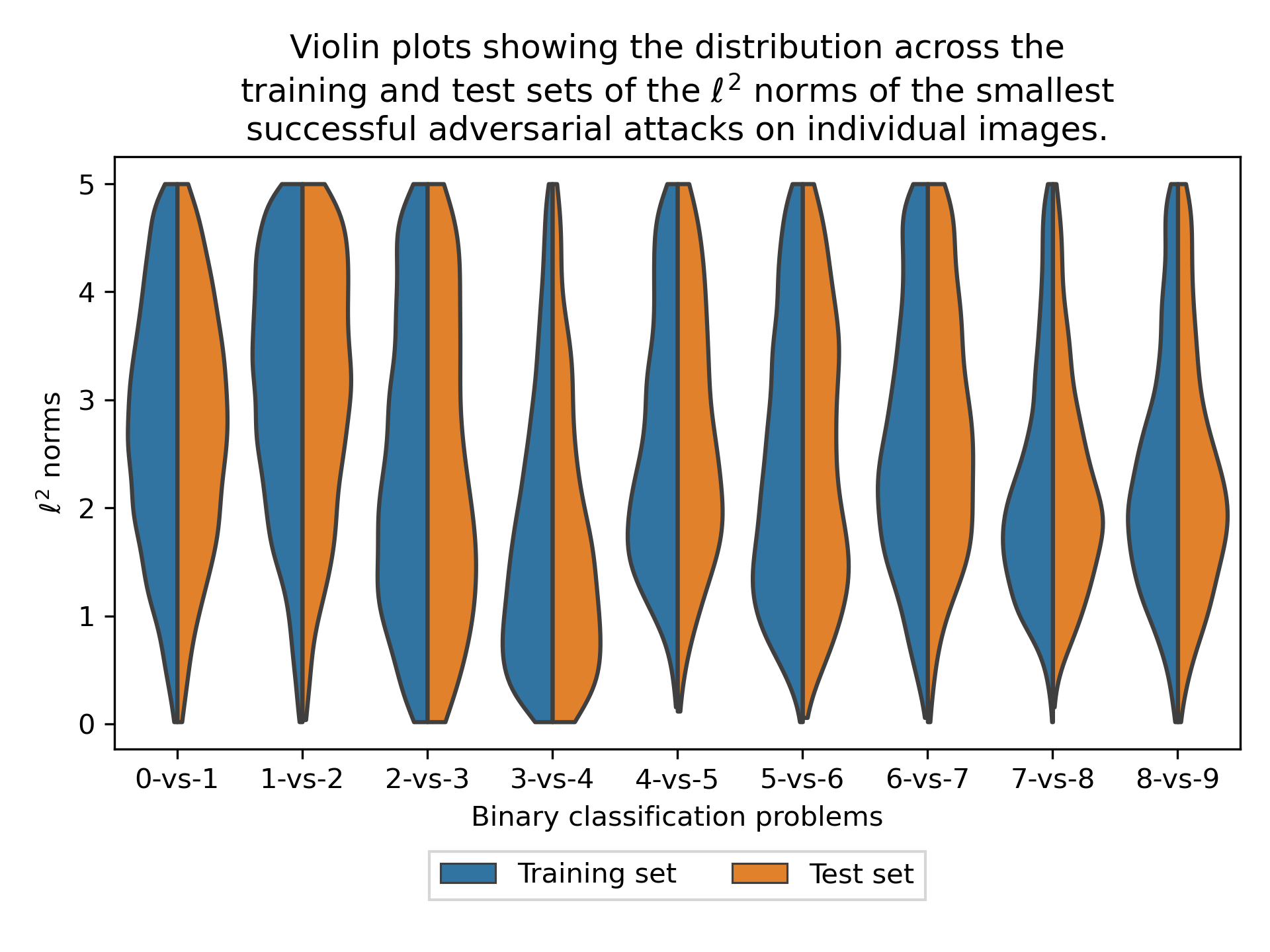}
        \caption{$\ell^2$ norms}
    \end{subfigure}
    
    \begin{subfigure}{\linewidth}
        \centering
        \includegraphics[width=\linewidth]{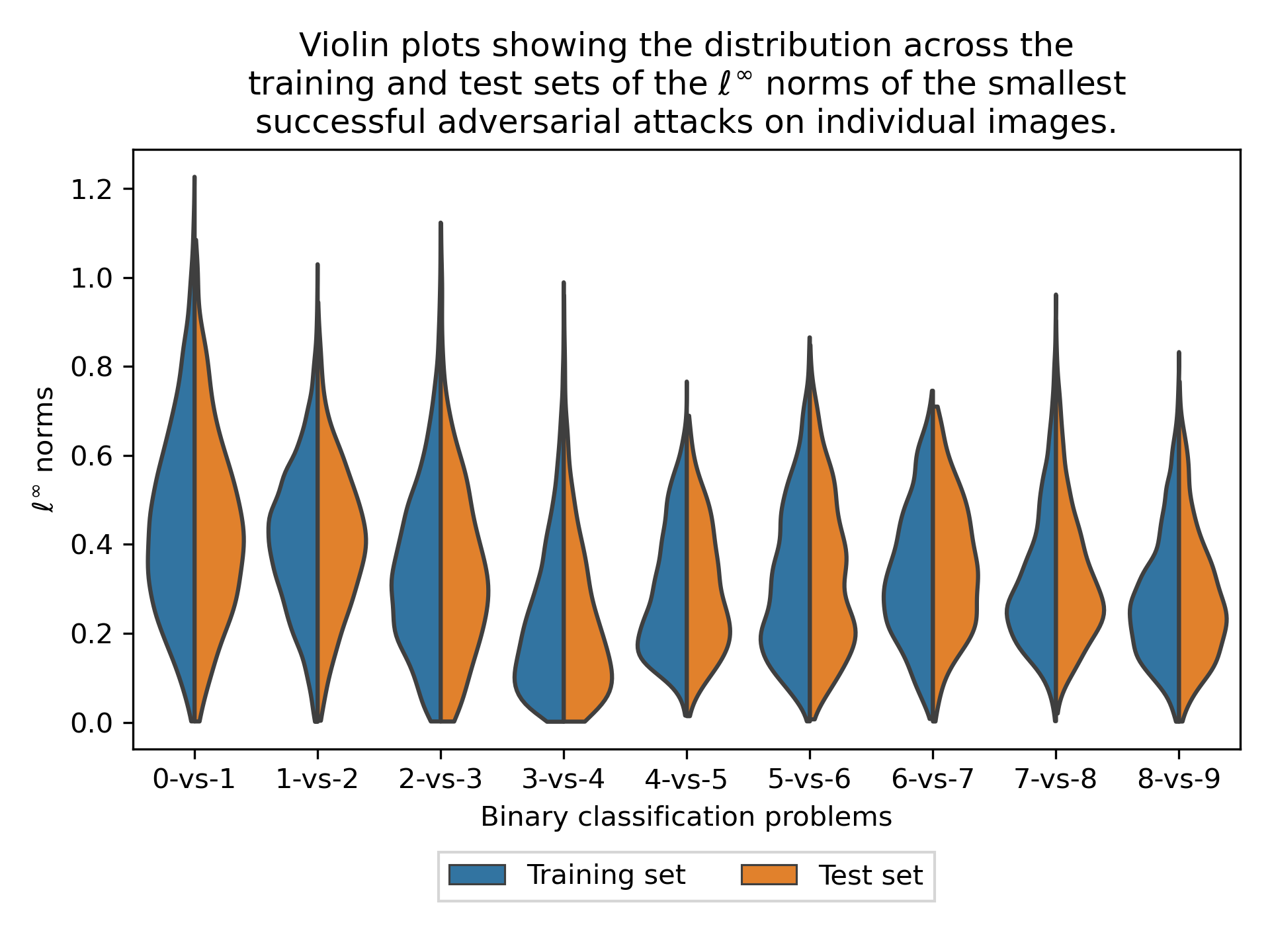}
        \caption{$\ell^\infty$ norms}
    \end{subfigure}
    \caption{Fashion MNIST --- Distribution of norms of smallest successful adversarial attacks on each image.}
    \label{fig:fashionmnist:adversarialViolins}
\end{figure}

\begin{figure}
    \begin{subfigure}{\linewidth}
        \centering
        \includegraphics[width=\linewidth]{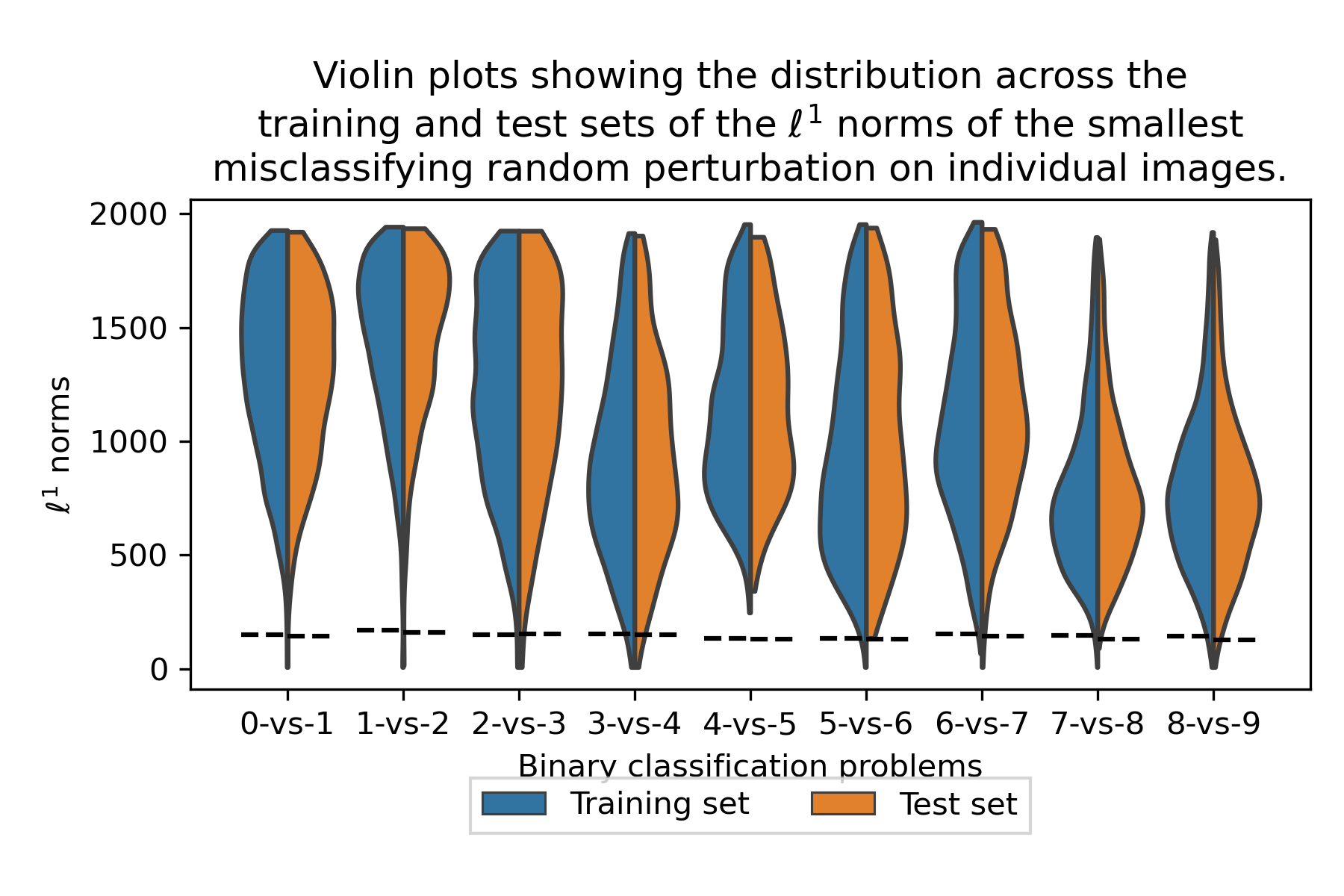}
        \caption{$\ell^1$ norms}
    \end{subfigure}

    \begin{subfigure}{\linewidth}
        \centering
        \includegraphics[width=\linewidth]{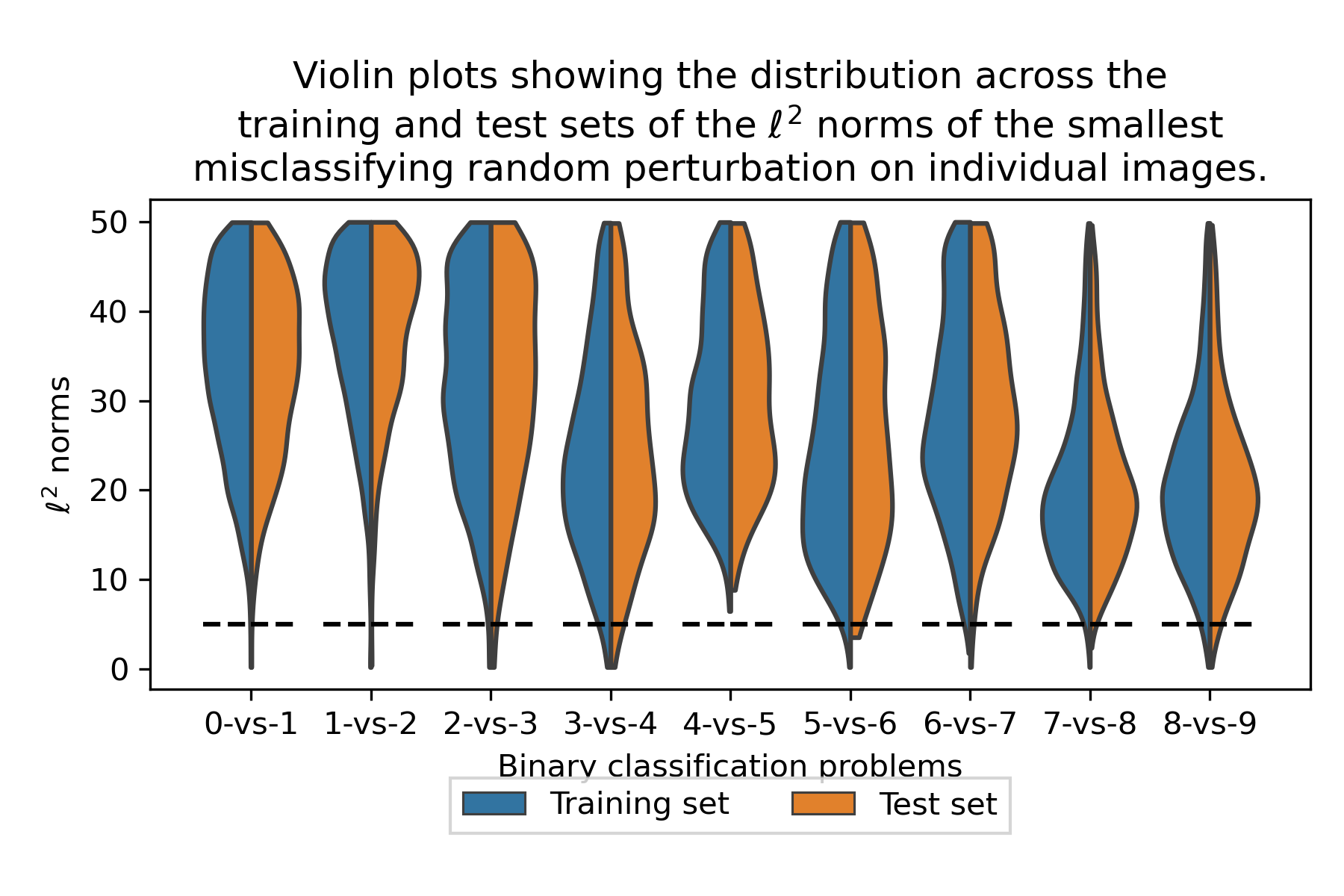}
        \caption{$\ell^2$ norms}
    \end{subfigure}
    
    \begin{subfigure}{\linewidth}
        \centering
        \includegraphics[width=\linewidth]{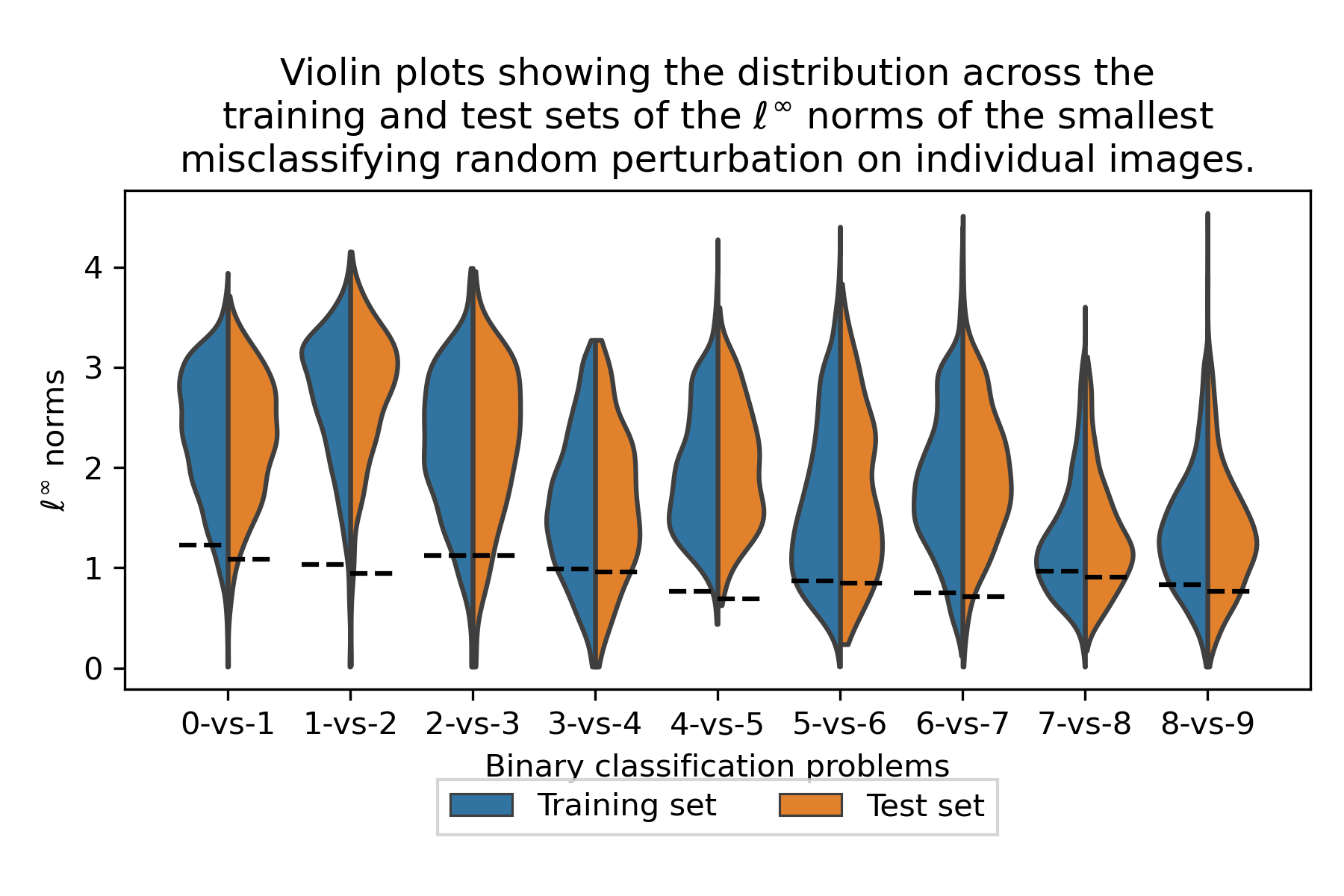}
        \caption{$\ell^\infty$ norms}
    \end{subfigure}
    \caption{Fashion MNIST --- Distribution of norms of the smallest misclassifying random perturbations found on each image. Black dashed lines indicate the size of the largest adversarial attack required on each data set.}
    \label{fig:fashionmnist:randomViolins}
\end{figure}

Violin plots for the distributions of the $\ell^1$, Euclidean and $\ell^{\infty}$ norms of the successful adversarial and random perturbations are given in Figures~\ref{fig:fashionmnist:adversarialViolins} and \ref{fig:fashionmnist:randomViolins}.

\begin{table*}
\centering
\centering
\scalebox{0.675}{
\begin{tabular}{r|ccccccccc}
\toprule
{} &        0 vs 1 &        1 vs 2 &        2 vs 3 &        3 vs 4 &        4 vs 5 &        5 vs 6 &        6 vs 7 &        7 vs 8 &        8 vs 9
\\
\midrule
Accuracy &  99.44, 99.35 &  99.51, 99.40 &  97.73, 96.85 &  97.77, 96.45 &  99.95, 99.95 &  99.96, 99.80 &  99.98, 99.95 &  99.84, 99.70 &  99.41, 99.35
\\
Adversarial susceptibility    &  56.42, 56.87 &  61.81, 62.73 &  26.29, 27.36 &  62.10, 61.43 &  29.51, 29.96 &  38.73, 38.28 &  26.95, 28.21 &  53.58, 53.01 &  55.36, 55.81
\\
Random susceptibility ($\delta = 1$)  &          0, 0 &          0, 0 &       0.03, 0 &          0, 0 &          0, 0 &          0, 0 &       0.06, 0 &       0.02, 0 &          0, 0
\\
Random susceptibility ($\delta = 2$)  &          0, 0 &          0, 0 &    0.26, 0.38 &    0.07, 0.08 &          0, 0 &    0.58, 0.39 &    1.30, 1.42 &    0.87, 1.04 &    0.05, 0.09
\\
Random susceptibility ($\delta = 5$)  &    2.48, 3.27 &    0.45, 0.32 &    4.25, 5.47 &    7.56, 7.68 &  10.71, 13.52 &  22.06, 23.69 &  16.73, 19.15 &  30.39, 31.98 &  13.72, 13.35
\\
Random susceptibility ($\delta = 10$)  &  52.12, 52.92 &  50.39, 49.32 &  54.30, 55.47 &  56.84, 55.44 &  56.63, 57.43 &  61.88, 59.55 &  67.92, 71.45 &  84.33, 84.58 &  84.77, 83.95
\\
\bottomrule
\end{tabular}
}
\caption{Fashion MNIST --- Accuracy and susceptibility of the networks to adversarial and random attacks, reported as percentages in the form `train, test'}
\label{tbl:fmnist:sus}
\end{table*}

\begin{table*}[h]
\centering
\scalebox{0.675}{
\begin{tabular}{rr|ccccccccc}
\toprule
{} & {} &         0 vs 1 &         1 vs 2 &         2 vs 3 &         3 vs 4 &         4 vs 5 &         5 vs 6 &         6 vs 7 &         7 vs 8 &         8 vs 9 
\\
\midrule
Adv. attack $\ell^1$ norm & train &  68.18 (30.79) &  83.46 (34.56) &  59.53 (36.77) &  50.06 (34.27) &  59.45 (26.16) &  49.61 (27.06) &  54.79 (26.78) &  47.29 (25.40) &  52.12 (24.77) 
\\
& test  &  68.12 (30.56) &  83.34 (35.36) &  59.12 (38.72) &  49.78 (35.26) &  61.72 (26.30) &  52.11 (27.52) &  57.76 (26.42) &  47.41 (23.36) &  52.84 (25.64) 
\\
Adv. attack $\ell^2$ norm & train &    2.75 (1.17) &    3.07 (1.16) &    2.39 (1.33) &    1.76 (1.20) &    2.60 (1.15) &    2.40 (1.24) &    2.63 (1.20) &    2.22 (1.08) &    2.32 (1.12) 
\\
& test  &    2.75 (1.15) &    3.06 (1.19) &    2.36 (1.40) &    1.75 (1.24) &    2.70 (1.16) &    2.52 (1.26) &    2.79 (1.17) &    2.24 (1.01) &    2.34 (1.14) 
\\
Adv. attack $\ell^\infty$ norm & train &    0.45 (0.22) &    0.41 (0.17) &    0.34 (0.19) &    0.23 (0.16) &    0.29 (0.14) &    0.32 (0.17) &    0.33 (0.15) &    0.32 (0.15) &    0.28 (0.14) 
\\
& test  &    0.44 (0.21) &    0.41 (0.17) &    0.34 (0.19) &    0.23 (0.17) &    0.30 (0.14) &    0.33 (0.17) &    0.35 (0.15) &    0.33 (0.15) &    0.29 (0.15) 
\\
\bottomrule
\end{tabular}
}
\caption{Fashion MNIST --- Means and standard deviations of the norms of the smallest successful adversarial attack on each image in the training and test set, reported in the form `mean (standard deviation)'.}\label{tbl:fmnist:norms}
\end{table*}

\subsubsection{Experimental results on the German Traffic Sign Recognition Benchmark dataset (GTSRB)}\label{sec:experimentResults:gtsrb}

The German Traffic Sign Recognition Benchmark (GTSRB) dataset consists of RGB colour images with size $30 \times 30 \times 3$, divided into more than 40 classes.
Here, we have demonstrated our results using six of these 40 classes, selected to be relatively visually distinct from each other, and therefore to produce binary classification problems which may be expected to be more robust to adversarial attacks.
The names of these data classes is given in Table~\ref{tbl:gtsrb:classnames}.
The network structure used in this case is the same as the Fashion MNIST dataset in Section~\ref{sec:experimentResults:fashionmnist}, which is as described in Table~\ref{tbl:architecture} but without Conv-5 or Conv-6.
The experimental setup was otherwise as described in Section~\ref{sec:experimentalSetup}.
For brevity, we report the results on the problems of the form `class $i$-vs-class $i+1$'.

Table~\ref{tbl:gtsrb:sus} shows the accuracy and susceptibility to adversarial and random perturbations of the network trained on each binary classification problem. The random perturbations are categorised by their size in the Euclidean norm in the form of $\delta$, since they are uniformly sampled at random from the ball with radius $\delta\epsilon$, where $\epsilon$ denotes the Euclidean norm of the smallest adversarial perturbation identified on each image, while Table~\ref{tbl:gtsrb:norms} shows the norms of the adversarial attack constructed with the smallest Euclidean norm on each image.

\begin{figure}
    \begin{subfigure}{\linewidth}
        \centering
        \includegraphics[width=\linewidth]{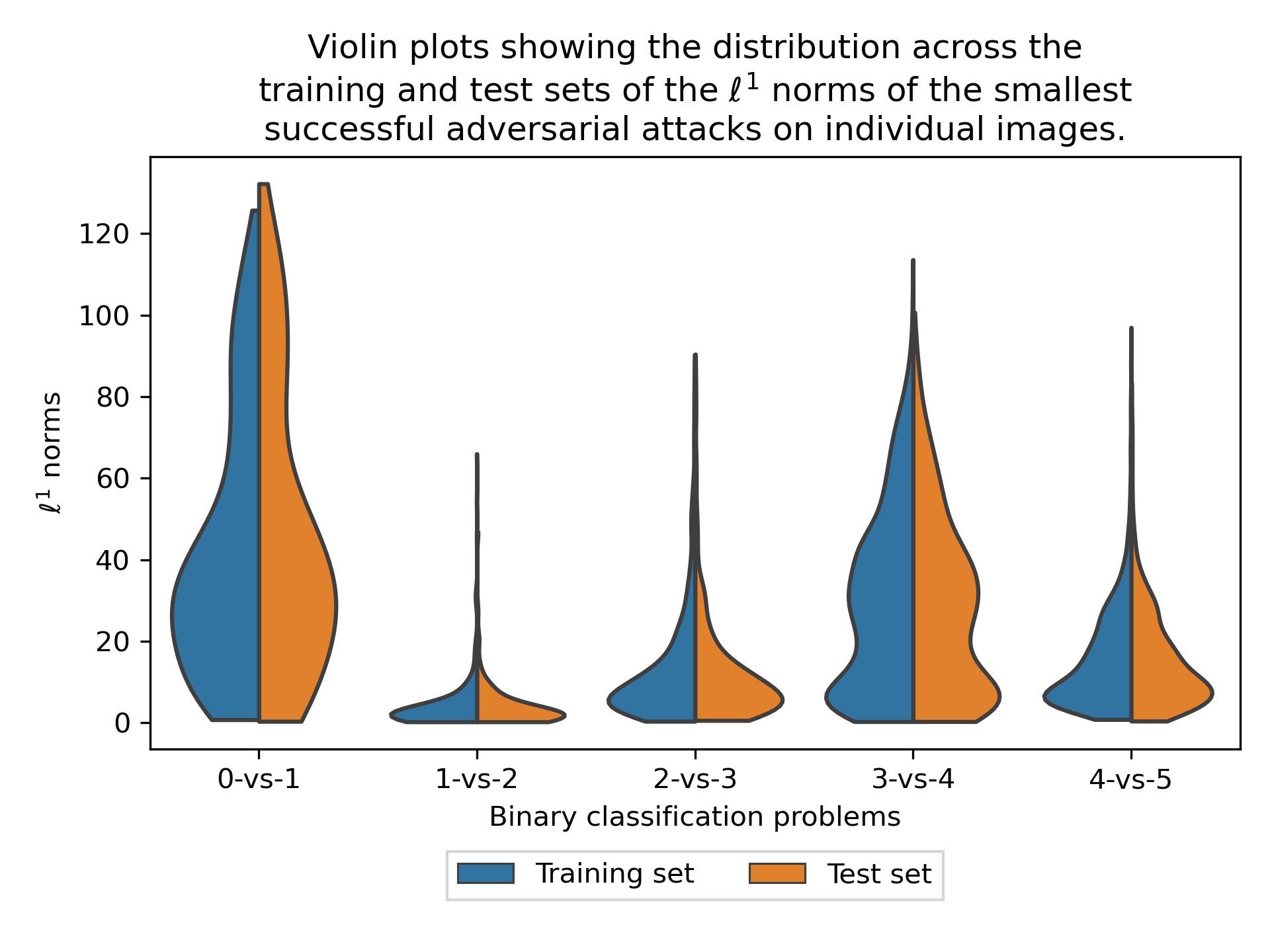}
        \caption{$\ell^1$ norms}
    \end{subfigure}

    \begin{subfigure}{\linewidth}
        \centering
        \includegraphics[width=\linewidth]{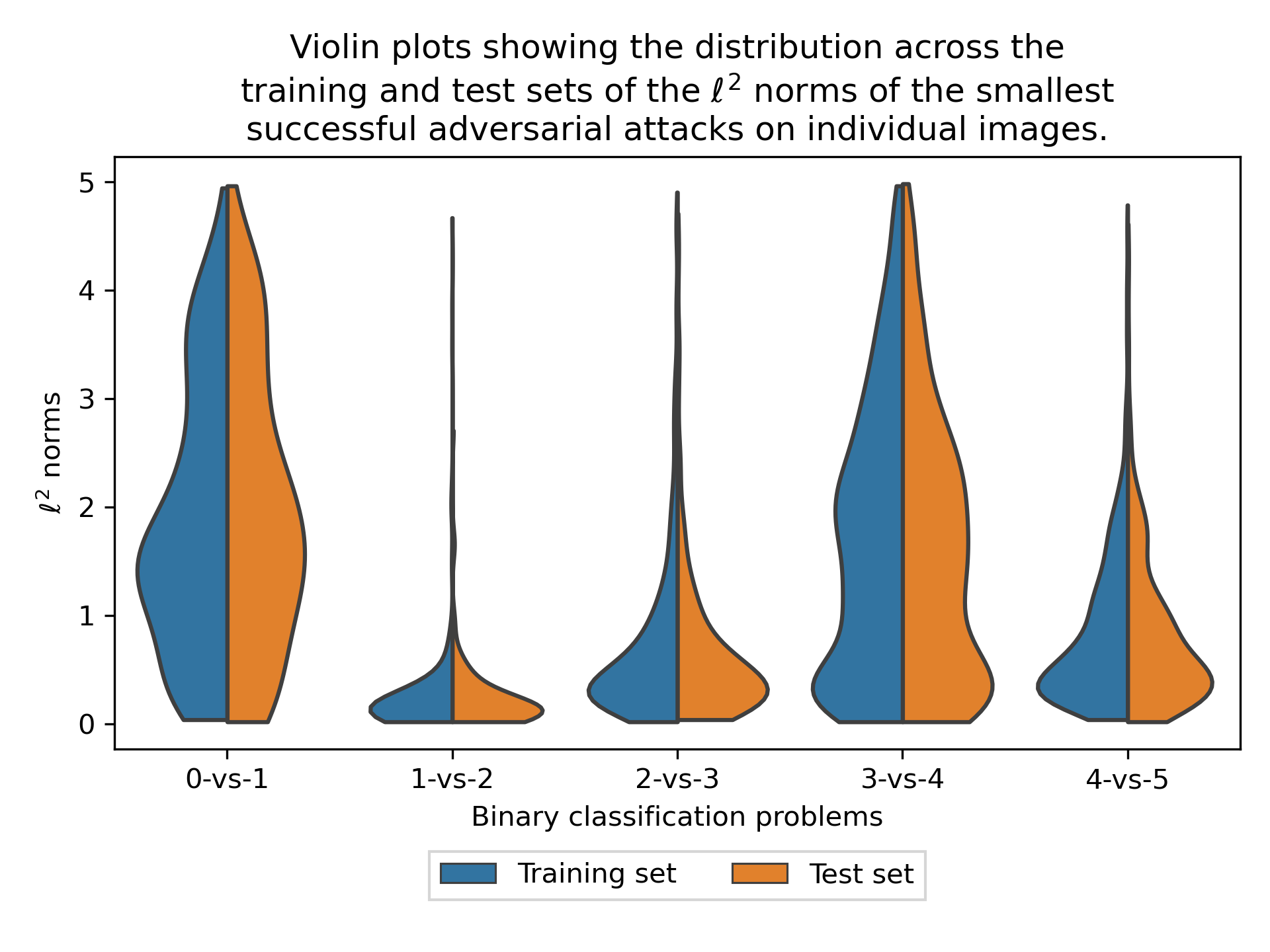}
        \caption{$\ell^2$ norms}
    \end{subfigure}
    
    \begin{subfigure}{\linewidth}
        \centering
        \includegraphics[width=\linewidth]{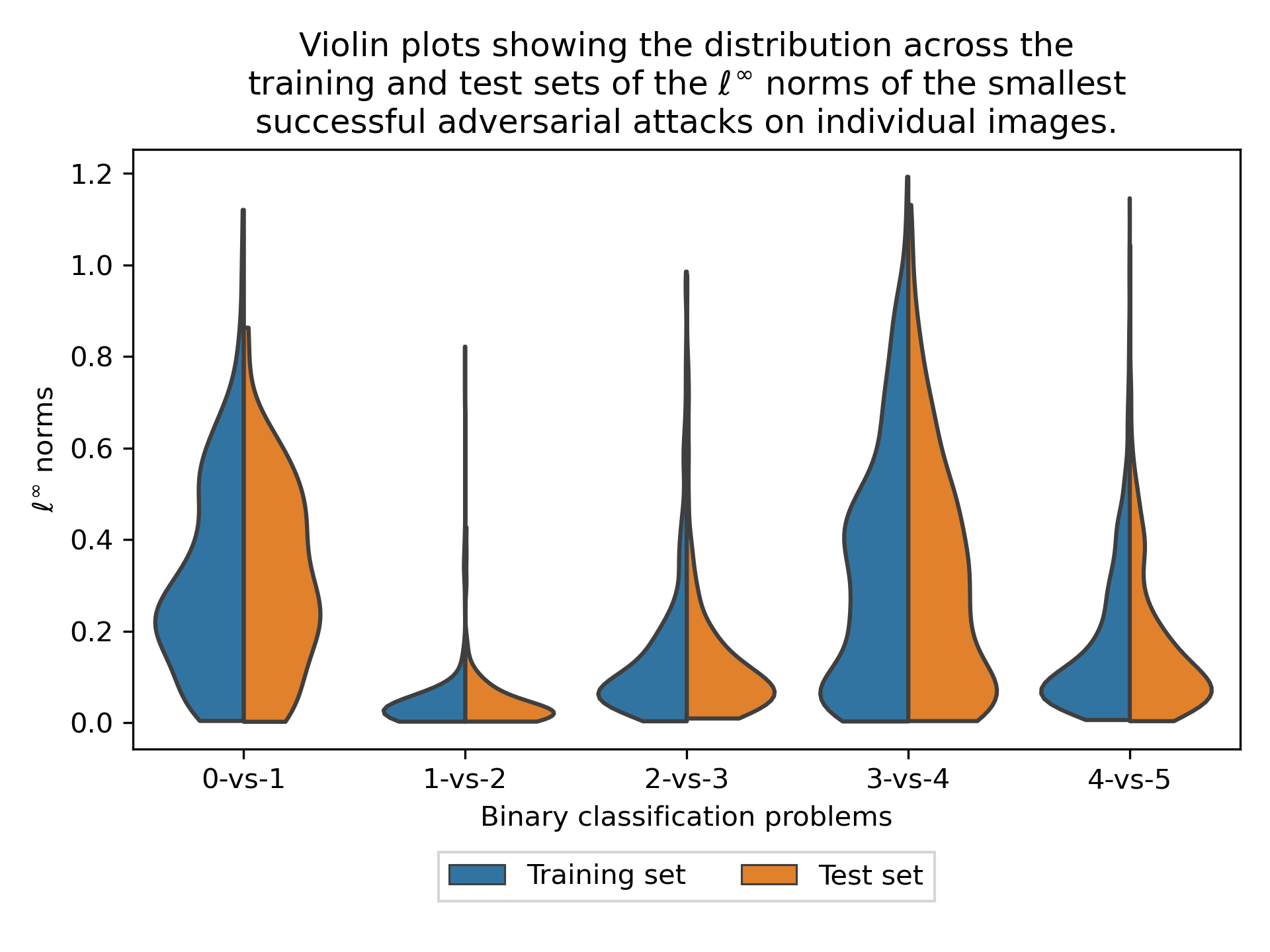}
        \caption{$\ell^\infty$ norms}
    \end{subfigure}
    \caption{GTSRB --- Distribution of norms of smallest successful adversarial attacks on each image.}
    \label{fig:gtsrb:adversarialViolin}
\end{figure}

\begin{figure}
    \begin{subfigure}{\linewidth}
        \centering
        \includegraphics[width=\linewidth]{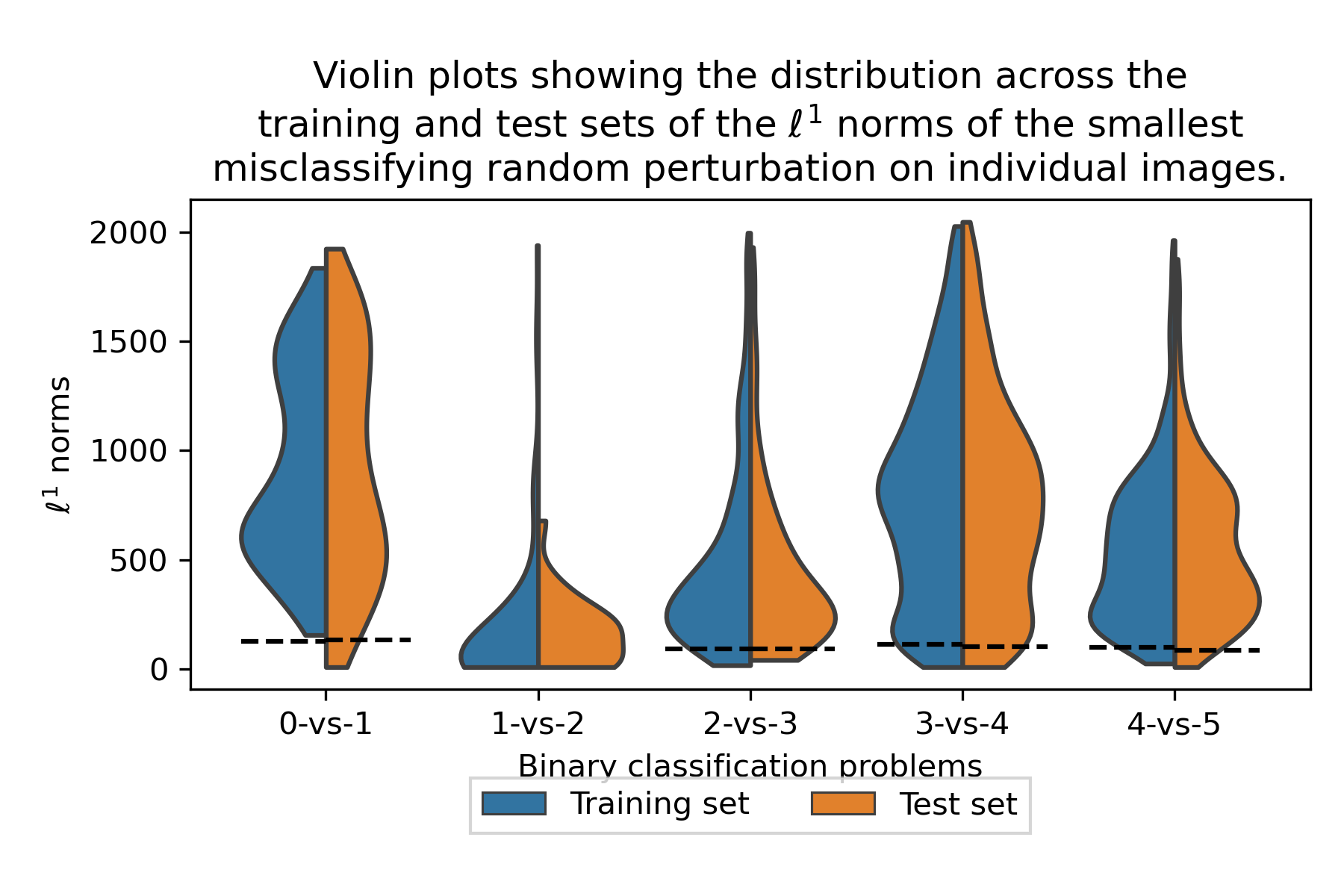}
        \caption{$\ell^1$ norms}
    \end{subfigure}

    \begin{subfigure}{\linewidth}
        \centering
        \includegraphics[width=\linewidth]{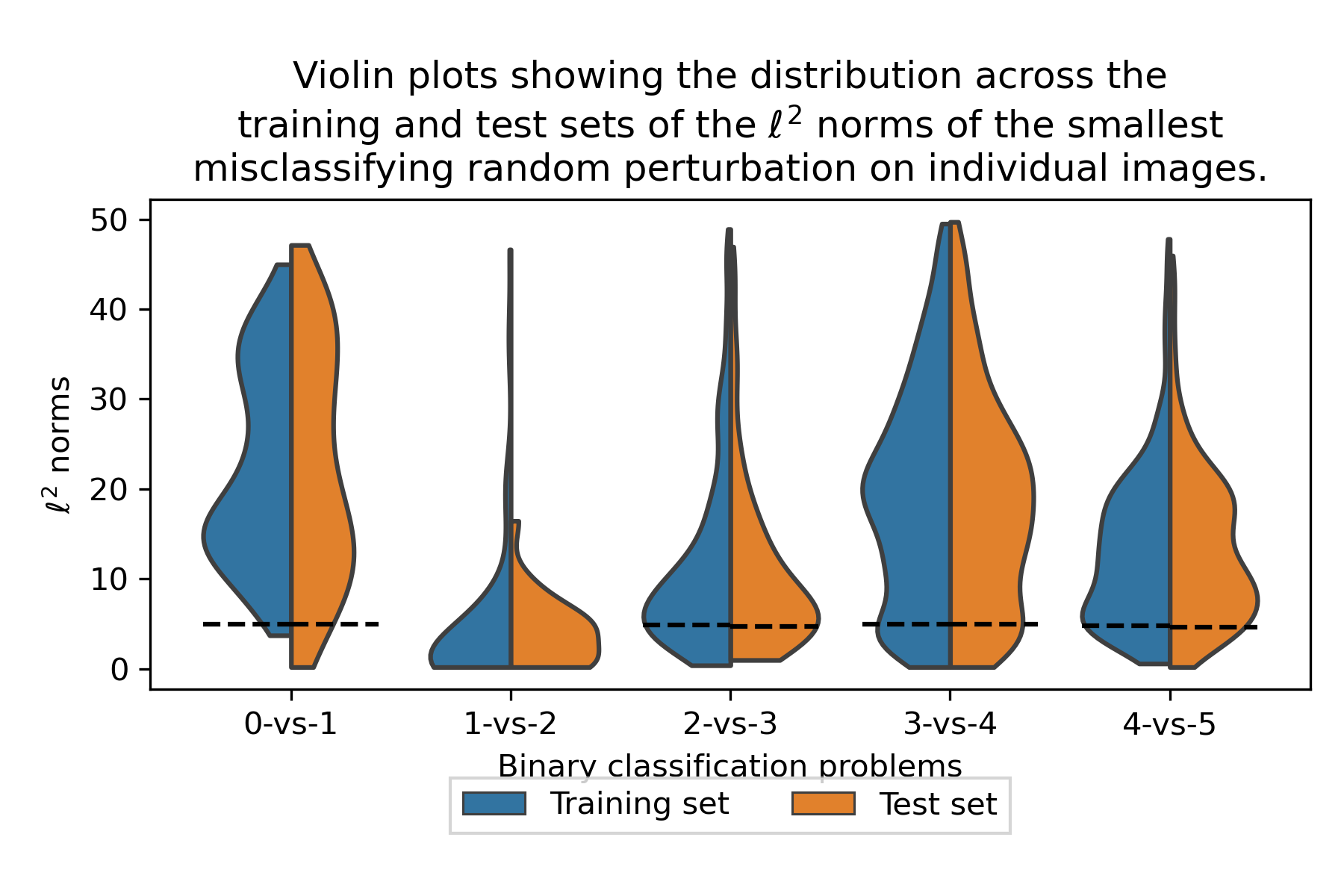}
        \caption{$\ell^2$ norms}
    \end{subfigure}
    
    \begin{subfigure}{\linewidth}
        \centering
        \includegraphics[width=\linewidth]{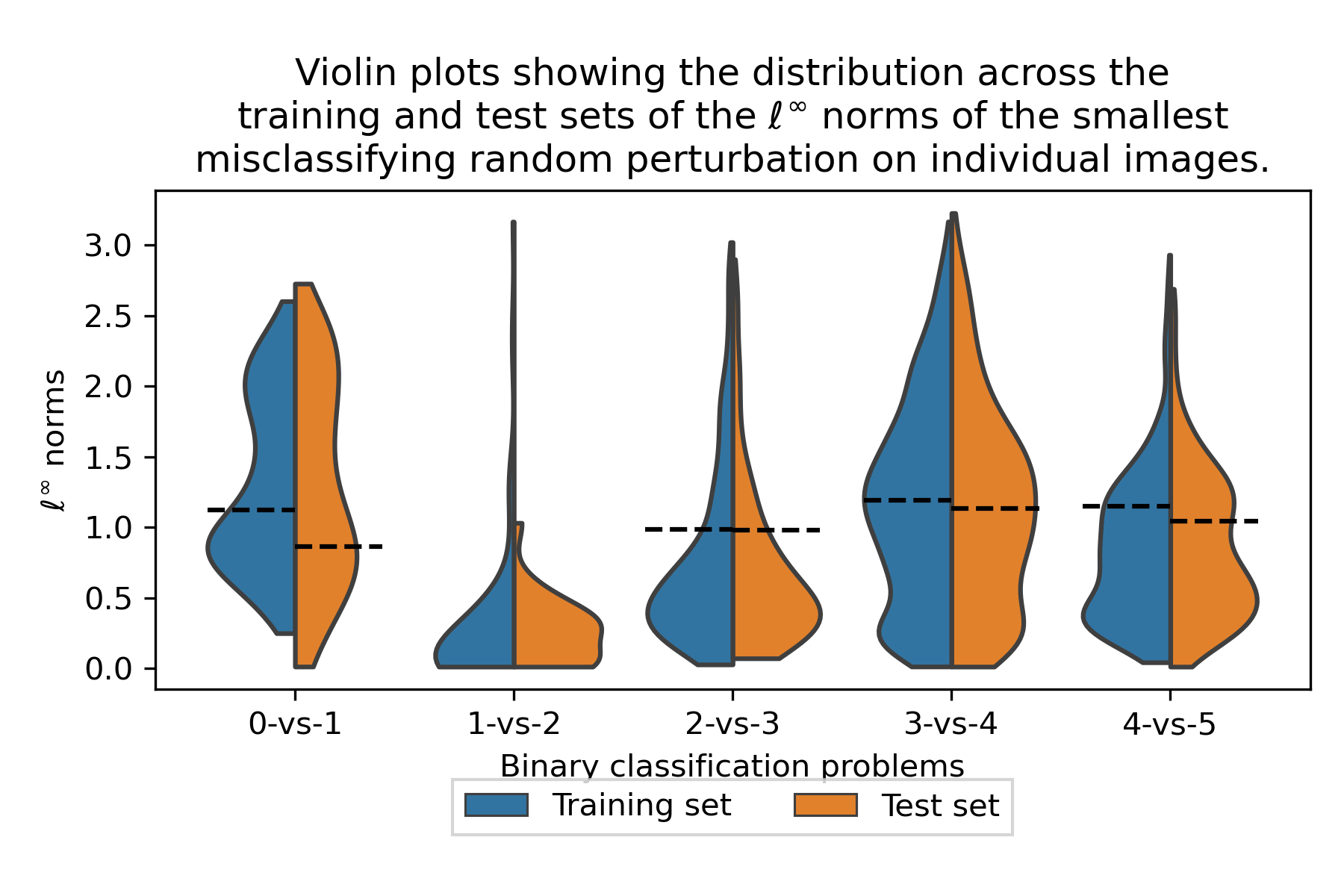}
        \caption{$\ell^\infty$ norms}
    \end{subfigure}
    \caption{GTSRB --- Distribution of norms of the smallest misclassifying random perturbations found on each image. Black dashed lines indicate the size of the largest adversarial attack required on each data set.}
    \label{fig:gtsrb:randomViolin}
\end{figure}

Violin plots for the distributions of the $\ell^1$, Euclidean and $\ell^{\infty}$ norms of the successful adversarial and random perturbations are shown in Figures~\ref{fig:gtsrb:adversarialViolin} and \ref{fig:gtsrb:randomViolin}.

\begin{table}
    \centering
    \centering
    \scalebox{0.675}{
    \begin{tabular}{cc}
        Index & Class name
        \\
        \midrule
        0 & Speed limit (20km/h)
        \\
        1 & End of no passing for vehicles $>$ 3.5 tons
        \\
        2 & Keep right
        \\
        3 & Turn right ahead
        \\
        4 & Road work
        \\
        5 & General caution
        \\
        6 & End of speed limit (80km/h)
    \end{tabular}
    }
    \caption{GTSRB --- Class names associated with each class index.}
    \label{tbl:gtsrb:classnames}
\end{table}

\begin{table*}
\centering
\centering
\scalebox{0.675}{
\begin{tabular}{r|ccccc}
\toprule
{} &        0 vs 1 &       1 vs 2 &        2 vs 3 &        3 vs 4 &        4 vs 5 \\
\midrule
Accuracy &  96.84, 98.51 &  98.04, 97.63 &  99.95, 99.28 &  98.32, 98.14 &  100.00, 100.00 \\
Adversarial susceptibility    &  94.44, 94.70 &  34.66, 32.98 &  62.96, 66.14 &  77.53, 77.00 &  88.95, 89.35 \\
Random susceptibility ($\delta = 1$)  &          0, 0 &    2.52, 1.84 &          0, 0 &          0, 0 &       0.12, 0 \\
Random susceptibility ($\delta = 2$)  &          0, 0 &    3.06, 2.76 &    0.17, 0.36 &    0.44, 0.19 &    0.36, 0.41 \\
Random susceptibility ($\delta = 5$)  &  32.18, 32.80 &    5.76, 3.23 &    4.96, 5.10 &  22.11, 23.63 &    2.63, 3.01 \\
Random susceptibility ($\delta = 10$) &  61.25, 62.40 &  15.65, 10.60 &  39.26, 36.07 &  83.26, 81.85 &  32.18, 31.64 \\
\bottomrule
\end{tabular}
}
\caption{GTSRB --- Accuracy and susceptibility of the networks to adversarial and random attacks, reported in the form `train, test'}
\label{tbl:gtsrb:sus}
\end{table*}

\begin{table*}
\centering
\centering
\scalebox{0.675}{
\begin{tabular}{rr|ccccc}
\toprule
{} & {} &          0 vs 1 &           1 vs 2 &           2 vs 3 &          3 vs 4 &           4 vs 5 
\\
\midrule
Adv. attack $\ell^1$ norm & train &  42.38 (31.92) &  4.20 (7.13) &  13.42 (14.52) &  29.80 (22.24) &  15.55 (12.56) 
\\
& test  &  46.09 (35.10) &  4.00 (5.91) &  12.22 (13.13) &  28.61 (22.18) &  15.36 (12.53) 
\\
Adv. attack $\ell^2$ norm & train &    1.84 (1.22) &  0.25 (0.48) &    0.74 (0.80) &    1.70 (1.25) &    0.85 (0.71) 
\\
& test  &    1.95 (1.29) &  0.24 (0.35) &    0.67 (0.72) &    1.61 (1.24) &    0.83 (0.71) 
\\
Adv. attack $\ell^\infty$ norm & train &    0.31 (0.21) &  0.05 (0.09) &    0.15 (0.16) &    0.33 (0.26) &    0.17 (0.16) 
\\
& test  &    0.31 (0.20) &  0.04 (0.06) &    0.14 (0.14) &    0.31 (0.25) &    0.17 (0.16) 
\\
\bottomrule
\end{tabular}
}
\caption{GTSRB --- Means and standard deviations of the norms of the smallest successful adversarial attack on each image in the training and test set, reported in the form `mean (standard deviation)'.}
\label{tbl:gtsrb:norms}
\end{table*}

\subsubsection{Experimental results on ImageNet}\label{sec:experimentResults:imagenet}
The ImageNet dataset~\cite{imagenet} consists of RGB images of various sizes from 1,000 classes.
We experimented using a pre-trained VGG19~\cite{simonyan2015deep} and ResNet50~\cite{he2016deep} network as described in Section~\ref{sec:experimentSetup:imagenet}.

Table~\ref{tbl:imagenet:sus} summarises the results of these experiments.
In contrast to previous experiments, we only report random susceptibility for $\delta = 10$.
Experiments with random perturbations for $\delta \in \{1, 2, 5\}$ produced virtually zero misclassifications, despite the apparently high adversarial susceptibility of the networks, and for brevity the detailed results are not reported here.

\begin{figure*}
    \centering
    \includegraphics[width=0.75\linewidth]{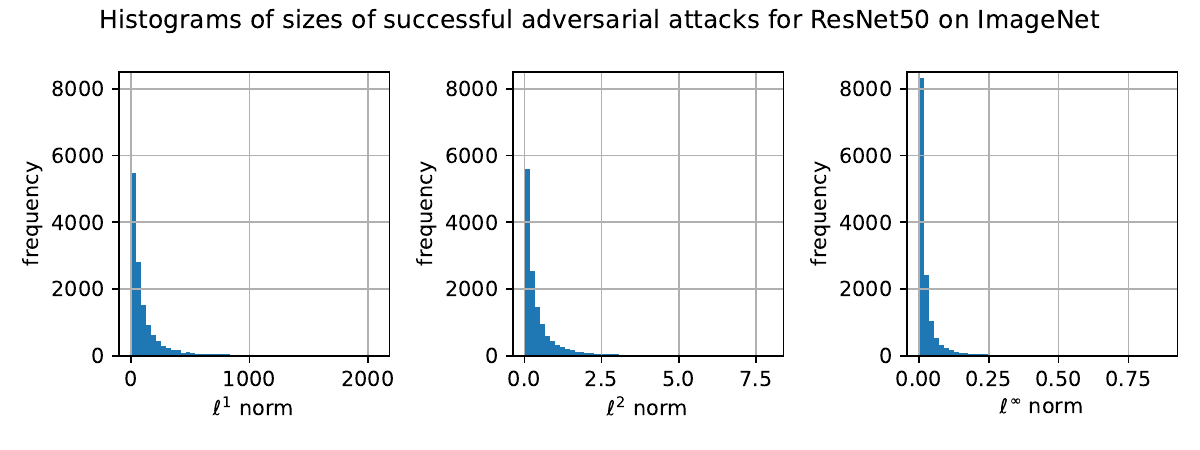}
    \caption{ImageNet --- Histograms showing the distribution of sizes of successful adversarial attacks on images from the ImageNet validation set for a pre-trained ResNet50 model.}
    \label{fig:imagenet:resnet50-adversarial-sizes}
\end{figure*}

\begin{figure*}
    \centering
    \includegraphics[width=0.75\linewidth]{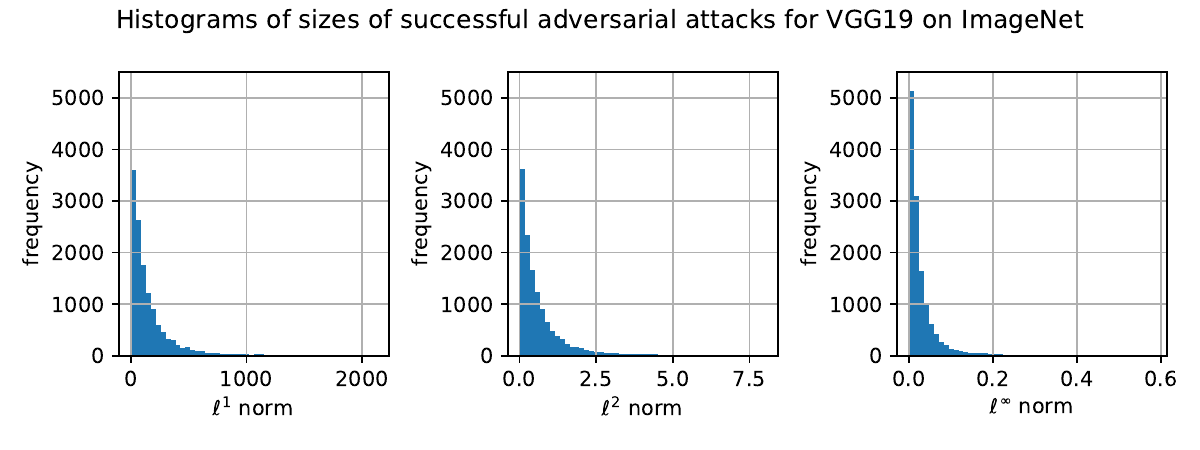}
    \caption{ImageNet --- Histograms showing the distribution of sizes of successful adversarial attacks on images from the ImageNet validation set for a pre-trained VGG19 model.}
    \label{fig:imagenet:vgg19-adversarial-sizes}
\end{figure*}

Figures~\ref{fig:imagenet:resnet50-adversarial-sizes} and \ref{fig:imagenet:vgg19-adversarial-sizes} show distributions of the sizes of successful adversarial attacks measured in different norms for each model.
It is clear from these plots that in both cases the majority of images are susceptible to small adversarial perturbations.
For ResNet50, 93.3\% of images were susceptible to an adversarial attack with $\ell^{\infty}$ norm (measuring the absolute value of the largest change to any individual pixel) less than 0.1, while this was 94.3\% for VGG19.
Despite this, the results in Table~\ref{tbl:imagenet:sus} show that $\leq 2.5\%$ of images were misclassified by any of the 2,000 random attacks sampled from the ball with radius 10 times larger than the Euclidean norm of the adversarial attack.
We observe that the rate of misclassification after these large random perturbations is significantly smaller than for the other datasets.
The theoretical results presented above suggest that this may be due to the much higher dimensionality of the images.
Both of the pretrained networks we use accept inputs of size $224\times224\times3$, meaning that they have 150,528 individual attributes, in contrast with 3,072 attributes for CIFAR-10 images.

\begin{table}
    \centering
    \scalebox{0.675}{
    \begin{tabular}{c|cc}
        \midrule
         & ResNet50 & VGG19 \\
        \midrule
        Accuracy  & 70.8\% & 66.52\%  \\
        Adversarial attack susceptibility & 94.2\% & 97.07\% \\
        Random attack susceptibility ($\delta = 10$) & 2.5\% & 1.4\% \\
                \midrule
    \end{tabular}
    }
    \caption{ImageNet --- Accuracy and susceptibility of the networks to adversarial and random attacks, reported for 20,480 images from the validation set.}
    \label{tbl:imagenet:sus}
\end{table}

\section{Conclusion}\label{sec:conclusion}

Our new framework for studying the paradox of apparent stability in classification problems allows for rigorous probabilistic bounds that are consistent with empirical observations concerning the simultaneous vulnerability to 
easily constructed worst-case adversarial attacks (Theorems~\ref{thm:susceptibility} and \ref{thm:gradientBasedAttack}) which may universally affect a whole data class (Theorem~\ref{thm:universality}), and robustness against randomly sampled perturbations (Theorem~\ref{thm:undetectability}).
The results are generic in the sense that they deal with 
small perturbations under which any smooth and accurate classifier will behave like the optimal linear classifier~\eqref{eq:classifier}.
As illustrated in Figure~\ref{fig:problemExplained} and Section~\ref{sec:commentsOnGeneralisbility}, the setup can be generalised to cover to broad range of input distributions and classification boundaries and multi-class setups.
In addition to quantifying vulnerabilities, our analysis also raises new issues concerning the most relevant and useful notions of stability in classification.

The overlapping unit ball model that we used, and the two half-ball model in Section~\ref{sec:hemispheres_model}, are closely tied to the use of the Euclidean norm.
We note that there are several applications where spherical input data arises naturally, including remote sensing, climate change modeling, global ionospheric prediction and environmental governance, \cite{Han21}.
It would of course be of interest to establish the extent to which these results can be extended to other choices of norm and input domain. 
We also note that more customised results could be investigated for specific classification tools by exploiting further information, for example, about the architecture, training regime and level of floating point accuracy.

\section*{Acknowledgements}

O.J.S, Q.Z, A.N.G. and I.Y.T were supported in part by the UKRI, EPSRC [UKRI Turing AI Fellowship ARaISE EP/V025295/2 and UKRI Trustworthy Autonomous Systems Node in Verifiability EP/V026801/2].
D.J.H. and A.B. were supported by EPSRC grant EP/V046527/1.

\bibliographystyle{elsarticle-num} 
\bibliography{references}

\appendix

\gdef\thesection{\Alph{section}}
\makeatletter
\renewcommand\@seccntformat[1]{\appendixname\ \csname the#1\endcsname.\hspace{0.5em}}
\makeatother

\section{Proof of Theorem~\ref{thm:fixedDataPoint}}\label{sec:proof:fixedDataPoint}
Expanding the probability as an integral, using the fact that the density of the uniform distribution $\mathcal{U}(B_{\delta}^n(x))$ is just the reciprocal of the volume of a ball with radius $\delta$, we have
\begin{align*}
    &P(z \sim \mathcal{U}(B_{\delta}^n(x)) \suchthat \sign(z \cdot \nu) \neq \sign(x \cdot \nu))
    \\&
    = 
    \frac{1}{V^n \delta^n}
    \int_{B_{\delta}^n(x)} \mathbb{I}_{\{\sign(z \cdot \nu) \neq \sign(x \cdot \nu)\}} dz.
\end{align*}
Since $\nu$ is the fixed normal vector to the plane $\Pi$, the integral here is simply measuring the volume of a spherical cap.
If we assume (without loss of generality) that $x \cdot \nu < 0$, then this cap may be expressed as the set 
\begin{equation*}
    C = \{z \in \mathbb{R}^n \suchthat \|z\| \leq \delta \text{ and } z \cdot \nu \geq 0 \}.
\end{equation*}
Since a spherical cap may be contained within a hemisphere of a different ball, we may prove the following bound:
\begin{lemma}[Spherical cap volume bound]\label{lem:capVolumeBound}
    Let $n$ be a positive integer, and $r \geq h > 0$. Then,
    \begin{align*}
        V^n_{\operatorname{cap}}(r, h) \leq \frac{1}{2} V^n r^n \Big( 1 - \Big(1 - \frac{h}{r} \Big)^2 \Big)^{\frac{n}{2}}.
    \end{align*}
\end{lemma}
By assumption, the height of the cap $C$ is $\delta - \epsilon$, and therefore
\begin{equation*}
    \int_{C} 1 dz \leq \frac{1}{2} V^n \delta^n \Big( 1 - \Big(1 - \frac{\delta - \epsilon}{\delta} \Big)^2 \Big)^{\frac{n}{2}},
\end{equation*}
and the result of the theorem follows.

\section{Proofs of results for the two balls model in Section~\ref{sec:two-balls-model}}

\subsection{Proof of Theorem~\ref{thm:accuracy}}\label{sec:accuracyProof}
Expanding the probability using the definition of the distribution $\mathcal{D}_{\epsilon}$, and the definition of the classifier $f$, we have
\begin{align*}
    &P((x, \ell) \sim \mathcal{D}_{\epsilon} \suchthat f(x) = \ell) 
    \\&
    = 
    \frac{1}{2}
    P(x \sim \mathcal{D}_{0} \suchthat f(x) = 0)
    +
    \frac{1}{2}
    P(x \sim \mathcal{D}_{1} \suchthat f(x) = 1)
\end{align*}
The factor of $\frac{1}{2}$ is due to the fact that samples with either label are sampled with equal probability.
Negating these two probabilities and expressing them as integrals using the densities $p_0$ and $p_1$ associated with $\mathcal{D}_0$ and $\mathcal{D}_1$ respectively (the existence of these densities is a requirement of the SmAC property), we have
\begin{align*}
    &P((x, \ell) \sim \mathcal{D}_{\epsilon} \suchthat f(x) = \ell) 
    \\&
    = 
    1 -
    \frac{1}{2}
    \int_{D_0} \mathbb{I}_{\{x_1 > 0\}} p_0(x) dx
    -
    \frac{1}{2}
    \int_{D_1} \mathbb{I}_{\{x_1 < 0\}} p_1(x) dx.
\end{align*}
The bound on the density $p$ provided by the SmAC property in Definition~\ref{def:smac} (recalling that $r = 1$ for both distributions) therefore implies that
\begin{align*}
    &P((x, \ell) \sim \mathcal{D}_{\epsilon} \suchthat f(x) = \ell) 
    \\&
    \geq
    1 -
    \frac{A}{2 V^n} 
    \Big(
        \int_{D_0} \mathbb{I}_{\{x_1 > 0\}} dx
        +
        \int_{D_1} \mathbb{I}_{\{x_1 < 0\}} dx
    \Big).
\end{align*}
By symmetry, the two integrals have the same value, so we only compute the first.
Since $\epsilon > 0$, this corresponds to the volume of a section of a ball which is smaller than a hemisphere, we may write it as
\begin{align*}
    \int_{D_0} \mathbb{I}_{\{x_1 > 0\}} dx = V^n_{\operatorname{cap}}(1, 1 - \epsilon).
\end{align*}
Lemma~\ref{lem:capVolumeBound} implies that
\begin{align*}
    V^n_{\operatorname{cap}}(1, 1 - \epsilon) \leq \frac{1}{2} V^n (1 - \epsilon^2)^{\frac{n}{2}},
\end{align*}
and the result therefore follows.

\subsection{Proof of Theorem~\ref{thm:susceptibility}}\label{sec:susecptibilityProof}
Using the definition of the classification function $f$ and conditioning on the class label, we may rewrite the probability in question as
\begin{align*}
    &P\big((x, \ell) \sim \mathcal{D}_\epsilon \suchthat \text{there exists } s \in \mathbb{B}^n_{\delta} \text{ such that } f(x + s) \neq \ell \big)
    \\&
    =
    \frac{1}{2} P(x \sim \mathcal{U}(D_0) \suchthat x_1 > -\delta)
    +
    \frac{1}{2} P(x \sim \mathcal{U}(D_1) \suchthat x_1 < \delta).
\end{align*}
Symmetry implies that these two probabilities have the same value, so we only need to compute the first one.
Negating the probability and expanding it as an integral using the density $p_0$ of $\mathcal{D}_0$, we find that
\begin{align*}
    P(x \sim \mathcal{U}(D_0) \suchthat x_1 > -\delta)
    =
    1 - \int_{D_0} \mathbb{I}_{\{x_1 < -\delta\}} p_0(x) dx.
\end{align*}
The bound on the density provided by the SmAC property (Definition~\ref{def:smac}) therefore implies that
\begin{align*}
    P(x \sim \mathcal{U}(D_0) \suchthat x_1 > -\delta)
    \geq
    1 - \frac{A}{V^n} \int_{D_0} \mathbb{I}_{\{x_1 < -\delta\}} dx.
\end{align*}
Since this integral corresponds to the volume of a spherical cap with height smaller than its radius (due to the fact that $1 > \delta > \epsilon > 0$), we may apply Lemma~\ref{lem:capVolumeBound} to show that
\begin{align*}
    \int_{D_0} \mathbb{I}_{\{x_1 < -\delta\}} dx 
    &= 
    V^n_{\operatorname{cap}}(1, 1 - (\epsilon - \delta))
    \\&
    \leq
    \frac{1}{2} V^n (1 - (\delta - \epsilon)^2)^{\frac{n}{2}},
\end{align*}
and the result therefore follows.

\subsection{Proof of Theorem~\ref{thm:undetectability}}\label{sec:undetectabilityProof}

To prove Theorem~\ref{thm:undetectability}, we begin by expanding the probability by conditioning on the label value, finding
\begin{align*}
    &P((x, \ell) \sim \mathcal{D}_{\epsilon}, s \sim \mathcal{U}(\mathbb{B}^n_{\delta}) \suchthat f(x + s) \neq \ell)
    \\&\qquad
    =
    \frac{1}{2}
    P(x \sim \mathcal{U}(D_0), s \sim \mathcal{U}(\mathbb{B}^n_{\delta}) \suchthat x_1 + s_1 > 0)
    \\&\qquad\qquad
    +
    \frac{1}{2}
    P(x \sim \mathcal{U}(D_1), s \sim \mathcal{U}(\mathbb{B}^n_{\delta}) \suchthat x_1 + s_1 < 0).
\end{align*}
The symmetry here implies that these two probabilities are equal, so we proceed by only bounding the first.
Expanding this as an integral using the density $p_0$ of $\mathcal{D}_0$ and the fact that $s$ is sampled from a uniform distribution, we observe that
\begin{align*}
    &P(x \sim \mathcal{U}(D_0), s \sim \mathcal{U}(\mathbb{B}^n_{\delta}) \suchthat x_1 + s_1 > 0)
    \\&\qquad
    =
    \frac{1}{V^n \delta^n} 
    \int_{D_0} \int_{\mathbb{B}^n_\delta} \mathbb{I}_{\{ s_1 > -x_1 \}} 
    ds \, p_0(x) \, dx.
\end{align*}
The bound on the density provided by the SmAC property (Definition~\ref{def:smac}) therefore implies that
\begin{align*}
    &P(x \sim \mathcal{U}(D_0), s \sim \mathcal{U}(\mathbb{B}^n_{\delta}) \suchthat x_1 + s_1 > 0)
    \\&\qquad
    \leq
    \frac{A}{(V^n)^2 \delta^n} 
    \int_{D_0} \int_{\mathbb{B}^n_\delta} \mathbb{I}_{\{ s_1 > -x_1 \}} 
    ds dx.
\end{align*}
For each fixed value of $x$, the inner integral is just computing the volume of a section of a ball with radius $\delta$. When $x_1 > 0$, this volume is at least a hemisphere, while when $x_1 < 0$ this volume is less than a hemisphere.
On the other hand, for fixed $s$, the integral over $x$ is also calculating the volume of a section of the unit ball.
Since the volume of the ball concentrates in high dimensions about its equator, a section which is smaller than a hemisphere may be expected to have small volume, while a section larger than a hemisphere may be expected to have large volume.
This is the intuition we apply here by splitting the integral into two parts: one for $x_1 < -t$ and one for $x_1 \geq -t$ for some arbitrary $t \in [0, \epsilon]$.
In the first part, we will be able to obtain `smallness' in our bound from the fact that we are integrating $s$ over just a spherical cap, while in the second case we are integrating $x$ over a spherical cap.
We write this splitting as 
\begin{align}\label{eq:undetectability:ballSplitting}
    &\int_{D_0} \int_{\mathbb{B}^n_\delta} \mathbb{I}_{\{ s_1 > -x_1 \}} ds dx
    \\&\qquad
    =
    \int_{D_0} \int_{\mathbb{B}^n_\delta} \mathbb{I}_{\{ x_1 < -t \}} \mathbb{I}_{\{ s_1 > -x_1 \}} ds dx
    \notag
    \\&\qquad\quad
    +
    \int_{D_0} \int_{\mathbb{B}^n_\delta} \mathbb{I}_{\{ x_1 > -t \}} \mathbb{I}_{\{ s_1 > -x_1 \}} ds dx.
    \notag
\end{align}
The first term of this splitting may be bounded above by extending the indicator function over $s$ to all those points with $s_1 > t$ (since $-x_1 > t$), which enables us to separate the integrals to find that
\begin{align*}
    &\int_{D_0} \int_{\mathbb{B}^n_\delta} \mathbb{I}_{\{ x_1 < -t \}} \mathbb{I}_{\{ s_1 > -x_1 \}} ds dx
    \\&
    \leq
    \int_{D_0} \mathbb{I}_{\{ x_1 < -t \}} dx \int_{\mathbb{B}^n_\delta} \mathbb{I}_{\{ s_1 > t \}} ds.
\end{align*}
These integrals may be expressed as volumes in the form
\begin{align*}
    &\int_{D_0} \mathbb{I}_{\{ x_1 < -t \}} dx \int_{\mathbb{B}^n_\delta} \mathbb{I}_{\{ s_1 > t \}} ds
    \\&
    =
    (V^n - V^{n}_{\operatorname{cap}}(1, 1 - (\epsilon - t))) V^{n}_{\operatorname{cap}}(\delta, \delta - t),
\end{align*}
and Lemma~\ref{lem:capVolumeBound} and the fact that the volume of a spherical cap is non-negative implies that
\begin{align*}
    \int_{D_0} \mathbb{I}_{\{ x_1 < -t \}} dx \int_{\mathbb{B}^n_\delta} \mathbb{I}_{\{ s_1 > t \}} ds
    \leq
    \frac{1}{2} (V^n)^2 \delta^n \Big(1 - \Big( \frac{t}{\delta} \Big)^2 \Big)^{\frac{n}{2}}.
\end{align*}

Returning to the second integral in~\eqref{eq:undetectability:ballSplitting}, we may similarly bound the indicator function over $s$ from above by simply the constant $1$.
This implies that
\begin{align*}
    \int_{D_0} \int_{\mathbb{B}^n_\delta} &\mathbb{I}_{\{ x_1 > -t \}} \mathbb{I}_{\{ s_1 > -x_1 \}} ds dx
    \leq
    \int_{D_0} \mathbb{I}_{\{ x_1 > -t \}} dx \int_{\mathbb{B}^n_\delta} ds
    \\&
    =
    V^{n}_{\operatorname{cap}}(1, 1 - (\epsilon - t)) V^n \delta^n,
\end{align*}
and Lemma~\ref{lem:capVolumeBound} consequently provides
\begin{align*}
    &\int_{D_0} \int_{\mathbb{B}^n_\delta} \mathbb{I}_{\{ x_1 > -t \}} \mathbb{I}_{\{ s_1 > -x_1 \}} ds dx
    \\&\qquad\qquad
    \leq
    \frac{1}{2} (V^n)^2 \delta^n (1 - (\epsilon - t)^2)^{\frac{n}{2}}.
\end{align*}

Combining these bounds and using the fact that $t \in [0, \epsilon]$ was arbitrary, we find that
\begin{align*}
    &P((x, \ell) \sim \mathcal{D}_{\epsilon}, s \sim \mathcal{U}(\mathbb{B}^n_{\delta}) \suchthat f(x + s) \neq \ell)
    \\&
    \leq
    \frac{1}{2} A
    \inf_{t \in [0, \epsilon]}
    \Big[
        \Big(1 - \Big( \frac{t}{\delta} \Big)^2 \Big)^{\frac{n}{2}}
        +
        (1 - (\epsilon - t)^2)^{\frac{n}{2}}
    \Big],
\end{align*}
and the theorem follows by noting that, for $t = \frac{\epsilon \delta}{1 + \delta}$, the two terms inside the infimum are equal (this choice of $t$ is valid since $\frac{\delta}{1 + \delta} \in [0, 1]$ for $\delta \geq 0$).

\subsection{Proof of Theorem~\ref{thm:gradientBasedAttack}}\label{sec:gradientAttackProof}

When $\ell = 0$, we have $|\tilde{f}(x) - \ell| = \sigma(g(x))$, since $\sigma(t) \in (0, 1)$ for $t \in \mathbb{R}$.
In this case, the attack may therefore be computed as
\begin{align}\label{eq:gradient-attack-proof:attack-direction}
    \textbf{e}_1 \sigma^{\prime}(g(x)) L^{\prime}(\sigma(g(x))),
\end{align}
since $g^{\prime}(x) = \textbf{e}_1$.
Since $\sigma$ and $L$ are assumed to be continuously differentiable and monotonically increasing, the Morse-Sard theorem~\cite{Morse:1939} implies that $\sigma^{\prime}(t), L^{\prime}(t) > 0$ everywhere except on a set of Lebesgue measure zero.
The SmAC property on the distribution $\mathcal{D}_{\epsilon}$ implies that the probability of sampling $x$ from a set of Lebesgue measure zero is zero, and therefore with probability 1 the attack direction~\eqref{eq:gradient-attack-proof:attack-direction} is a positive multiple of $\mathbf{e}_1$ for all $t \in \mathbb{R}$, as required.
Analogously, when $\ell = 1$ we have $|\tilde{f}(x) - \ell| = 1 - \sigma(g(x))$,
and we therefore obtain a negative mutiple of $\mathbf{e}_1$ with probability 1, and the result follows.

\subsection{Proof of Theorem~\ref{thm:universality}}\label{sec:universalityProof}
Since the statement and setup are symmetric with respect to the class label $\ell$, we focus only on the class $0$ and the statement for class $1$ follows analogously.
In this case, the statement that 
\[
    f(x + s) \neq \ell \text{ for all } s \in \mathbb{R}^n \text{ such that } d_{\ell}(z, s) > \gamma,
\]
is implied by the condition that $z_1 < x_1 + \gamma$, and therefore
\begin{align*}
    &
    P( x, z \sim \mathcal{U}(D_0) \suchthat f(x + s) \neq 0 \text{ for all } s \in \mathbb{R}^n 
    \\&\qquad\qquad\qquad\qquad\qquad\qquad\qquad\text{ such that } d_{0}(z, s) > \gamma )
    \\&\qquad\qquad\qquad
    \geq
    P( x, z \sim \mathcal{U}(D_0) \suchthat z_1 < x_1 + \gamma ).
\end{align*}
Introducing $t \in [0, \gamma]$, this probability is clearly at least
\begin{align*}
    P( x, z \sim \mathcal{U}(D_0) \suchthat z_1 < x_1 + \gamma \text{ and } x_1 > -\epsilon -t).
\end{align*}
We note that if $x_1 > - \frac{1}{2}\gamma - \epsilon$ and $z_1 < \frac{1}{2}\gamma - \epsilon$ then it follows that $z_1 < x_1 + \gamma$, and therefore
\begin{align*}
    &P( x, z \sim \mathcal{U}(D_0) \suchthat z_1 < x_1 + \gamma )
    \geq
    \\&\quad
    P( x, z \sim \mathcal{U}(D_0) \suchthat z_1 < \frac{1}{2}\gamma - \epsilon \text{ and } x_1 > - \frac{1}{2}\gamma - \epsilon).
\end{align*}
This last probability has the property of involving two events which separately depend on the independent variables $x$ and $z$, and may therefore be expressed as the product
\begin{align*}
    P( z \sim \mathcal{U}(D_0) \suchthat z_1 < \frac{1}{2}\gamma - \epsilon) P( x \sim \mathcal{U}(D_0) \suchthat x_1 > - \frac{1}{2}\gamma - \epsilon).
\end{align*}
Negating both of these probabilities and recalling the bound on the density provided by Definition~\ref{def:smac} with $r = 1$, we obtain the lower bounds
\begin{align*}
    P( z \sim \mathcal{U}(D_0) \suchthat z_1 < \frac{1}{2}\gamma - \epsilon)
    \geq
    1 - \frac{A}{V^n} 
    \int_{D_0} 
    \mathbb{I}_{\{ z_1 < \frac{1}{2}\gamma - \epsilon \}}
    dz,
\end{align*}
and
\begin{align*}
    P( x \sim \mathcal{U}(D_0) \suchthat x_1 > -\epsilon - \frac{1}{2}\gamma)
    \geq
    1 - \frac{A}{V^n} 
    \int_{D_0} 
    \mathbb{I}_{\{ x_1 < -\epsilon - \frac{1}{2}\gamma \}}
    dz.
\end{align*}
Since $D_0$ is simply a ball for which the centre has first coordinate $-\epsilon$, it follows that both of these integrals are simply the volume of a spherical cap, with value $V^n_{\operatorname{cap}}(1, 1 - \frac{1}{2} \gamma)$, and using Lemma~\ref{lem:capVolumeBound}, we therefore conclude that
\begin{align*}
    P( x, z \sim \mathcal{U}(D_0) \suchthat z_1 < x_1 + \gamma ) \geq  \Big(1 - A  \Big(1 - \frac{\gamma^2}{4} \Big)^{\frac{n}{2}} \Big)^2
\end{align*}

\section{Proofs of results for the two half balls model in Section~\ref{sec:hemispheres_model}}

\subsection{Proof of Theorem~\ref{thm:hemisphere:susceptibility}}\label{sec:supplementary:hemisphere:susceptibility-proof}
Using the definition of the classification function $f$, we may rewrite the probability in question as
\begin{align*}
    P\big(x &\sim \mathcal{D}_\epsilon \suchthat \text{there exists } s \in \mathbb{R}^n \text{ with } \|s\| \leq \delta 
    \\&
    \text{ such that } f(x + s) \neq f(x)\big)
    =
    P(x \sim \mathcal{D}_\epsilon \suchthat |x_1| < \delta).
\end{align*}
Expanding the probability as an integral, and using the fact that $\mathcal{D}_{\epsilon}$ is a uniform distribution over two disjoint half-balls and therefore has density $(V^n)^{-1}$, we may further express this as
\begin{align*}
    P(x \sim \mathcal{D}_\epsilon \suchthat |x_1| < \delta)
    &=
    \frac{1}{V^n}
    \Big(
        \int_{D_0} \mathbb{I}_{\{x \suchthat -\delta < x_1 < -\epsilon\}} dx 
        \\&\qquad
        + 
        \int_{D_1} \mathbb{I}_{\{x \suchthat \epsilon < x_1 < \delta\}} dx
    \Big),
\end{align*}
and the remaining problem is to compute the two remaining integrals, the values of which are equal by symmetry.
We may express the set $\{x \in D_1 \suchthat \epsilon < x_1 < \delta\}$, which geometrically represents the slab of the half-ball $D_1$ within $\delta - \epsilon$ distance of its planar face, as the complement of a spherical cap, implying
\begin{align*}
    \int_{D_1} \mathbb{I}_{\{x \suchthat \epsilon < x_1 < \delta\}} dx
    =
    \frac{1}{2} V^n - V^n_{\operatorname{cap}}(1, 1 - (\delta - \epsilon)),
\end{align*}
and therefore
\begin{align*}
    P(x \sim \mathcal{D}_\epsilon \suchthat |x_1| < \delta)
    =
    1 - \frac{2V^n_{\operatorname{cap}}(1, 1 - (\delta - \epsilon))}{V^n}.
\end{align*}
We may further estimate this from below, to show the exponential behaviour of this quantity with respect to $n$, by enveloping the spherical cap within a small half-ball.
The Pythagorean theorem implies that
\begin{align*}
    \{x \in D_1 \suchthat \epsilon < x_1 < \delta\} \subset G
\end{align*}
where
\begin{align*}
    G = \{ x \in \mathbb{R}^n \suchthat x_1 > \epsilon \text{ and } \|x - \epsilon \, \mathbf{e}_1\|^2 \leq 1 - (\delta - \epsilon)^2 \},
\end{align*}
and therefore
\begin{align*}
    \frac{2V^n_{\operatorname{cap}}(1, 1 - (\delta - \epsilon))}{V^n}
    \leq
    (1 - (\delta - \epsilon)^2)^{n/2},
\end{align*}
which proves the theorem.

\subsection{Proof of Theorem~\ref{thm:hemisphere:undetectability}}
\label{sec:supplementary:hemisphere:undetectability-proof}
Since $D_0$ and $D_1$ are disjoint half-balls of a unit ball, it follows that the density associated with $\mathcal{D}_\epsilon$ is simply $(V^n)^{-1}$, while the density associated with $\mathcal{U}(\mathbb{B}^n_{\delta})$ is $(\delta^n V^n)^{-1}$.
Writing the probability as an integral with this density, we therefore find that
\begin{align*}
    &P\big(x \sim \mathcal{D}_\epsilon, s \sim \mathcal{U}(\mathbb{B}^n_\delta) \suchthat f(x + s) \neq f(x)\big)
    \\&\qquad
    =
    \frac{1}{\delta^n (V^n)^2} \Big(
        \int_{D_0} \int_{\mathcal{B}^n_{\delta}} \mathbb{I}_{\{x, s \suchthat s_1 > -x_1 \}} ds dx
        \\&\qquad\qquad\qquad\qquad
        +
        \int_{D_1} \int_{\mathcal{B}^n_{\delta}} \mathbb{I}_{\{x, s \suchthat s_1 < -x_1 \}} ds dx
    \Big).
\end{align*}
Since the values of these two integrals are equal by symmetry, we proceed by only estimating the first.
For $x \in D_0$, we have $x_1 < -\epsilon$, and therefore 
\begin{align*}
    \int_{D_0} \int_{\mathcal{B}^n_{\delta}} \mathbb{I}_{\{x, s \suchthat s_1 > -x_1 \}} ds dx
    &\leq
    \int_{D_0} dx \int_{\mathcal{B}^n_{\delta}} \mathbb{I}_{\{s \suchthat s_1 > \epsilon \}} ds 
    \\&
    =
    \frac{1}{2} V^n \int_{\mathcal{B}^n_{\delta}} \mathbb{I}_{\{s \suchthat s_1 > \epsilon \}} ds.
\end{align*}
The remaining integral over $s$ now takes the form of the volume of a spherical cap, which we may bound by enveloping the cap in a small half-ball.
Arguing as in the proof of Theorem~\ref{thm:hemisphere:susceptibility}, it follows that
\begin{align*}
    \int_{\mathcal{B}^n_{\delta}} \mathbb{I}_{\{s \suchthat s_1 > \epsilon \}} ds
    &=
    V^n_{\operatorname{cap}}(\delta, (\delta^2 - \epsilon^2)^{1/2})
    \\&
    \leq
    \frac{1}{2} \delta^n V^n \Big(1 - \Big( \frac{\epsilon}{\delta} \Big)^2 \Big)^{n/2}.
\end{align*}
The result therefore follows by combining the components above.

\subsection{Proof of Theorem \ref{thm:hemisphere:no_hiding}}\label{sec:supplementary:hemisphere:nohide-proof}

Consider the hyperplane $h$ passing through the origin and whose normal is ${\bf{e}}_1$. This hyperplane is the decision boundary of the classifier $f$ (see (\ref{eq:hemisphere:classifier})) which separates $D_0$ and $D_1$ in the sense that, with probability one, the classifier assigns correct labels to samples drawn from $\mathcal{D}_\epsilon$, $\epsilon=0$.

Pick any $\Delta\in(0,1)$. The probability $p$ that a point $x\sim\mathcal{D}_\epsilon$ lands within the $\Delta$-distance from the hyperplane $h$ is bounded from below as:
\[
p > 1 - (1-\Delta^2)^{\frac{n}{2}}.
\]
Conversely, if one picks the value of $p\in(0,1)$, then the value of $\Delta$ corresponding to this probability must satisfy:
\[
\Delta<(1-(1-p)^{\frac{2}{n}})^{\frac{1}{2}}=\rho(p,n).
\]
In what follows, we are interested in the event
\[
E_1(x,s,\delta,n): \ f(x+s)\neq f(x), \ s\sim\mathcal{U}(\mathbb{B}_\delta^n).
\]

It is clear that the event
\[
E_2 (x,s,\delta,\Delta,n): \ f(x+s)\neq f(x),  \  \ |x\cdot {\bf e}_1| \leq \Delta, \  s\sim\mathcal{U}(\mathbb{B}_\delta^n)
\]
implies event $E_1(x,s,\delta,n)$.  Hence
\begin{equation}\label{eq:prob_bounds:1}
\begin{split}
& P(f(x+s)\neq f(x), \ s\sim\mathcal{U}(\mathbb{B}_\delta^n)) \geq \\
& P(f(x+s)\neq f(x), \ s\sim\mathcal{U}(\mathbb{B}_\delta^n) \ \mbox{and} \ |x\cdot {\bf e}_1|\leq \Delta)\\
&=P(f(x+s)\neq f(x), \ s\sim\mathcal{U}(\mathbb{B}_\delta^n) \given |x\cdot {\bf e}_1|\leq \Delta) p,
\end{split}
\end{equation}
where the last equality follows from the definition of the conditional probability and the fact that $p=P(|x\cdot {\bf e}_1|\leq \Delta)$ is the probability of $x\sim\mathcal{D}_\epsilon$ landing within the $\Delta$-distance from the hyperplane $h$. 

Consider the event
\[
\begin{split}
& E_3(x,s,\delta,p,n): \\
&  \left[0\leq  x\cdot {\bf e}_1 \leq \Delta \ \mbox{and} \ s\cdot {\bf e}_1 \leq - \rho(p,n)\right] \ \mbox{or} \\
&  \ \ \ \ \ \ \ \ \ \ \ \ \ \ \ \ \ \ \ \ \left[-\Delta \leq   x\cdot {\bf e}_1 < 0  \ \mbox{and} \ s\cdot {\bf e}_1 \geq \rho(p,n)\right].
\end{split}
\]

Given that $\Delta<\rho(p,n)$, the event $E_3 (x,s,\delta,p,n)$ implies $E_2(x,s,\delta,\Delta,n)$. Hence taking (\ref{eq:prob_bounds:1}) into account, the following holds true:
\[
P(E_1(x,s,\delta,n))\geq P(E_2(x,s,\delta,\Delta,n)) \geq P(E_3(x,s,\delta,p,n)).
\]
Therefore, a lower bound for $P(E_3(x,s,\delta,p,n))$ is also a lower bound for $P(E_1(x,s,\delta,n))$.  

Noticing that $x$ and $s$ are independent, we obtain
\[
\begin{split}
P(E_3(x,s,\delta,p,n))=&P(0\leq  x\cdot {\bf e}_1 \leq \Delta)P (s\cdot {\bf e}_1 \leq - \rho(p,n)) +\\
&P(-\Delta \leq   x\cdot {\bf e}_1 < 0) P(s\cdot {\bf e}_1 \geq \rho(p,n))\\
= &\frac{p}{2} P (s\cdot {\bf e}_1 \leq - \rho(p,n)) + \frac{p}{2} P(s\cdot {\bf e}_1 \geq \rho(p,n)).
\end{split}
\]
Observe that the symmetry of $\mathcal{U}(\mathbb{B}_{\delta}^n)$ implies $P(s\cdot {\bf e}_1 \geq \rho(p,n))=P (s\cdot {\bf e}_1 \leq - \rho(p,n))$, and hence
\[
P(E_3(x,s,\delta,p,n))=p P(s\cdot {\bf e}_1 \geq \rho(p,n)).
\]

Let us now bound the probability of $E_3(x,s,\delta,p,n)$ from below. 

{\it Case 1}: $0<\delta\leq \rho(p,n)$. In this case
\[
f(x+s)=f(x) \ \mbox{for all} \ s\in\mathbb{B}_\delta^n,
\]
and hence $P(E_3(x,s,\delta,p,n))=0$.

{\it Case 2}: $\delta>\rho(p,n)$. The probability of $P(s\cdot {\bf e}_1 \geq \rho(p,n))$ is the ratio 
\[
\frac{V^n_{\mathrm{cap}}(\delta,\delta-\rho(p,n))}{V^n_{\mathrm{cap}}(\delta,\delta)},
\]
where $V^n_{\mathrm{cap}}(\delta,\delta-\rho(p,n))$ is the volume of the spherical cap whose radius is $\delta$ and whose height is $\delta-\rho(p,n)$. 

Consider $V^n_{\mathrm{cap}}(\delta,\delta-\rho(p,n))$ \cite{li2010concise}:
\[
V^n_{\mathrm{cap}}(\delta,\delta-\rho(p,n))=\frac{\pi^{(n-1)/2}}{\Gamma(\frac{n-1}{2}+1)}\delta^n \int_{0}^{\cos^{-1}(\rho(p,n)/\delta)} \sin^n(\theta) d\theta.
\]
Rewriting the integral through the change of variables $t=\cos(\theta)$ results in
\[
V^n_{\mathrm{cap}}(\delta,\delta-\rho(p,n))=\frac{\pi^{(n-1)/2}}{\Gamma(\frac{n-1}{2}+1)}\delta^n \int_{\rho(p,n)/\delta}^1 (1-t^2)^{\frac{n-1}{2}}dt,
\]
and hence
\begin{equation}\label{eq:spherical_cap_prob}
P(s\cdot {\bf e}_1 \geq \rho(p,n))=\frac{\int_{\rho(p,n)/\delta}^1 (1-t^2)^{\frac{n-1}{2}}dt}{\int_{0}^1 (1-t^2)^{\frac{n-1}{2}}dt}.
\end{equation}

Let us now bound the integral 
\begin{equation}\label{eq:integral_to_bound}
\int_{\rho(p,n)/\delta}^1 (1-t^2)^{\frac{n-1}{2}}dt
\end{equation}
from below. First, observe that
\begin{equation}\label{eq:intermediate_bound}
(1-t^2) \geq (1 - t^2 \alpha + \frac{t^4 \alpha^2}{2})
\end{equation}
for any $\alpha>1$ and
\[
0<t \leq \frac{\sqrt{2(\alpha-1)}}{\alpha}.
\]
At the same time, using Taylor's theorem we see that if $t, \alpha > 0$ there exists a $c \in (0,t\alpha^2)$ so that
\begin{equation}\label{eq:gaussian_bound}
\begin{split}
 & e^{-\alpha t^2} = 1 - t^2 \alpha + \frac{t^4 \alpha^2}{2} - \frac{e^{-c} t^6 \alpha^3}{3!} <  1 - t^2 \alpha + \frac{t^4 \alpha^2}{2}
 \\
 & 1 - t^2 \alpha + \frac{t^4 \alpha^2}{2} > e^{-\alpha t^2}. 
 \end{split}
\end{equation}

Applying the same argument, one can conclude that for any $p\in(0,1)$  and all $n\geq 1$
the following holds true:
\[
(1-p)^{\frac{2}{n}}=e^{\log(1-p) \frac{2}{n}}> 1 + \log(1-p)\frac{2}{n}.
\]
This implies that
\[
\rho(p,n)< \frac{\sqrt{2|\log(1-p)|}}{\sqrt{n}}
\]
for all $n\geq 1$.

Let
\[
\tau(\alpha)=\frac{\sqrt{2(\alpha-1)}}{\alpha},
\]
$N$ be defined by
\[
N(\alpha,p,\delta):= \max\left\{1,\frac{2|\log(1-p)|}{\delta^2 \tau(\alpha)^2 }\right\},
\]
and \[\beta(p,\delta,n):=\frac{\sqrt{2|\log(1-p)|}}{\sqrt{n}\delta}.\]
Suppose that $n > N(\alpha,p,\delta)$. Then we must have
\begin{equation*}\label{eq:choice:1}
\frac{\rho(p,n)}{\delta}<\beta(p,\delta,n)<\tau(\alpha).
\end{equation*}

In particular, for $n>N(\alpha,p,\delta)$, the integral (\ref{eq:integral_to_bound}) can be bounded from below as
\begin{equation}\label{eq:integral_bound}
\int_{\rho(p,n)/\delta}^1 (1-t^2)^{\frac{n-1}{2}}dt > \int_{\beta(p,\delta,n)}^{\tau(\alpha)} (1-t^2)^{\frac{n-1}{2}}dt
\end{equation}

Pick an 
\[
n>N(\alpha,p,\delta),
\]
and consider the right-hand-side of (\ref{eq:spherical_cap_prob}). According to (\ref{eq:integral_bound})
\[
P(s\cdot {\bf e}_1 \geq \rho(p,n)) > \frac{\int_{\beta(p,\delta,n)}^{\tau(\alpha)} (1-t^2)^{\frac{n-1}{2}}dt}{\int_{0}^1 (1-t^2)^{\frac{n-1}{2}}dt}.
\]
Invoking (\ref{eq:intermediate_bound}) and (\ref{eq:gaussian_bound}) we arrive at 
\[
P(s\cdot {\bf e}_1 \geq \rho(p,n)) > \frac{\int_{\beta(p,\delta,n)}^{\tau(\alpha)} e^{-\alpha  \frac{t^2(n-1)}{2}}dt}{\int_{0}^1 (1-t^2)^{\frac{n-1}{2}}dt}.
\]
Changing the integration variable as  $t {\sqrt{n-1} =\xi}$ yields:
\[
P(s\cdot {\bf e}_1 \geq \rho(p,n))
> \frac{\int_{\beta(p,\delta,n)\sqrt{n-1}}^{\tau(\alpha)\sqrt{n-1}} e^{-\alpha  \frac{\xi^2}{2}}d\xi}{\int_{0}^{\sqrt{n-1}} \left(1-\frac{\xi^2}{n-1}\right)^{\frac{n-1}{2}}d\xi}.
\]
Note that (cf. \cite{gorban2016approximation}, p. 135, inequality (23))
\[
\left(1-\frac{\xi^2}{(n-1)}\right)^{{n-1}}\leq e^{- \xi^2}
\]
for all $n>1$, $n\in\Natural$ and any $\xi^2 \geq 0$.

Therefore 
\[
P(s\cdot {\bf e}_1 \geq \rho(p,n))> \frac{\int_{\beta(p,\delta,n)\sqrt{n-1}}^{\tau(\alpha)\sqrt{n-1}} e^{-\alpha  \frac{\xi^2}{2}}d\xi}{\int_{0}^{\sqrt{n-1}} e^{-\frac{\xi^2}{2}}d\xi}.
\]
for all $n>N(\alpha,p,\delta)$. Changing the integration variable yet again in the top integral as $\zeta=\sqrt{\alpha}\xi$ and pre-multiplying both the nominator and the denominator by $1/\sqrt{2\pi}$ results in:
\begin{equation}\label{eq:lower_boundr_raw}
P(s\cdot {\bf e}_1 \geq \rho(p,n))> \frac{\frac{1}{\sqrt{\alpha}} \frac{1}{\sqrt{2\pi}}\int_{\sqrt{\alpha}\beta(p,\delta,n)\sqrt{n-1}}^{\sqrt{\alpha}\tau(\alpha)\sqrt{n-1}} e^{-\frac{\zeta^2}{2}}d\zeta}{\frac{1}{\sqrt{2\pi}}\int_{0}^{\sqrt{n-1}} e^{-\frac{\xi^2}{2}}d\xi}.
\end{equation}

Recalling the standard cumulative distribution function 
\[
\Phi(s)=\frac{1}{\sqrt{2\pi}}\int_{-\infty}^{s} e^{-\frac{\xi^2}{2}}d\xi,
\]
the right-hand side of (\ref{eq:lower_boundr_raw}) becomes
\[
\frac{\frac{1}{\sqrt{\alpha}}\left(\Phi\left(\sqrt{\alpha}\tau(\alpha)\sqrt{n-1}\right)-\Phi\left(\sqrt{\alpha}\beta(p,\delta,n)\sqrt{n-1}\right)\right)}{\Phi(\sqrt{n-1})-\frac{1}{2}}.
\]

Therefore, for any $p\in(0,1)$, $\delta>0$, $\alpha>1$, and $n> N(\alpha,p,\delta)$
\[
\begin{split}
&P(E_3(x,s,\delta,p,n)) >\\
& \frac{p\left(\Phi\left(\sqrt{\alpha}\tau(\alpha)\sqrt{n-1}\right)-\Phi\left(\sqrt{\alpha}\beta(p,\delta,n)\sqrt{n-1}\right)\right)}{\sqrt{\alpha}\left(\Phi(\sqrt{n-1})-\frac{1}{2}\right)}.
\end{split}
\]

For any fixed $\alpha>1$ and $n\rightarrow\infty$, the right-hand side of the above expression reduces to
\[
\frac{p\frac{1}{\sqrt{\alpha}}\left(1-\Phi\left(\sqrt{\alpha}\frac{\sqrt{2|\log(1-p)|}}{\delta}\right)\right)}{\frac{1}{2}}.
\]
Given that the value of $\alpha$ can be chosen arbitrarily in $(1,\infty)$, in the limit 
\[
\begin{split}
&\lim_{n\rightarrow\infty} P(E_3(x,s,\delta,p,n))\geq\\
& \ \ \ \ \ \ \ \ \  \sup_{\alpha>1}\frac{2p}{\sqrt{\alpha}} \left(1-\Phi\left(\frac{\sqrt{\alpha}\sqrt{2|\log(1-p)|}}{\delta}\right)\right)\\
& \ \ \ \ \ \ \ \ \  = 2 p\left(1-\Phi\left(\frac{\sqrt{2|\log(1-p)|}}{\delta}\right)\right), \ \delta>0.
\end{split}
\]

Finally, taking $\sup$ over $p\in(0,1)$, results in the following asymptotic bound:
\[
\sup_{p\in(0,1)}  2 p \left(1-\Phi\left(\frac{\sqrt{2|\log(1-p)|}}{\delta}\right)\right).
\]
$\square$

\section{Proofs of results for the general model in Section~\ref{sec:general-model}}

\subsection{Accuracy of the general model: Proof of Theorem~\ref{thm:supplementary:generalisation:accuracy} and Corollary~\ref{corr:supplementary:generalisation:accuracy}}\label{sec:supplementary:generalisation:accuracyProof}

\subsubsection{Proof of Theorem~\ref{thm:supplementary:generalisation:accuracy}}

We can measure the accuracy of the classifier for class $0$ as
\begin{align*}
    \accuracy_0(f) = P(x \sim \distribution \suchthat f(x) = 0) = P(x \sim \distribution \suchthat d_S(x) \leq 0).
\end{align*}
For any $t \in \mathbb{R}$, the condition that $d_S(x) \leq t$ can be rewritten as
\begin{align*}
    d_{\pi}(x) - \sfunc(\proj x) \leq t,
\end{align*}
which is implied by the condition that
\begin{align*}
    d_{\pi}(x) + |\sfunc(\proj x)| \leq t.
\end{align*}
Introducing the events
\begin{equation}\label{eq:summplementary:generalisation:eventsAB}
\begin{aligned}
A(\alpha) &\suchthat \quad x \sim \distribution \text{ is such that } |\sfunc(\proj x)| \leq \alpha,
\\
B(\beta) &\suchthat \quad x \sim \distribution \text{ is such that } d_{\pi}(x) \leq \beta,
\end{aligned}
\end{equation}
parameterised by the arbitrary values $\alpha \geq 0$, $\beta \in \Real$, we find that
\begin{align}\label{eq:supplementary:generalisation:combiningEvents}
    A(\alpha) \wedge B(t - \alpha)
    \quad \Rightarrow \quad
    d_{\pi}(x) + |\sfunc(\proj x)| \leq t.
\end{align}
Putting these pieces together, we find that
\begin{align*}
    &\accuracy_0(f) 
    \geq 
    P(x \sim \distribution \suchthat d_{\pi}(x) + |\sfunc(\proj x)| \leq 0)
    \\&
    \geq 
    P(x \sim \distribution \suchthat A(\alpha) \wedge B(-\alpha)).
\end{align*}
Negating this event and applying the union bound, we therefore find that
\begin{align*}
    \accuracy_0(f) 
    &\geq 
    P(x \sim \distribution \suchthat |\sfunc(\proj x)| \leq \alpha)
    \\&\qquad
    -
    P(x \sim \distribution \suchthat d_{\plane}(x) > -\alpha).
\end{align*}
Since $\alpha \geq 0$ was arbitrary, it therefore follows that
\begin{align*}
    \accuracy_0(f) 
    \geq 
    \sup_{\alpha \geq 0}
    \big[
        &P(x \sim \distribution \suchthat |\sfunc(\proj x)| \leq \alpha)
        \\&\,
        -
        P(x \sim \distribution \suchthat d_{\plane}(x) > -\alpha)
    \big].
\end{align*}

\subsubsection{Proof of Corollary~\ref{corr:supplementary:generalisation:accuracy}}
Since $\sfunc \equiv 0$ in this case, it follows that
\begin{align*}
    P(x \sim \mathcal{E} \suchthat |\phi(x)| \leq \alpha) = 1,
\end{align*}
for all $\alpha \geq 0$.
We may therefore take $\alpha = 0$ in the second term of Theorem~\ref{thm:supplementary:generalisation:accuracy}, and we proceed by bounding
\begin{align*}
    P(x \sim \mathcal{E} \suchthat d_{\plane} (x) \geq 0)
\end{align*}
from above.

Recalling the bound on the density $p$ of $\mathcal{E}$ in Definition~\ref{def:smac},
we have
\begin{align*}
    P(x \sim \mathcal{E} \suchthat d_{\plane} (x) \geq 0)
    \leq
    \frac{A}{V^nr^n} 
    \int_{\mathbb{B}^n_r(c)} \mathbb{I}_{\{ x \suchthat d_{\plane}(x) \geq 0 \}} dx,
\end{align*}
and the definition of $d_{\plane}$ implies that this is
\begin{align*}
    \int_{\mathbb{B}^n_r(c)} \mathbb{I}_{\{ x \suchthat d_{\plane}(x) \geq 0 \}} dx
    =
    \int_{\mathbb{B}^n_r(c)} \mathbb{I}_{\{ x \suchthat (x - w) \cdot \planenormal \geq 0 \}} dx,
\end{align*}
which is zero for $(w - c) \cdot \planenormal > r$ and simply a spherical cap otherwise.
Note that the assumption that $d_{\plane}(c) = -\eta$ for some $\eta > 0$ implies that this spherical cap is less than half the ball $\mathbb{B}^n_r(c)$.
Therefore, Lemma~\ref{lem:capVolumeBound} implies that
\begin{align*}
    P(x \sim \mathcal{E} \suchthat f(x) = 0)
    \geq
    1
    -
    \frac{1}{2}A \Big(1 - \Big(\frac{\eta}{r}\Big)^2\Big)^{\frac{n}{2}}
\end{align*}

\subsection{Susceptibility to adversarial perturbations of the general model: Proof of Theorem~\ref{thm:supplementary:generalisation:susceptibility} and Corollary~\ref{corr:supplementary:generalisation:susceptibility}}\label{sec:supplementary:generalisation:susceptibilityProof}

\subsubsection{Proof of Theorem~\ref{thm:supplementary:generalisation:susceptibility}}

The susceptibility of points sampled from class $0$ to an adversarial attack with Euclidean norm $\delta$ may be measured analogously using the function
\begin{align*}
    \susceptibility_0(f) 
    =
    P(x \sim \distribution \suchthat \text{ there exists } s \in \mathbb{B}^n_\delta \text{ with } f(x + s) \neq 0).
\end{align*}
The set of points $x$ satisfying the condition in this probability may be seen to be those contained in the union $R \cup T$ of the disjoint sets
\begin{align*}
    R = \{ x \in \mathbb{R}^n \suchthat d_S(x) > 0 \}
\end{align*}
and
\begin{align*}
    T = \{ x \in \mathbb{R}^n \suchthat d_S(x) \leq 0 \text{ and } \sigma(x) \leq \delta \};
\end{align*}
in the first case, since these points are already misclassified it follows that $f(x + s) \neq 0$ for $s = 0 \in \mathbb{B}^n_{\delta}$, while in the second case the points are correctly classified but they lie within Euclidean distance $\delta$ of the decision surface $S$, due to the definition of $\sigma$.
To simplify this condition slightly, we observe that 
\begin{align*}
    \{ x \in \mathbb{R}^n \suchthat d_S(x) \geq -\delta \} \subset R \cup T,
\end{align*}
and therefore
\begin{align*}
    \susceptibility_0(f) 
    \geq
    P(x \sim \distribution \suchthat d_S(x) \geq -\delta)
\end{align*}
Arguing as above, we have
\begin{align*}
    d_S(x) = d_{\pi}(x) - \sfunc(\proj x) \geq -\delta,
\end{align*}
which is implied by the condition that
\begin{align*}
    |\sfunc(\proj x)| - d_{\pi}(x) \leq \delta.
\end{align*}
Recalling the events $A(\alpha)$ and $B(\beta)$ from~\eqref{eq:summplementary:generalisation:eventsAB}, we see that for any $\alpha \geq 0$ this event is in turn implied by the event
\begin{align*}
    A(\alpha) \wedge \notword B(\alpha - \delta),
\end{align*}
from which it follows that
\begin{align*}
    \susceptibility_0(f)
    \geq
    P(x \sim \distribution \suchthat A(\alpha) \wedge \notword B(\alpha - \delta)),
\end{align*}
and negating this event and applying the union bound therefore implies that
\begin{align*}
    \susceptibility_0(f)
    \geq
    P(x \sim \distribution \suchthat A(\alpha)) - P(x \sim \distribution \suchthat B(\alpha - \delta)),
\end{align*}
and, since $\alpha \geq 0$ was arbitrary,
\begin{align*}
    \susceptibility_0(f)
    \geq
    \sup_{\alpha \geq 0}
    \big[
    &P(x \sim \distribution \suchthat |\sfunc(\proj x)| < \alpha) 
    \\&\,
    - P(x \sim \distribution \suchthat d_{\pi}(x) < \alpha - \delta)
    \big].
\end{align*}

\subsubsection{Proof of Corollary~\ref{corr:supplementary:generalisation:susceptibility}}

To prove the Corollary, we start from the result of Theorem~\ref{thm:supplementary:generalisation:susceptibility}.
Setting $\sfunc \equiv 0$ and selecting $\alpha = 0$, we find that
\begin{align*}
    &P(x \sim \mathcal{E} \suchthat \text{ there exists } s \in \mathbb{B}^n_\delta \text{ with } f(x + s) \neq 0)
    \\&
    \geq
    1 -
    P(x \sim \mathcal{E} \suchthat d_{\pi}(x) < - \delta).
\end{align*}
To prove the result, we therefore bound this final term on the right from above.

Recalling the bound on the density $p$ of $\mathcal{E}$ in Definition~\ref{def:smac},
we have
\begin{align*}
    P( x \sim \mathcal{E} \suchthat d_{\plane}(x) < -\delta )
    \leq
    \frac{A}{V^n r^n} 
    \int_{\mathbb{B}^n_r(c)} \mathbb{I}_{\{ x \suchthat d_{\plane}(x) < -\delta \}} dx.
\end{align*}
Here, the assumption that $\delta \in (\eta, r]$ implies that this integral is over a spherical cap which is smaller than a hemisphere, and so we conclude that
\begin{align*}
    P( x \sim \mathcal{E} \suchthat d_{\plane}(x) < -\delta )
    \leq
    \frac{1}{2} A \Big( 1 - \Big( \frac{\delta - \eta}{r} \Big)^2 \Big)^{\frac{n}{2}},
\end{align*}
and the result follows.

\subsection{Probability of sampling misclassifying random perturbations for the general model: Proof of Lemma~\ref{lem:supplementary:generalisation:sLowerBound}, Theorem~\ref{thm:supplementary:generalisation:undetectability} and Corollary~\ref{corr:supplementary:generalisation:undetectability}}\label{sec:supplementary:generalisation:undetectabilityProof}

\subsubsection{Proof of Lemma~\ref{lem:supplementary:generalisation:sLowerBound}}
Geometrically, for any point $x \in \mathbb{R}^n$, the Lipschitz condition on $\sfunc$ defines a cone $C(x)$ of points $y$ such that $y \in C(x)$ implies that $d_S(y) \leq 0$, where
\begin{align*}
    C(x) = \{ y \in \mathbb{R}^n \suchthat  (y - \sproj(x)) \cdot \planenormal \leq m \|y - \sproj(x)\| \},
\end{align*}
with $m = - \cos \theta$ and $\theta = \arctan(L^{-1})$.

Suppose that $z \in \mathbb{R}^n$ is a point which $f$ classifies as class 0, i.e. such that $d_S(z) \leq 0$.
The Lipschitz condition on $\sfunc$ provides a cone of points $C(z)$ containing $z$ which are guaranteed to also be assigned class $0$ by $f$.
This allows us to use a geometric argument to find a lower bound on $\sigma$ in terms of $d_S$.
Placing a ball $\mathbb{B}^n_{\epsilon}(z)$ of radius $\epsilon$ around $z$ for some $\epsilon \geq 0$, we can observe that the cone $C(z)$ is tangent to this ball when $\epsilon = |d_S(z)| \sin \theta$.
This is due to the fact that $z$ lies on the central axis of $C(z)$ (which is oriented in the direction of $\planenormal$) and $|d_S(z)|$ therefore measures the distance from $z$ to the vertex of $C(z)$.
This therefore implies the lower bound that
\begin{align*}
    \sigma(z) \geq |d_S(z)| \sin \theta,
\end{align*}
which we may view as the companion to the upper bound~\eqref{eq:generalisation:sBound}.
This allows us to control $\sigma$ from below using $d_S$, which would not have been possible without such a regularity condition on the surface $S$.

\subsubsection{Proof of Theorem~\ref{thm:supplementary:generalisation:undetectability}}

Define the probability of randomly sampling an adversarial perturbation as
\begin{align*}
    \undetectability_0(f)
    =
    P(x \sim \distribution, s \sim \mathcal{U}(\mathbb{B}^n_{\delta}) \suchthat f(x + s) \neq 0),
\end{align*}
for fixed $\delta$ as in the statement of the theorem.
If $x$ is correctly classified, and sampled with $\sigma(x) > \delta$ then there is no possibility of sampling a perturbation $s$ which can destabilise it.
We may therefore ignore these points, implying that
\begin{align*}
    \undetectability_0(f)
    &=
    P(x \sim \distribution, s \sim \mathcal{U}(\mathbb{B}^n_{\delta}) \suchthat f(x + s) \neq 0 
    \\&\qquad\qquad
    \text{ and } (\sigma(x) \leq \delta \text{ or } f(x) \neq 0)).
\end{align*}
Recalling the definition of $f$, we can rewrite this as
\begin{align*}
    \undetectability_0(f)
    &=
    P(x \sim \distribution, s \sim \mathcal{U}(\mathbb{B}^n_{\delta}) \suchthat d_S(x + s) > 0 
    \\&\qquad\qquad
    \text{ and } (\sigma(x) \leq \delta \text{ or } d_S(x) > 0)).
\end{align*}
To obtain an upper bound, we slightly refine this splitting of the points $x$ by treating those points which are very close to the decision boundary along with those which are already misclassified.
Specifically, let $t \in [0, \delta]$ and introduce
\begin{align*}
    K = \{ x \in \mathbb{R}^n \suchthat d_S(x) > 0 \text{ or } \sigma(x) \leq t \},
\end{align*}
which contains those points which are misclassified by $f$ alongside those points which are correctly classified but very close to the decision boundary, 
and
\begin{align*}
    U = \{ x \in \mathbb{R}^n \suchthat d_S(x) \leq 0 \text{ and } \sigma(x) \in (t, \delta] \},
\end{align*}
which contains the correctly classified points which are in a small strip close to, but separated from, the decision boundary.
Since these two sets are disjoint and between them contain all of the points which are susceptible to a perturbation of size $\delta$, we have
\begin{align}\label{eq:supplementary:generalisation:undetectabilitySplitting}
    &\undetectability_0(f) 
    \\&
    =
    P(x \sim \mathcal{D}, s \sim \mathcal{U}(\mathbb{B}^n_{\delta}) \suchthat x \in K \text{ and } f(x + s) \neq 0)
    \notag
    \\&\qquad
    +
    P(x \sim \mathcal{D}, s \sim \mathcal{U}(\mathbb{B}^n_{\delta}) \suchthat x \in U \text{ and } f(x + s) \neq 0),
    \notag
\end{align}
and we proceed by obtaining bounds on these two terms separately.
Analogously to the proof of Theorem~\ref{thm:undetectability}, the philosophy here is that the first term is `small' since it only contains those points which are misclassified by a slightly worse classifier, while the second term is small because only a small fraction of the sampled perturbations $s \in \mathbb{B}^n_\delta$ are sufficiently large to push the points across the decision boundary.

To bound the first term of~\eqref{eq:supplementary:generalisation:undetectabilitySplitting}, we use the lower bound of Lemma~\ref{lem:supplementary:generalisation:sLowerBound} on $\sigma$ in terms of $d_S$ to show that the condition $\sigma(x) \leq t$ implies that $|d_S(x)| \leq \frac{t}{\sin\theta}$.
From this, the set inclusion
\begin{align*}
    K
    \subset
    V =
    \Big\{ x \in \mathbb{R}^n \suchthat d_S(x) \geq -\frac{t}{\sin\theta} \Big \}
\end{align*}
follows, enabling us to simplify the term to be bounded as
\begin{align*}
    &P(x \sim \mathcal{D}, s \sim \mathcal{U}(\mathbb{B}^n_{\delta}) \suchthat x \in K \text{ and } f(x + s) \neq 0)
    \\&
    \leq
    P(x \sim \mathcal{D} \suchthat x \in K)
    \leq
    P(x \sim \mathcal{D} \suchthat x \in V).
\end{align*}
Recalling the definition of the events $A(\alpha)$ and $B(\beta)$ introduced in~\eqref{eq:summplementary:generalisation:eventsAB}, for any $\alpha \geq 0$ the event $A(\alpha) \wedge B(- \alpha - \frac{t}{\sin\theta})$ implies that the event $x \not\in V$ holds, and therefore
\begin{align*}
    P(x \sim \mathcal{D} \suchthat x \in V) 
    &= 1 - P(x \sim \mathcal{D} \suchthat x \not\in V)
    \\&
    \leq
    1 - P\Big( A(\alpha) \wedge B \Big( - \alpha - \frac{t}{\sin\theta} \Big) \Big).
\end{align*}
Inverting this final probability, the union bound implies that
\begin{align*}
    P(x \sim \mathcal{D} \suchthat x \in V) 
    &\leq 
    P( \notword A(\alpha)) 
    \\&\qquad
    + P\Big( \notword B \Big( - \alpha - \frac{t}{\sin\theta} \Big) \Big),
\end{align*}
and since $\alpha \geq 0$ was arbitrary it follows that for any $t \in [0, \delta]$
\begin{align}\label{eq:supplementary:generalisation:undetectability:firstTermBound}
    &P(x \sim \mathcal{D} \suchthat x \in V) 
    \\&
    \leq 
    \inf_{\alpha \geq 0}
    \Big(
        P( x \sim \mathcal{D} \suchthat |\sfunc(\proj x)| \geq \alpha)
        \notag
        \\&\qquad\qquad
        + 
        P\Big( x \sim \mathcal{D} \suchthat d_{\plane}(x) \geq - \alpha - \frac{t}{\sin\theta} \Big)
    \Big),
    \notag
\end{align}
which completes our bound on the first term of~\eqref{eq:supplementary:generalisation:undetectabilitySplitting}.

Turning to the second term of~\eqref{eq:supplementary:generalisation:undetectabilitySplitting}, 
we can simplify things by including the set $U$ into the larger set
\begin{align*}
    U \subset
    G = 
    \{ x \in \mathbb{R}^n \suchthat d_S(x) < -t
    \},
\end{align*}
where the inclusion holds due to the upper bound~\eqref{eq:generalisation:sBound} on $\sigma$.
The reason for this inclusion is that it allows us to study the intersection of the cone $C(x)$ of points with the same classification as $x$ (the existence of which is ensured by the Lipschitz property on $\sfunc$) with the ball of perturbed data points $\mathbb{B}^n_{\delta}(x)$.
Specifically, for any $x \in G$, define the set 
\begin{align*}
    H(x) = \mathbb{B}^n_{\delta}(x) \setminus (\mathbb{B}^n_{\delta}(x) \cap C(x))
\end{align*}
of perturbations of $x$ which are taken outside the cone $C(x)$ of points guaranteed to be correctly classified.

Suppose that $L \leq 1$.
Then, for $t > \delta L$ the set $H(x)$ may be included in a spherical cap which forms less than a hemisphere of $\mathbb{B}^n_{\delta}(x)$, and which may itself be contained in the larger spherical cap
\begin{align*}
    H(x) \subset \{ y \in \mathbb{B}^n_\delta(x) \suchthat (y - x) \cdot \planenormal > |d_S(x)| - \delta L \}.
\end{align*}
Since $x \in G$ implies that $d_S(x) < -t$, it follows that
\begin{align*}
    H(x) \subset J(x) = \{ y \in \mathbb{B}^n_\delta(x) \suchthat (y - x) \cdot \planenormal > t - \delta L \},
\end{align*}
and Lemma~\ref{lem:capVolumeBound} implies that the volume of $J(x)$ may be bounded by
\begin{align*}
    \frac{1}{2} V^n \delta^n \Big(1 - \Big( \frac{t}{\delta} - L \Big)^2 \Big)^{\frac{n}{2}}
\end{align*}
Consequently, since perturbations are sampled uniformly from $\mathbb{B}^n_{\delta}$, we obtain the bound
\begin{align*}
    &P(x \sim \mathcal{D}, s \sim \mathcal{U}(\mathbb{B}^n_{\delta}) \suchthat x \in U \text{ and } f(x + s) \neq 0)
    \\&
    \leq
    \frac{1}{2} \Big(1 - \Big( \frac{t}{\delta} - L \Big)^2 \Big)^{\frac{n}{2}}
    P(x \sim \mathcal{D} \suchthat x \in G).
\end{align*}
To compute the probability of sampling $x \in G$, 
we once again recall the definition of the events $A(\alpha)$ and $B(\beta)$ introduced in~\eqref{eq:summplementary:generalisation:eventsAB}, and observe that for any $\gamma \geq 0$
\begin{align*}
    A(\gamma) \wedge \notword B(\gamma - t)
    \quad \Rightarrow \quad
    d_S(x) > -t,
\end{align*}
and therefore, since
\begin{align*}
    &P(x \sim \mathcal{D} \suchthat x \in G) = 1 - P(x \sim \mathcal{D} \suchthat d_S(x) > -t)
    \\&
    \leq
    1 - P(x \sim \mathcal{D} \suchthat A(\gamma) \wedge \notword B(\gamma - t)),
\end{align*}
it follows from negating this event, applying the union bound, and recalling that $\gamma \geq 0$ was arbitrary, that
\begin{align}\label{eq:supplementary:generalisation:undetectability:secondTermBound}
    &P(x \sim \mathcal{D} \suchthat x \in G) 
    \\&
    \leq 
    \inf_{\gamma \geq 0} 
    \big[
        P(x \sim \mathcal{D} \suchthat d_{\plane}(x) \leq \gamma - t)
        \\&\qquad\qquad
        +
        P(x \sim \mathcal{D} \suchthat |\sfunc(\proj x)| > \gamma) 
    \big].
\end{align}

For $L > 1$, however, it is not possible to take $t > \delta L$ since $t \in [0, \delta]$ and so selecting $t = \delta$ provides an optimal result here.
In this case, the set $U$ is empty, so this term is simply zero.

Combining the bounds~\eqref{eq:supplementary:generalisation:undetectability:firstTermBound} and~\eqref{eq:supplementary:generalisation:undetectability:secondTermBound}, we therefore find that
\begin{align*}
    &\undetectability_0(f) 
    \\&
    \leq
    \inf_{\substack{\alpha, \gamma \geq 0 \\ t \in T(L)}}
    \Big[
        P( x \sim \mathcal{D} \suchthat |\sfunc(\proj x)| \geq \alpha)
        \\&\qquad\qquad
        + 
        P\Big( x \sim \mathcal{D} \suchthat d_{\plane}(x) \geq - \alpha - \frac{t}{\sin\theta} \Big)
        \\&\qquad\qquad
        +
        \Delta(L)
        \frac{1}{2} \Big(1 - \Big( \frac{t}{\delta} - L \Big)^2 \Big)^{\frac{n}{2}}
        \cdot
        \\&\qquad\qquad\qquad
        \cdot
        \big(
            P(x \sim \mathcal{D} \suchthat d_{\plane}(x) \leq \gamma - t)
            \\&\qquad\qquad\qquad
            +
            P(x \sim \mathcal{D} \suchthat |\sfunc(\proj x)| > \gamma) 
        \big)
    \Big],
    \notag
\end{align*}
where $\Delta(L) = 1$ for $L \leq 1$ and 0 for $L > 1$, and the set $T(L) = [\min\{L, 1\}\delta, \delta]$.

\subsubsection{Proof of Corollary~\ref{corr:supplementary:generalisation:undetectability}}

Since $\sfunc \equiv 0$, it follows that $L = 0$ and therefore $\sin \theta = 1$.
Applying these facts to the result of Theorem~\ref{thm:supplementary:generalisation:undetectability}, and selecting $\alpha = \gamma = 0$, we immediately find that
\begin{align}\label{eq:supplementary:generalisation:undetectability:corollarySplitting}
    &P(x \sim \mathcal{E}, s \sim \mathbb{B}^n_{\delta} \suchthat f(x + s) \neq 0)
    \\&\qquad
    \leq
    \inf_{t \in [0, \delta]}
    \Big[
        P( x \sim \mathcal{E} \suchthat d_{\plane}(x) \geq - t )
        \notag
        \\&\qquad\qquad
        +
        \frac{1}{2} \Big(1 - \Big( \frac{t}{\delta} \Big)^2 \Big)^{\frac{n}{2}}
        \big(
            P(x \sim \mathcal{E} \suchthat d_{\plane}(x) \leq - t)
        \big)
    \Big].
    \notag
\end{align}
Using the crude bound
\begin{align*}
    P\Big( x \sim \mathcal{E} \suchthat d_{\plane}(x) \geq - t \Big) \leq 1,
\end{align*}
this may be simplified to
\begin{align*}
    &P(x \sim \mathcal{E}, s \sim \mathbb{B}^n_{\delta} \suchthat f(x + s) \neq 0)
    \\&\quad
    \leq
    \inf_{t \in [0, \delta]}
    \Big[
        P( x \sim \mathcal{E} \suchthat d_{\plane}(x) \geq - t ) + \frac{1}{2} \Big(1 - \Big( \frac{t}{\delta} \Big)^2 \Big)^{\frac{n}{2}}
    \Big].
\end{align*}
Recalling the bound on the density $p$ of $\mathcal{E}$ in Definition~\ref{def:smac},
we have
\begin{align*}
    P( x \sim \mathcal{E} \suchthat d_{\plane}(x) \geq - t )
    \leq
    \frac{A}{V^n r^n} 
    \int_{\mathbb{B}^n_r(c)} \mathbb{I}_{\{ x \suchthat d_{\plane}(x) \geq -t \}} dx,
\end{align*}
and, arguing as in the proof of Corollary~\ref{corr:supplementary:generalisation:accuracy}, we note that this may be bounded by
\begin{align*}
    P(x \sim \mathcal{E} \suchthat d_{\plane} (x) \geq -t)
    \leq
    \frac{1}{2}A \Big(1 - \Big(\frac{\eta - t}{r}\Big)^2\Big)^{\frac{n}{2}}
\end{align*}
for any $t \in [0, \delta]$.
Substituting this bound into~\eqref{eq:supplementary:generalisation:undetectability:corollarySplitting} and selecting $t = \frac{\eta \delta}{r + \delta}$ (which is a valid choice of $t$ because $\frac{\eta \delta}{r + \delta} \in [0, \frac{\eta}{r + 1}]$ for $\delta \in [0, 1]$ and $\eta \in [0, r)$) produces the result.

\subsection{Universality of adversarial perturbations for the general model: Proof of Theorem~\ref{thm:supplementary:generalisation:universality} and Corollary~\ref{corr:supplementary:generalisation:universality}}
\label{sec:supplementary:generalisation:universalityProof}

\subsubsection{Proof of Theorem~\ref{thm:supplementary:generalisation:universality}}
Since $\phi$ satisfies the Lipschitz condition with parameter $L$, a simple geometric argument shows that if $x \in \mathbb{R}^n$ is such that
\begin{align*}
    d_S(z)  \leq d_S(x) - 2L\delta + \gamma,
\end{align*}
then $f(z + s) > \gamma \implies f(x + s) > 0$ for all $s \in \mathbb{B}^n_\delta$.
Therefore, we bound the probability
\begin{align*}
    P(x, z \sim \mathcal{D} \suchthat d_S(z) \leq d_S(x) - 2L\delta + \gamma).
\end{align*}
For any $t \in \Real$, this probability is at least the probability that
\begin{align*}
    P(x, z \sim \mathcal{D} \suchthat &d_S(z) \leq d_S(x) - 2L\delta + \gamma
    \\&
    \text{ and } 
    d_S(x) > t + L\delta - \frac{1}{2}\gamma).
\end{align*}
When $d_S(x) > t + L\delta - \frac{1}{2}\gamma$, the condition that $d_S(z) \leq t - L\delta + \frac{1}{2} \gamma$ implies that $d_S(z) \leq d_S(x) - 2L\delta + \gamma$, and therefore the probability above is bounded from below by
\begin{align*}
    P(x, z \sim \mathcal{D} \suchthat d_S(z) \leq t - L\delta + \frac{1}{2}\gamma \text{ and } d_S(x) > t + L\delta - \frac{1}{2}\gamma),
\end{align*}
which may be expressed as the product
\begin{align*}
    &P(z \sim \mathcal{D} \suchthat d_S(z) \leq t - L\delta + \frac{1}{2}\gamma)\cdot
    \\&\quad\cdot
    P(x \sim \mathcal{D} \suchthat d_S(x) > t + L\delta - \frac{1}{2}\gamma),
\end{align*}
since $x$ and $z$ are sampled independently.

Arguing as in the proofs of the previous theorems, using the definitions of the events $A(\alpha)$ and $B(\beta)$ introduced in~\eqref{eq:summplementary:generalisation:eventsAB}, we find that, for any $\alpha \geq 0$,
\begin{align*}
    &P(z \sim \mathcal{D} \suchthat d_S(z) \leq t - L\delta + \frac{1}{2}\gamma) 
    \\
    &\quad\geq 
    P(z \sim \mathcal{D} \suchthat |\phi(\Pi z)| \leq \alpha)
    - P(z \sim \mathcal{D} \suchthat d_\pi(z) > t + \chi),
\end{align*}
and
\begin{align*}
    &P(x \sim \mathcal{D} \suchthat d_S(x) > t + L\delta - \frac{1}{2}\gamma) 
    \\
    &\quad\geq 
    P(x \sim \mathcal{D} \suchthat |\phi(\Pi x)| \leq \alpha)
    - P(x \sim \mathcal{D} \suchthat d_\pi(x) \leq t - \chi),
\end{align*}
where $\chi = \frac{1}{2}\gamma - L\delta - \alpha$.
The result of the theorem therefore follows since $\alpha$ and $t$ were arbitrary.

\subsubsection{Proof of Corollary~\ref{corr:supplementary:generalisation:universality}}
In this scenario $\sfunc \equiv 0$, and we have $L = 0$ and may therefore take $\alpha = 0$.
This then implies that $\chi = \frac{1}{2} \gamma$, and so, selecting $t = -\eta$ (where we recall that $\eta = d_S(c)$ for the SmAC distribution $\mathcal{E}$), the bound from Theorem~\ref{thm:supplementary:generalisation:universality} becomes
\begin{align*}
    &P( x, z \sim \mathcal{E} \suchthat f(x + s) \neq 0 \text{ for all } s \in S_z(\delta))
    \\&
    \geq
        \Big( 
            1 - P(z \sim \mathcal{E} \suchthat d_\pi(z) > -\eta + \frac{1}{2} \gamma)
        \Big)
        \cdot
        \\&\qquad\qquad
        \cdot
        \Big( 
            1 - P(x \sim \mathcal{E} \suchthat d_\pi(x) \leq -\eta - \frac{1}{2} \gamma)
        \Big).
\end{align*}
Noting that the bound does not depend on $\delta$, we may switch to using the generic $S_z$ rather than $S_z(\delta)$.

The bound on the density guaranteed by the SmAC property in Definition~\ref{def:smac} implies that
\begin{align*}
    &P( x, z \sim \mathcal{E} \suchthat f(x + s) \neq 0 \text{ for all } s \in S_z)
    \\&
    \geq
        \Big( 
            1 - \frac{A}{V^nr^n} \int_{\mathbb{B}^n_r(c)} \mathbb{I}_{d_\pi(z) > -\eta + \frac{1}{2} \gamma} dz
        \Big)
        \cdot
        \\&\qquad\qquad
        \cdot
        \Big( 
            1 - \frac{A}{V^nr^n} \int_{\mathbb{B}^n_r(c)} \mathbb{I}_{d_\pi(x) \leq -\eta - \frac{1}{2} \gamma}dx
        \Big),
\end{align*}
and we observe that both integrals are simply the volume of a spherical cap with height $r - \frac{1}{2}\gamma$, and Lemma~\ref{lem:capVolumeBound} therefore implies that
\begin{align*}
    &P( x, z \sim \mathcal{E} \suchthat f(x + s) \neq 0 \text{ for all } s \in S_z)
    \\&\qquad
    \geq
    \Big(
        1 - A \Big( 1 - \frac{\gamma^2}{4r^2} \Big)^{\frac{n}{2}}
    \Big)^2
\end{align*}

\end{document}